\documentclass[10pt,twocolumn,letterpaper]{article}

\usepackage[pagenumbers]{wacv} 

\definecolor{wacvblue}{rgb}{0.21,0.49,0.74}
\usepackage[pagebackref,breaklinks,colorlinks,allcolors=wacvblue]{hyperref}

\usepackage{tabularx} 
\usepackage{threeparttable} 
\usepackage{adjustbox}

\newcommand{\tablefont}{\fontsize{7.5pt}{9pt}\selectfont}

\usepackage{anyfontsize}

\usepackage{caption}
\usepackage{amssymb}
\usepackage{graphics}
\newcommand{\newcheckmark}{\raisebox{0.6ex}{\scalebox{1.0}{\textcolor{teal}{$\sqrt{}$}}}}
\newcommand{\newcrossmark}{\scalebox{1.0}[1]{\textcolor{purple}{$\times$}}}

\usepackage[accsupp]{axessibility}  

\def\wacvPaperID{929} 
\def\confName{WACV}
\def\confYear{2026}

\title{Reverse Personalization}

\newcommand{\authorspace}{\hspace{3em}}

\author{
  Han-Wei Kung$^1$ \authorspace Tuomas Varanka$^2$ \authorspace Nicu Sebe$^1$ \\
  \vspace{-3mm} \\
  $^1$University of Trento \authorspace $^2$University of Oulu \\
  \vspace{-3mm} \\
  {\tt\small hanwei.kung@unitn.it}
}

\newcolumntype{R}[1]{>{\centering\arraybackslash}m{#1}}

\begin{document}

\twocolumn[{
  \renewcommand\twocolumn[1][]{#1}%
  \maketitle

  \begin{center}
    \captionsetup{type=figure}
    \setlength{\tabcolsep}{0.5pt}

    \resizebox{1.\textwidth}{!}{
      \begin{tabular}{
          *{1}{>{\centering\arraybackslash}m{\dimexpr.14\linewidth-2\tabcolsep}}
        }
        \multicolumn{1}{c}{\small Input} \\
        \midrule
        \multicolumn{1}{c}{\includegraphics[width={\dimexpr.14\linewidth}]{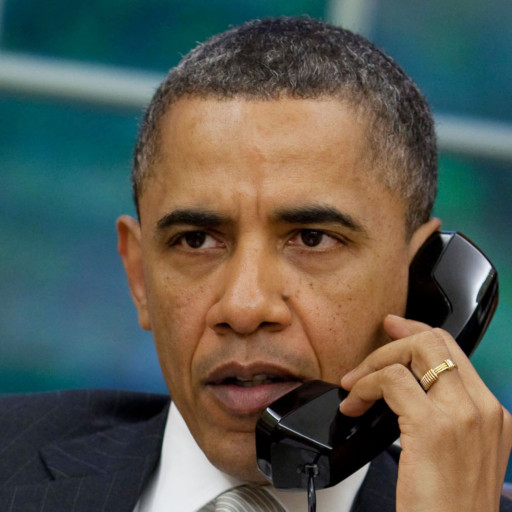}} \\
        \multicolumn{1}{c}{\includegraphics[width={\dimexpr.14\linewidth}]{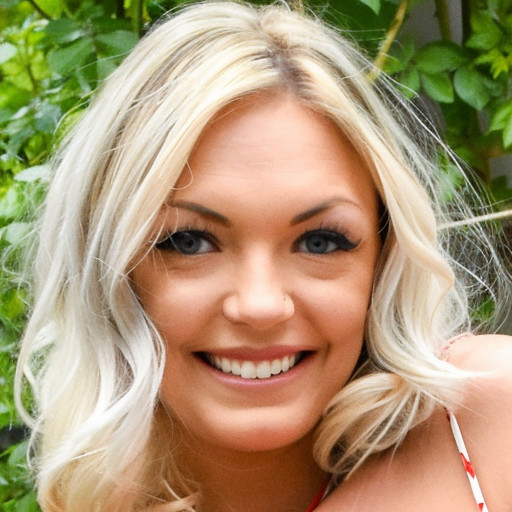}} \\
        \\
        \midrule
        \multicolumn{1}{l}{\small Identity anonymized} \\
        \multicolumn{1}{l}{\small Subject agnostic} \\
        \multicolumn{1}{l}{\small Attr. \& scene kept} \\
        \multicolumn{1}{l}{\small Attr. controllable} \\
      \end{tabular}
      \quad
      \begin{tabular}{
          *{1}{>{\centering\arraybackslash}m{\dimexpr.14\linewidth-2\tabcolsep}}
        }
        \multicolumn{1}{c}{\small LDFA~\cite{klemp2023ldfa}} \\
        \midrule
        \multicolumn{1}{c}{\includegraphics[width={\dimexpr.14\linewidth}]{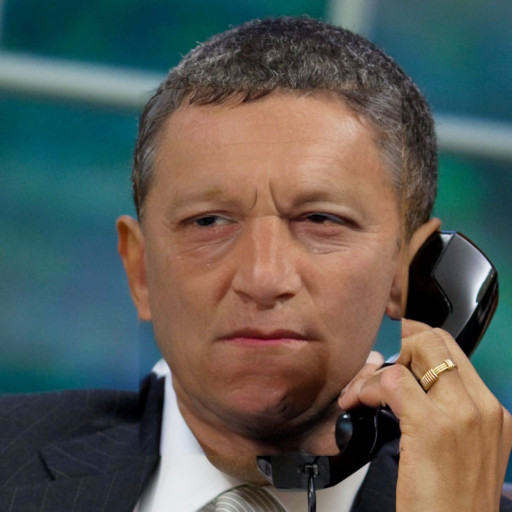}} \\
        \multicolumn{1}{c}{\includegraphics[width={\dimexpr.14\linewidth}]{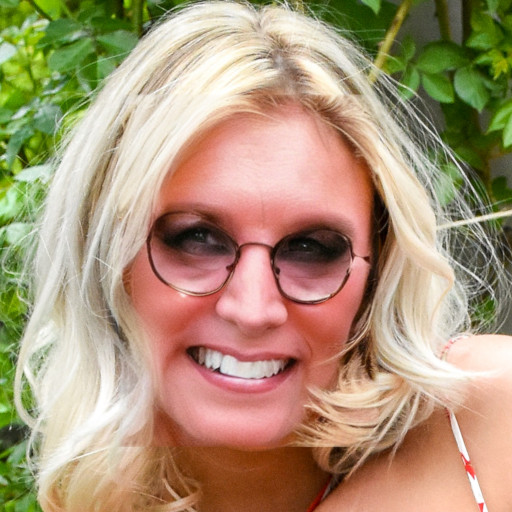}} \\
        \\
        \midrule
        \multicolumn{1}{c}{\small $\newcheckmark$ Yes} \\
        \multicolumn{1}{c}{\small $\newcheckmark$ Yes} \\
        \multicolumn{1}{c}{\small $\newcrossmark$ Poor} \\
        \multicolumn{1}{c}{\small $\newcrossmark$ No} \\
      \end{tabular}
      \quad
      \begin{tabular}{
          *{1}{>{\centering\arraybackslash}m{\dimexpr.14\linewidth-2\tabcolsep}}
        }
        \multicolumn{1}{c}{\small RiDDLE~\cite{li2023riddle}} \\
        \midrule
        \multicolumn{1}{c}{\includegraphics[width={\dimexpr.14\linewidth}]{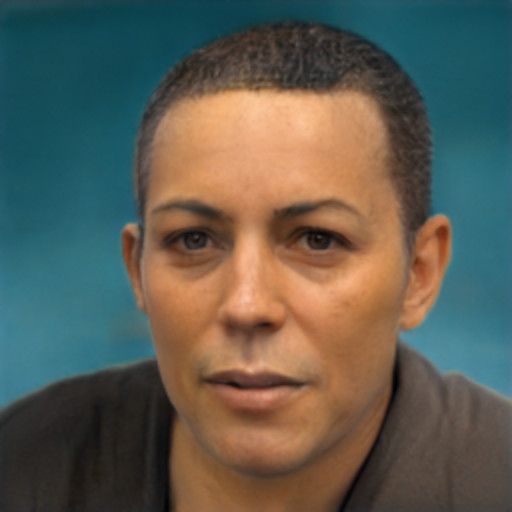}} \\
        \multicolumn{1}{c}{\includegraphics[width={\dimexpr.14\linewidth}]{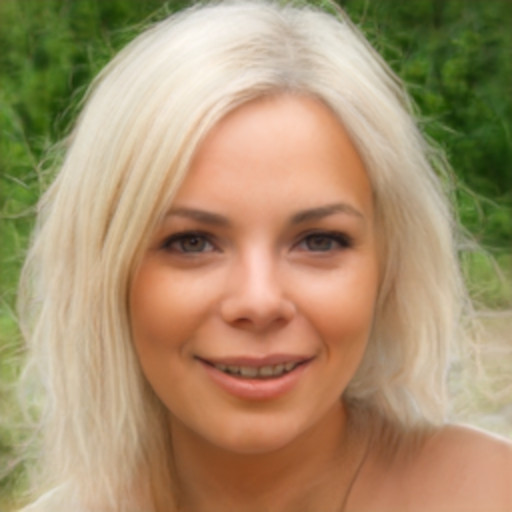}} \\
        \\
        \midrule
        \multicolumn{1}{c}{\small $\newcheckmark$ Yes} \\
        \multicolumn{1}{c}{\small $\newcheckmark$ Yes} \\
        \multicolumn{1}{c}{\small $\newcrossmark$ Poor} \\
        \multicolumn{1}{c}{\small $\newcrossmark$ No} \\
      \end{tabular}
      \quad
      \begin{tabular}{
          *{1}{>{\centering\arraybackslash}m{\dimexpr.14\linewidth-2\tabcolsep}}
        }
        \multicolumn{1}{c}{\small Textual Inversion~\cite{gal2022image}} \\
        \midrule
        \multicolumn{1}{c}{\includegraphics[width={\dimexpr.14\linewidth}]{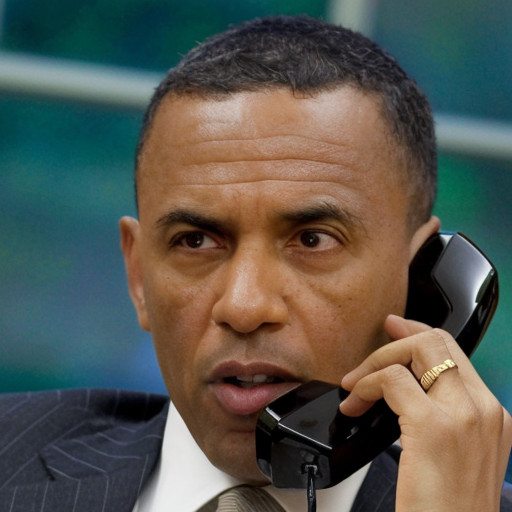}} \\
        \multicolumn{1}{c}{\includegraphics[width={\dimexpr.14\linewidth}]{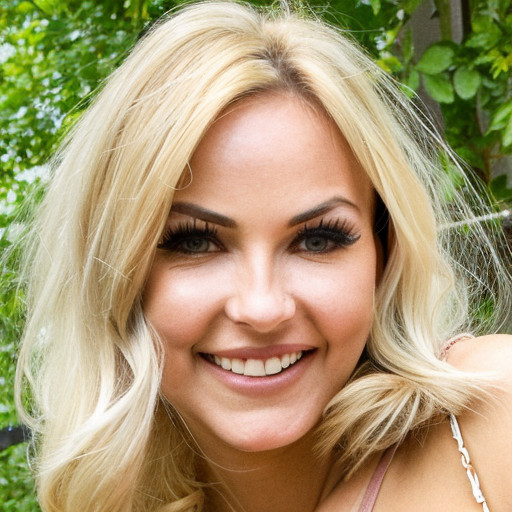}} \\
        \\
        \midrule
        \multicolumn{1}{c}{\small $\newcheckmark$ Yes} \\
        \multicolumn{1}{c}{\small $\newcrossmark$ No} \\
        \multicolumn{1}{c}{\small $\newcheckmark$ Good} \\
        \multicolumn{1}{c}{\small $\newcrossmark$ No} \\
      \end{tabular}
      \quad
      \begin{tabular}{
          *{3}{>{\centering\arraybackslash}m{\dimexpr.14\linewidth-2\tabcolsep}}
        }
        \multicolumn{3}{c}{\small Ours} \\
        \midrule
        \multicolumn{1}{c}{\includegraphics[width={\dimexpr.14\linewidth}]{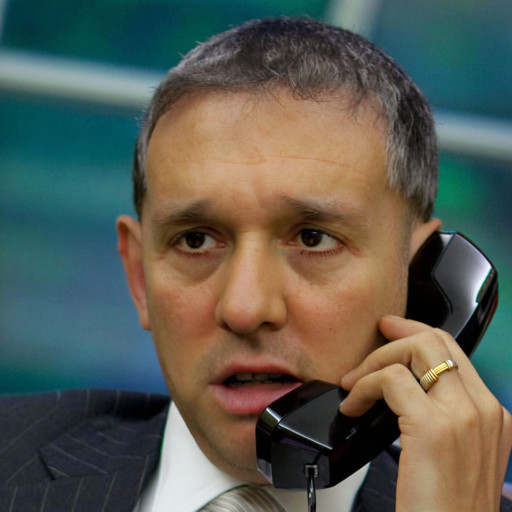}} & \multicolumn{1}{c}{\includegraphics[width={\dimexpr.14\linewidth}]{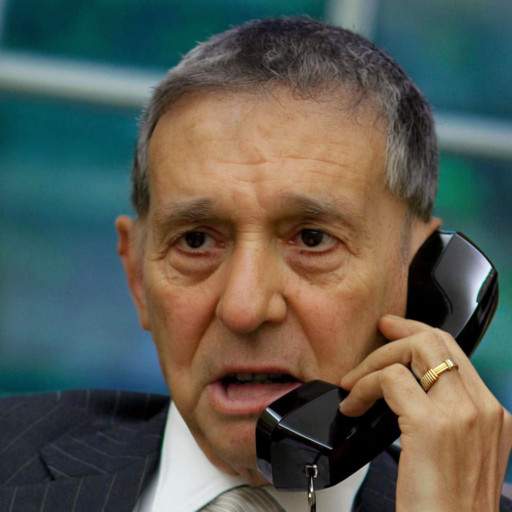}} & \multicolumn{1}{c}{\includegraphics[width={\dimexpr.14\linewidth}]{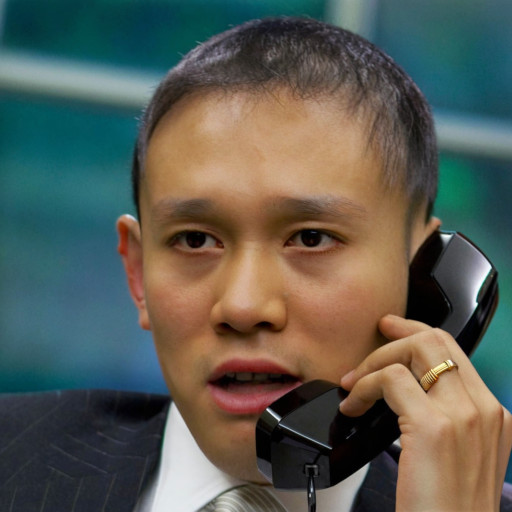}} \\
        \multicolumn{1}{c}{\includegraphics[width={\dimexpr.14\linewidth}]{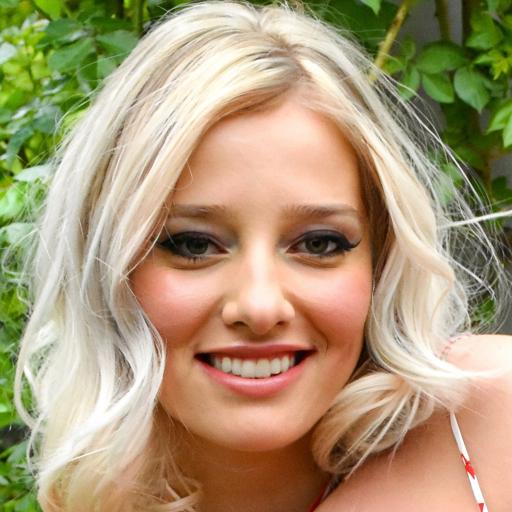}} & \multicolumn{1}{c}{\includegraphics[width={\dimexpr.14\linewidth}]{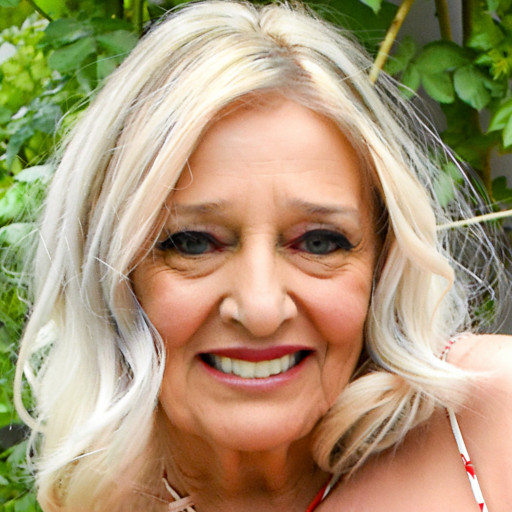}} & \multicolumn{1}{c}{\includegraphics[width={\dimexpr.14\linewidth}]{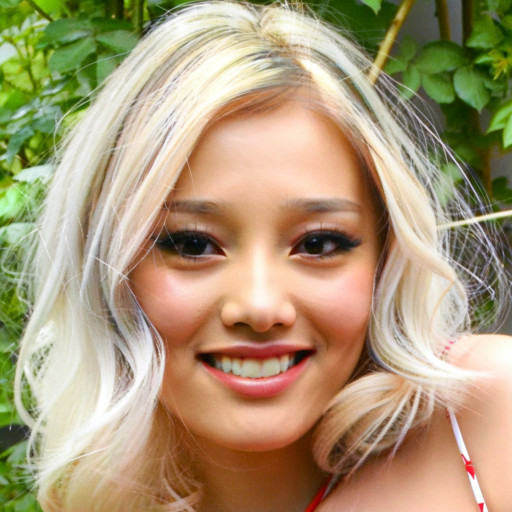}} \\
        \multicolumn{1}{c}{\footnotesize Default} & \multicolumn{1}{c}{\footnotesize Change age (older)} & \multicolumn{1}{c}{\footnotesize Change race (Asian)} \\
        \midrule
        \multicolumn{3}{c}{\small $\newcheckmark$ Yes} \\
        \multicolumn{3}{c}{\small $\newcheckmark$ Yes} \\
        \multicolumn{3}{c}{\small $\newcheckmark$ Good} \\
        \multicolumn{3}{c}{\small $\newcheckmark$ Yes} \\
      \end{tabular}
    }

    \captionof{figure}{Our reverse personalization method removes identity-specific features while preserving both the original facial attributes and surrounding scene---without requiring subject finetuning. It also provides intuitive user control over which attributes are retained or modified, enabling flexible and customizable anonymization for downstream applications.}
    \label{fig:teaser}
  \end{center}
}]


\begin{abstract}
  Recent text-to-image diffusion models have demonstrated remarkable generation of realistic facial images conditioned on textual prompts and human identities, enabling creating personalized facial imagery. However, existing prompt-based methods for removing or modifying identity-specific features rely either on the subject being well-represented in the pre-trained model or require model fine-tuning for specific identities. In this work, we analyze the identity generation process and introduce a reverse personalization framework for face anonymization. Our approach leverages conditional diffusion inversion, allowing direct manipulation of images without using text prompts. To generalize beyond subjects in the model's training data, we incorporate an identity-guided conditioning branch. Unlike prior anonymization methods, which lack control over facial attributes, our framework supports attribute-controllable anonymization. We demonstrate that our method achieves a state-of-the-art balance between identity removal, attribute preservation, and image quality. Source code and data are available at \url{https://github.com/hanweikung/reverse-personalization}.
\end{abstract}


\section{Introduction}

Given an image of a human face, how can we remove identity-specific features while preserving non-identity attributes? Recent advances in diffusion models have enabled the creation of realistic visual content~\cite{rombach2022high,podell2023sdxl}, including face synthesis~\cite{kim2023dcface,huang2023collaborative,boutros2023idiff,papantoniou2024arc2face}. Several studies~\cite{wang2024instantid,guo2024pulid,li2024photomaker} have demonstrated the effectiveness of these models in generating faces from prompts and identities. Additionally, research~\cite{hertz2022prompt,mokady2023null,huberman2024edit,brack2024ledits++} has shown that manipulating attention weights on specific tokens within prompts can influence semantic alignment and provide fine-grained control.

Building on this insight, we observe that adjusting attention weights on celebrity names within prompts controls the likeness of generated images. However, this approach presents one limitation. If the input image shows a non-celebrity, their identity may not exist in the model's learned feature space. In such cases, modifying attention weights has a negligible effect, as the model lacks a representation of the individual.

To overcome this limitation, we first employ \textit{diffusion inversion} techniques~\cite{song2020denoising,huberman2024edit}, which map an input image into the latent space of a pre-trained model. This enables us to synthesize facial images with specific traits using conditioning methods. Beyond text prompts, recent advances have introduced expressive conditioning techniques, such as face embeddings~\cite{valevski2023face0,ye2023ip,xu2024text} and semantic masks~\cite{huang2023collaborative,zhang2023adding,deng2025z}. Inspired by developments in personalization~\cite{gal2022image,ye2023ip,wang2024instantid}, we adopt identity-conditioned generation using face embeddings, allowing us to extract identity features from arbitrary input images.

Our approach offers several key advantages. As an inversion-based method, it avoids model retraining, thus preserving the original generative capabilities of the diffusion model. This also ensures compatibility with other identity-conditioned generation methods and enables flexible control over high-level facial attributes. Importantly, our method supports image-only inputs, eliminating reliance on textual instructions.

A challenge then arises: how can we modulate how generated faces reflect the identity in the input image? By analyzing the generation process, we find that increasing the classifier-free guidance~\cite{ho2022classifier} scale amplifies identity-defining features. The model appears to first synthesize a neutral face, then progressively injects characteristic traits. This insight led us to experiment with guidance scales, hypothesizing that they could suppress identity-specific features and produce a \textit{reverse} identity. Our experiments confirm this hypothesis, motivating the development of a novel mechanism termed \textit{reverse personalization}.

Reverse personalization offers compelling benefits in \textit{face anonymization}, where the goal is to protect personal identity while preserving utility. This capability is important across sectors such as healthcare~\cite{yang2022digital,mishima2019evaluation} and security~\cite{grgic2011scface,kasim2024watchlist}. Anonymization techniques help address ethical concerns about surveillance and individual autonomy, while ensuring compliance with data privacy regulations including GDPR~\cite{gdpr} and CCPA~\cite{ccpa}.

Despite progress in GAN- and diffusion-based anonymization, current methods face persistent challenges. Many struggle to strike a balance between removing identity-specific features and preserving non-identity attributes and realism~\cite{cao2024face,meden2021privacy}. Furthermore, research has shown that individuals' willingness to disclose personal information---particularly sensitive demographic attributes such as age, race, and gender---is context-dependent~\cite{phillips2009disclose}. For instance, in workplace settings, individuals may withhold such details to avoid bias or misunderstandings, especially in environments with status differences or limited trust~\cite{phillips2009disclose}. Conversely, in healthcare contexts, disclosing demographic information can lead to improved outcomes by enabling more personalized and effective treatment~\cite{fiske2025weighing}. However, existing face anonymization methods lack control over whether such attributes are retained or altered~\cite{kuang2024facial}. In contrast, our reverse personalization framework addresses this limitation, enabling users to flexibly control facial attributes in anonymized outputs.

We analyze identity-conditioned generation and apply our findings to face anonymization. We demonstrate that reverse personalization removes identifiable facial features while maintaining realism and attribute consistency. We contribute:

\begin{itemize}
  \item \textit{Conditional inversion:} we present a conditioning strategy that guides the inversion process to facilitate identity manipulation.
  \item \textit{Reverse personalization:} we propose a guidance mechanism that steers the generative process away from identity-defining features, enabling anonymization while maintaining realism. Our method achieves an optimal balance between identity obfuscation, attribute preservation, and visual quality.
  \item \textit{Attribute-controllable anonymization:} our approach includes intuitive controls for adjusting facial features like age, gender, and ethnicity, making it easy to generate customizable, anonymized results.
\end{itemize}


\section{Related Work}

\paragraph{Personalization.}

The field of personalized text-to-image generation~\cite{gal2022image,ruiz2023dreambooth,kumari2023multi,chen2023subject,sohn2023styledrop} focuses on adapting diffusion models to synthesize images of specific subjects, styles, or concepts that are meaningful to individual users. This line of research enables models to learn visual concepts from a few example images, allowing for tailored image generation beyond general models.

Textual Inversion~\cite{gal2022image} optimizes token embeddings linked to a placeholder token, allowing the model to incorporate new concepts into text-driven generation with a few subject images. Similarly, DreamBooth~\cite{ruiz2023dreambooth} introduces a fine-tuning approach that binds an identifier to a specific subject. It leverages the semantic priors of pre-trained diffusion models while employing a class-specific prior preservation loss to maintain visual consistency with the broader data distribution.

Despite effectiveness, these methods suffer from computational costs, requiring several minutes to hours for fine-tuning. To overcome this limitation, parameter-efficient fine-tuning methods have emerged. HyperDreamBooth~\cite{ruiz2024hyperdreambooth} introduces a hypernetwork-based architecture that generates personalized weights from a single image, offering a speedup---as much as 25 times faster than DreamBooth~\cite{ruiz2023dreambooth} and 125 times faster than Textual Inversion~\cite{gal2022image}. JeDi~\cite{zeng2024jedi} reduces computational demands by learning the joint distribution of multiple text-image pairs of a common subject, enabling finetuning-free personalization. IP-Adapter~\cite{ye2023ip} adopts an alternative strategy based on image prompt adaptation, allowing users to guide image generation using reference images while retaining controllable text prompts.

These personalization techniques have unlocked a variety of applications, including identity-preserving face generation~\cite{xu2024text,liu2024towards,li2024photomaker,xiao2024fastcomposer,wang2024instantid,guo2024pulid,shiohara2024face2diffusion}, virtual try-on~\cite{xie2024dreamvton}, and customized content creation~\cite{nam2025visual}. However, while most efforts focus on preserving identity, style, or subject appearance, we repurpose these principles toward the opposite goal---removing identifiable facial traits. Our work extends the technical foundations of personalization to enable attribute-controllable face anonymization.

\paragraph{Face anonymization.}

Face anonymization, or face de-identification, refers to concealing or altering facial traits to protect individual privacy in images. Traditional methods, such as blurring, pixelation, and masking, are simple and effective at preventing identification by humans. However, these approaches degrade the visual quality of images, destroying important contextual information like facial expressions, pose, and background---limiting their usefulness for downstream computer vision tasks~\cite{meden2021privacy,cao2024face}.

In response, modern face anonymization techniques strive for balance between privacy protection and data utility. Rather than obscuring faces, these methods aim to prevent identity recognition while preserving identity-agnostic attributes crucial for tasks such as emotion analysis~\cite{kim2022optimal,jiang2022disentangling} and estimating head pose~\cite{liu2022arhpe} and gaze~\cite{wang2022contrastive}. Among these, GAN-based methods~\cite{maximov2020ciagan,dall2022graph,helou2023vera,cai2024disguise} have been explored. DP~\cite{hukkelaas2019deepprivacy} and its successor DP2~\cite{hukkelaas2023deepprivacy2} employ conditional GANs~\cite{mirza2014conditional} to generate synthetic faces that maintain pose and background context. GANonymization~\cite{hellmann2024ganonymization} further advances this idea by focusing on preserving facial expressions during anonymization. Methods like FALCO~\cite{barattin2023attribute} and RiDDLE~\cite{li2023riddle} exploit the latent space of StyleGAN2~\cite{karras2020analyzing} to generate realistic anonymized faces while retaining non-identity features.

Driven by recent progress in generative modeling, diffusion models have emerged as promising for face anonymization. For example, LDFA~\cite{klemp2023ldfa} combines face detection with a latent diffusion model to generate in-painted faces. FAMS~\cite{Kung_2025_WACV} uses only reconstruction loss, eliminating dependence on identity losses derived from face recognition models or the use of facial landmarks and masks, which are prone to inaccuracies.

Despite these advancements, challenges remain. Achieving an balance between anonymization and preservation of image utility is difficult~\cite{hukkelaas2023does}. Additionally, most methods offer limited controllability over which sensitive attributes are preserved or concealed during anonymization. To address the limitations, our reverse personalization framework leverages the precise reconstruction and generative power of diffusion models. Our method enables user-controlled anonymization by allowing selective preservation or removal of specific facial attributes, while maintaining realism and utility for identity-agnostic tasks.


\section{Method}

\subsection{Preliminary}

\paragraph{DDPM Inversion.}

Denoising Diffusion Probabilistic Models (DDPMs)~\cite{ho2020denoising} define a forward process that gradually adds Gaussian noise to data and a learned reverse process that denoises it step by step. The reverse process reconstructs \( x_{t-1} \) from a noisy input \( x_t \) as follows:

\begin{equation}
	x_{t-1} = \hat{\mu}_t(x_t) + \sigma_t z_t, \quad z_t \sim \mathcal{N}(0, I),
\end{equation}

\noindent where \( \hat{\mu}_t(x_t) \) is the predicted mean, \( \sigma_t \) is the variance schedule, and \( z_t \) is standard Gaussian noise. To recover the noise used at each step, we compute:

\begin{equation}
	z_t = \frac{x_{t-1} - \hat{\mu}_t(x_t)}{\sigma_t}.
\end{equation}

\paragraph{IP-Adapter.}

The IP-Adapter~\cite{ye2023ip} is a lightweight module that enables diffusion models to use image prompts by modifying their attention mechanisms. Specifically, it integrates visual information without requiring model retraining. The modified attention is computed as:

\begin{equation}
  \begin{aligned}
    \mathbf{Z}^{\text{new}} & = \text{Attention}(\mathbf{Q}, \mathbf{K}, \mathbf{V}) \\
                            & + \lambda_{ipa} \cdot \text{Attention}(\mathbf{Q}, \mathbf{K}', \mathbf{V}'),
  \end{aligned}
\end{equation}

\noindent where \( \mathbf{Q}, \mathbf{K}, \mathbf{V} \) are the original queries, keys, and values, and \( \mathbf{K}', \mathbf{V}' \) are keys and values derived from the image prompt. The IP-Adapter~\cite{ye2023ip} scale \( \lambda_{ipa} \) controls the influence of the visual prompt. This mechanism facilitates flexible multimodal conditioning while preserving the pretrained model's structure.

\subsection{Motivation}

\begin{figure}
	\centering
  \setlength{\tabcolsep}{2pt}
	\resizebox{1.\linewidth}{!}{
		\begin{tabular}{
			*{4}{m{\dimexpr.25\linewidth-2\tabcolsep}}
			}
      \multicolumn{1}{c}{\small Input} & \multicolumn{1}{c}{\small \shortstack{Null-text \\ Inversion~\cite{mokady2023null}}} & \multicolumn{1}{c}{\small \shortstack{Textual \\ Inversion~\cite{gal2022image}}} & \multicolumn{1}{c}{\small Ours} \\
			\multicolumn{1}{c}{\includegraphics[width={\dimexpr.25\linewidth}]{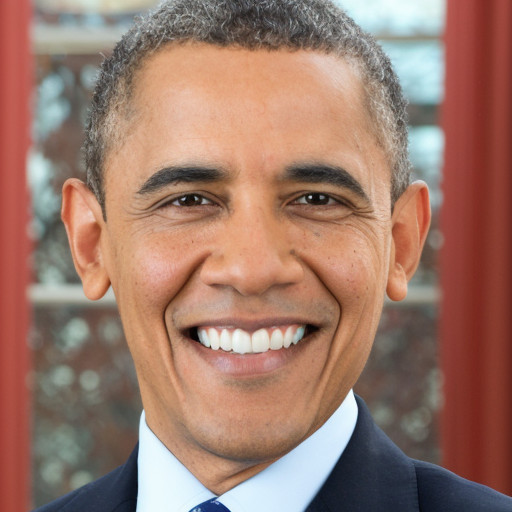}} &
      \multicolumn{1}{c}{\includegraphics[width={\dimexpr.25\linewidth}]{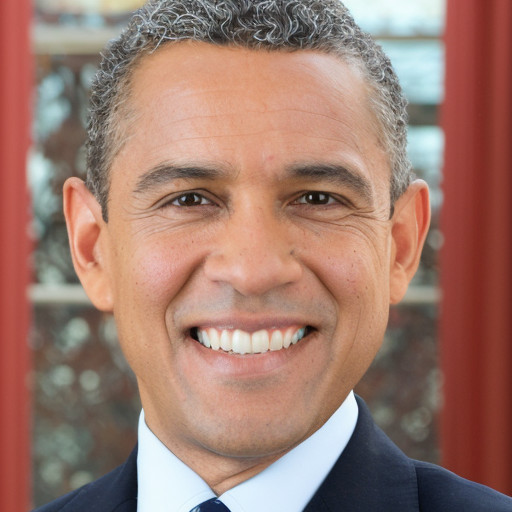}} &
      \multicolumn{1}{c}{\includegraphics[width={\dimexpr.25\linewidth}]{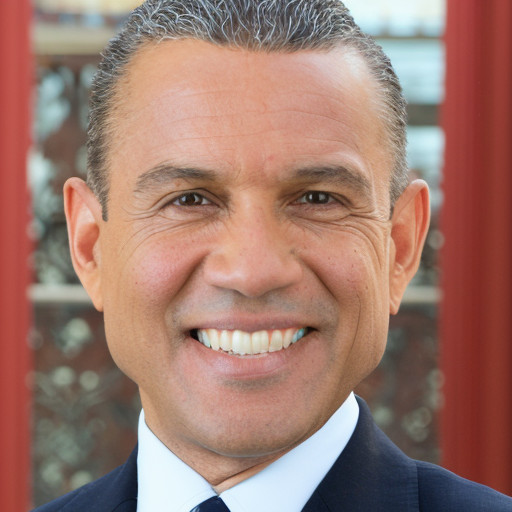}} &
      \multicolumn{1}{c}{\includegraphics[width={\dimexpr.25\linewidth}]{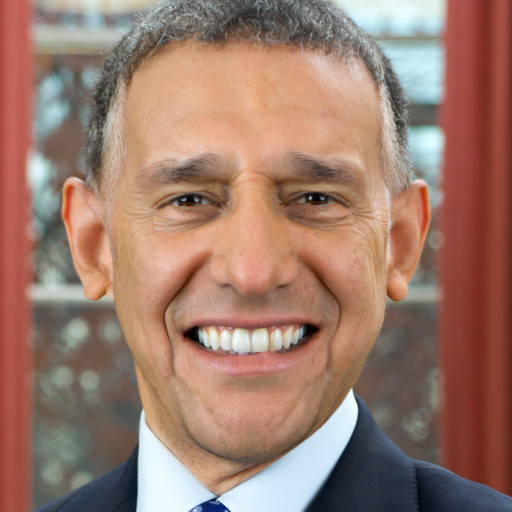}} \\
			\multicolumn{1}{c}{\includegraphics[width={\dimexpr.25\linewidth}]{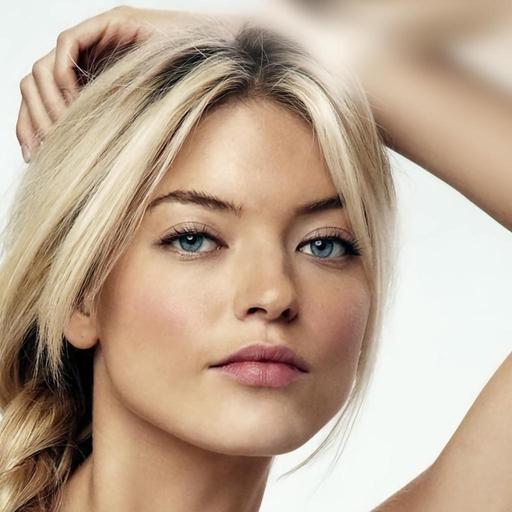}} &
      \multicolumn{1}{c}{\includegraphics[width={\dimexpr.25\linewidth}]{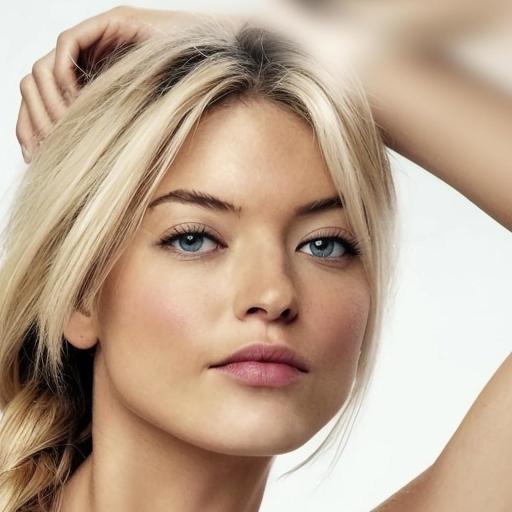}} &
      \multicolumn{1}{c}{\includegraphics[width={\dimexpr.25\linewidth}]{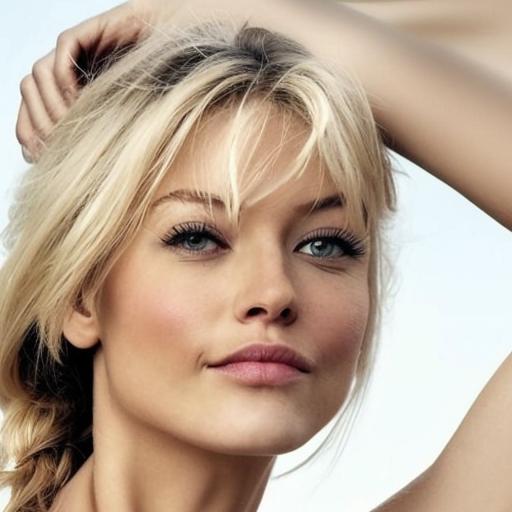}} &
      \multicolumn{1}{c}{\includegraphics[width={\dimexpr.25\linewidth}]{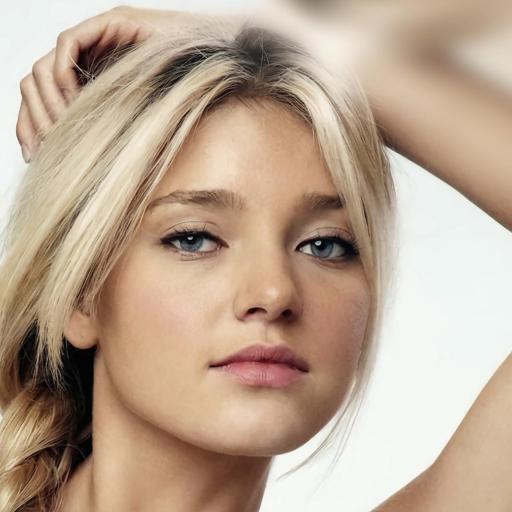}} \\
      \multicolumn{1}{l}{\small \shortstack{No prior knowl- \\ edge required}} & \multicolumn{1}{c}{\small $\newcrossmark$ No} & \multicolumn{1}{c}{\small $\newcheckmark$ Yes} & \multicolumn{1}{c}{\small $\newcheckmark$ Yes} \\
      \midrule
      \multicolumn{1}{l}{\small \shortstack{No fine-tuning \\ needed}} & \multicolumn{1}{c}{\small $\newcrossmark$ No} & \multicolumn{1}{c}{\small $\newcrossmark$ No} & \multicolumn{1}{c}{\small $\newcheckmark$ Yes} \\
		\end{tabular}
	}
  \caption{Prompt-based attention reweighting methods such as Null-text Inversion~\cite{mokady2023null} can modify facial identity only when the diffusion model knows the subject (e.g., well-known figures, such as Obama). Personalization methods like Textual Inversion~\cite{gal2022image} require fine-tuning with multiple reference images. Unlike these methods, our reverse personalization approach can modify facial identity without prior model knowledge or fine-tuning.}
	\label{fig:motivation}
\end{figure}

Given an image of a person, our goal is to modify defining facial features while preserving non-identifying features and the surrounding context. Achieving balance between data utility and identity protection is central to face anonymization, which supports scientific progress~\cite{helou2023vera} and privacy~\cite{newton2005preserving}.

An intuitive approach leverages inversion techniques in diffusion models, which map real images into latent representations compatible with a pretrained generator. One such method, Null-text Inversion~\cite{mokady2023null}, employs DDIM inversion~\cite{song2020denoising} to extract a sequence of latent codes. These are then used to optimize the null-text embedding in classifier-free guidance~\cite{ho2022classifier}, allowing the model to reconstruct the input image. This optimized embedding enables methods like Prompt-to-Prompt~\cite{hertz2022prompt} to reweight attention maps during generation, offering fine-grained control. For instance, as shown in \cref{fig:motivation}, attenuating the influence of the token ``Obama'' in the prompt ``a photo of Obama'' alters identity-specific facial features while preserving non-identity-relevant attributes like expression and the background context.

However, this strategy has a limitation: it relies on the model's knowledge of the subject's identity. When the individual is poorly represented in the training data---the woman in \cref{fig:motivation}---adjusting prompt weights has little effect on the output.

To address this, model personalization techniques, such as Textual Inversion~\cite{gal2022image}, are applied. Textual Inversion~\cite{gal2022image} fine-tunes the model's word embeddings on few images of a concept (e.g., the women in \cref{fig:teaser,fig:motivation}) and associates with a unique token $S_*$. This token can be used in prompts (e.g., ``a photo of $S_*$'') to trigger identity-aware generation.

Although this approach moves beyond the model's prior knowledge, it has challenges. It demands computational resources to fine-tune and may underperform when limited reference images are available, failing to capture the identity.

\subsection{Reverse personalization}

\begin{figure*}[ht]
  \centering

  \newlength{\imageAheight}
  \newlength{\imageBheight}

  \settoheight{\imageAheight}{\includegraphics[width={.48\linewidth}]{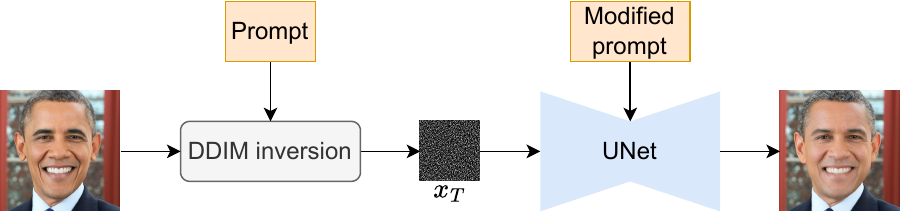}}
  \settoheight{\imageBheight}{\includegraphics[width={.48\linewidth}]{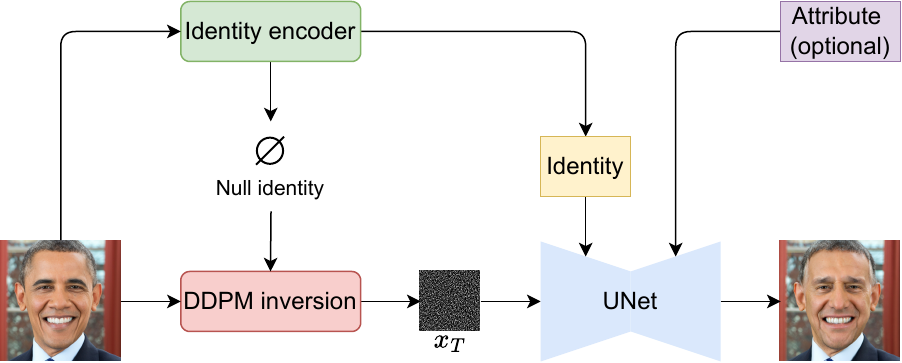}}

  \newlength{\diffAB}
  \setlength{\diffAB}{0.5\dimexpr\imageBheight - \imageAheight\relax}

  \newlength{\diffBA}
  \setlength{\diffBA}{0.5\dimexpr\imageAheight - \imageBheight\relax}

  \begin{subfigure}[b]{.48\textwidth}
    \centering
    \raisebox{\diffAB}{\includegraphics[width={1.0\linewidth}]{./images/schematic/reweighting.pdf}}
    \caption{Conventional prompt-based reweighting method.}
  \end{subfigure}
  \hfill
  \begin{subfigure}[b]{.48\textwidth}
    \centering
    \raisebox{\diffBA}{\includegraphics[width={1.0\linewidth}]{./images/schematic/ours.pdf}}
    \caption{Our identity-based method.}
  \end{subfigure}

  \caption{Our reverse personalization framework. Unlike prior prompt-based reweighting approaches that guide both inversion and generation using textual prompts, our method leverages identity information to condition these processes. The framework also supports intuitive user control, enabling selective retention or modification of semantic facial attributes.}
  \label{fig:schematic}
\end{figure*}

To address the limitations of prompt-based reweighting and fine-tuning methods, we propose an approach to modify identity-specific facial features without relying on prompt reweighting or computationally intensive personalization. Our method leverages identity embeddings extracted from facial images and integrates them through identity-conditioned adapters to modulate the diffusion process. A schematic overview of our approach is illustrated in \cref{fig:schematic}.

In standard diffusion model inversion, the denoising network uses prompt embeddings to guide the recovery of noise trajectories, ensuring that the reconstructed image matches the input prompt semantically and visually. This enables prompt-based image editing, where modifying the prompt can yield targeted changes in the output.

However, in our reverse personalization setting---where only one image is provided and the goal is to alter identity---prompt substitution is not applicable. Instead, we introduce \emph{null-identity embeddings} to guide the inversion process without identity-specific conditioning.

We further improve inversion efficiency using DPM-Solver++~\cite{lu2025dpm}, a higher-order solver that surpasses DDPM-based methods in speed~\cite{brack2024ledits++}. This also ensures perfect reconstruction of the input image, preserving structural and contextual details while enabling modification of identity-specific features.

Our inversion process is formulated as:
\begin{equation}
	z_t = \frac{x_{t-1} - \hat{\mu}_t(x_t, x_{t+1}, {\varnothing}_{id})}{\sigma_t}, \qquad t=T,\ldots,1,
\end{equation}
where $\hat{\mu}_t(x_t, x_{t+1}, {\varnothing}_{id})$ is the second-order estimate of the denoised sample, ${\varnothing}_{id}$ denotes the null identity embedding, and $\sigma_t$ is the noise variance at step $t$.

Text-to-image diffusion models employ classifier-free guidance~\cite{ho2022classifier} to improve text fidelity by interpolating between a conditional and an unconditional prediction. In contrast, we reinterpret classifier-free guidance~\cite{ho2022classifier} to control identity alignment. Specifically, we condition one forward pass on the identity embedding and leave the other unconditioned. Our reverse personalization guidance is defined as:
\begin{equation}
  \begin{aligned}
    \hat{\epsilon}_\theta({x}_t, t, c_{id}) & = \lambda_{cfg} \cdot \epsilon_\theta({x}_t, t, c_{id}) \\
                                            & + (1 - \lambda_{cfg}) \cdot \epsilon_\theta({x}_t, t, {\varnothing}_{id}),
  \end{aligned}
\end{equation}
where $\lambda_{cfg}$ is the guidance scale, $c_{id}$ is the identity embedding, and ${\varnothing}_{id}$ is the null identity.

\begin{figure}
	\centering
  \setlength{\tabcolsep}{2pt}
  \resizebox{1.\linewidth}{!}{
    \begin{tabular}{
        *{5}{m{\dimexpr0.20\linewidth-2\tabcolsep}}
      }
      \multicolumn{2}{c}{\small Generated} & \multicolumn{1}{c}{\small Input} & \multicolumn{2}{c}{\small Generated} \\
      \cmidrule(lr){1-2} \cmidrule(lr){3-3} \cmidrule(lr){4-5}
      \multicolumn{1}{c}{\includegraphics[width=.20\linewidth]{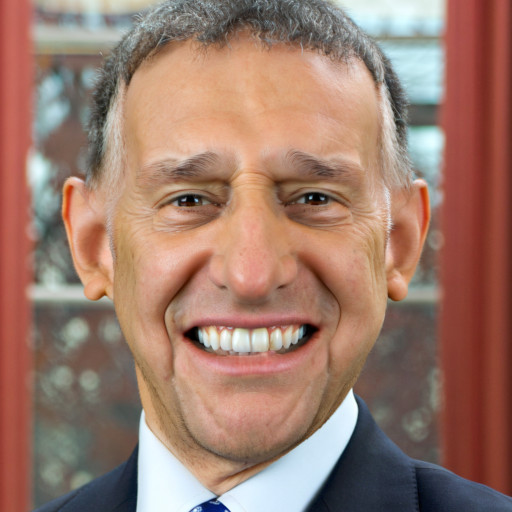}} &
      \multicolumn{1}{c}{\includegraphics[width=.20\linewidth]{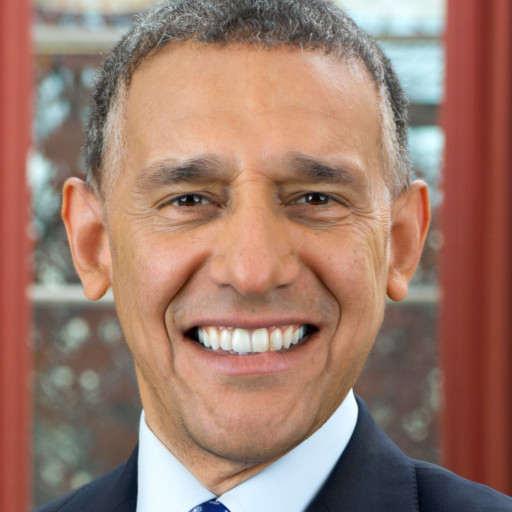}} &
      \multicolumn{1}{c}{\includegraphics[width=.20\linewidth]{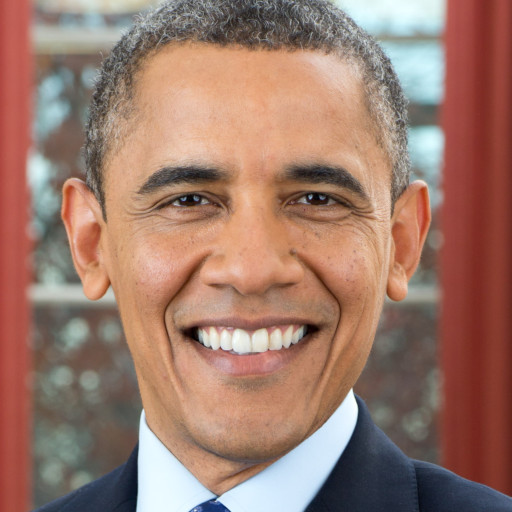}} &
      \multicolumn{1}{c}{\includegraphics[width=.20\linewidth]{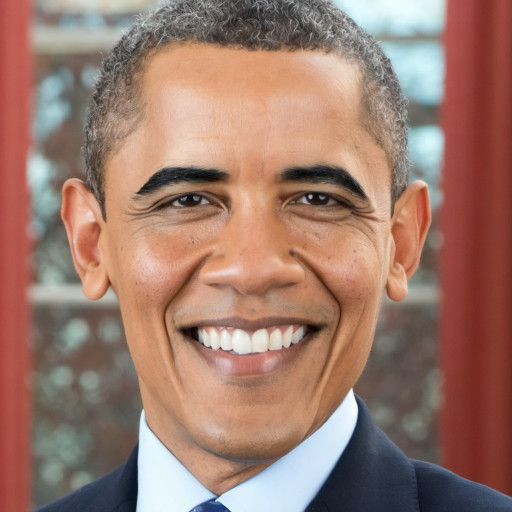}} &
      \multicolumn{1}{c}{\includegraphics[width=.20\linewidth]{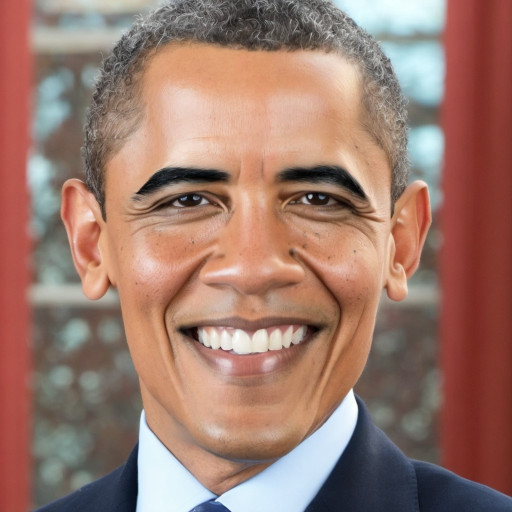}} \\
      \multicolumn{1}{c}{\small -8.0} & \multicolumn{1}{c}{\small -4.0} & \multicolumn{1}{c}{\small $\lambda_{cfg}$} & \multicolumn{1}{c}{\small 4.0} & \multicolumn{1}{c}{\small 8.0} \\
      \midrule
      \multicolumn{1}{c}{\small 0.773} & \multicolumn{1}{c}{\small 0.607} & \multicolumn{1}{c}{\small ID Distance} & \multicolumn{1}{c}{\small 0.170} & \multicolumn{1}{c}{\small 0.229} \\
      \multicolumn{2}{c}{\small $\underleftarrow{\rule[-4pt]{0pt}{2pt}{\text{Increase}}}$} & & \multicolumn{2}{c}{\small $\underrightarrow{\rule[-4pt]{0pt}{2pt}{\text{Increase}}}$} \\
    \end{tabular}
  }
	\caption{We visualize how varying the classifier-free guidance~\cite{ho2022classifier} scale affects identity preservation. Identity distance from the input increases with scale---especially in the negative direction---confirming a divergence from the input identity.}
	\label{fig:cfg-viz}
\end{figure}

While standard personalization approaches use a positive guidance scale to reinforce identity, we instead apply a \emph{negative guidance scale} to steer generation away from the provided identity. Our motivation comes from an observation: as the guidance scale increases, the model tends to enhance and exaggerate facial characteristics. This suggests that negating the scale may attenuate those features. We explore this hypothesis and visualize the results in \cref{fig:cfg-viz}.

\subsection{Attribute-controllable anonymization}

Most anonymization methods do not allow users to choose which facial features to keep or change, limiting an important part of user control. However, research has shown that individuals' privacy preferences are shaped by a balance between perceived risks and potential benefits~\cite{phillips2009disclose}. In some cases, retaining certain personal attributes can serve societal and individual goals---from improving healthcare~\cite{fiske2025weighing} and workplace equity~\cite{browne2003intersection} to advancing research~\cite{acciai2023estimating} and social justice~\cite{edwards2019risk}. To address this, our approach enables intuitive and consistent control over these attributes through natural language prompts. This is possible through our integration of image prompt adapters, which condition the diffusion model on visual identity cues without disrupting its interpretation of textual instructions.

To control facial attributes while anonymizing an image, we first invert the image to its latent noise representation \( x_T \). During reverse diffusion process, we guide the model using a new textual prompt \( \hat{c}_{attr} \) that specifies the desired facial attributes (e.g., ``a young woman'' or ``an elderly man'').

We modify the sampling equation by replacing the original attribute prompt \( c_{attr} \) with the updated prompt \( \hat{c}_{attr} \), while reusing the original noise trajectory \( z_t \). The sampling step becomes:

\begin{equation}
	\hat{x}_{t-1} = \hat{\mu}_t(\hat{x}_t, \hat{x}_{t+1}, c_{id}, \hat{c}_{attr}) + \sigma_t z_t.
\end{equation}

By substituting the previously estimated noise \( z_t \), we obtain:

\begin{equation}
  \begin{aligned}
    \hat{x}_{t-1} & = \hat{\mu}_t(\hat{x}_t, \hat{x}_{t+1}, c_{id}, \hat{c}_{attr}) + x_{t-1} \\ 
                  & - \hat{\mu}_t(x_t, x_{t+1}, {\varnothing}_{id}, c_{attr}).
  \end{aligned}
\end{equation}

This formulation ensures that the anonymized output respects both the conditioning identity and the desired semantic attributes, allowing precise and interpretable control over the anonymization process.


\section{Experiments}

We evaluate our method across benchmarks, comparing it with prior face anonymization approaches in terms of identity removal, attribute preservation, image quality, and controllability. We also present ablation studies to analyze key design choices.

\subsection{Hyperparameter analysis}

\begin{figure}
  \centering
  \begin{subfigure}{0.48\linewidth}
    \includegraphics[width=1.0\linewidth]{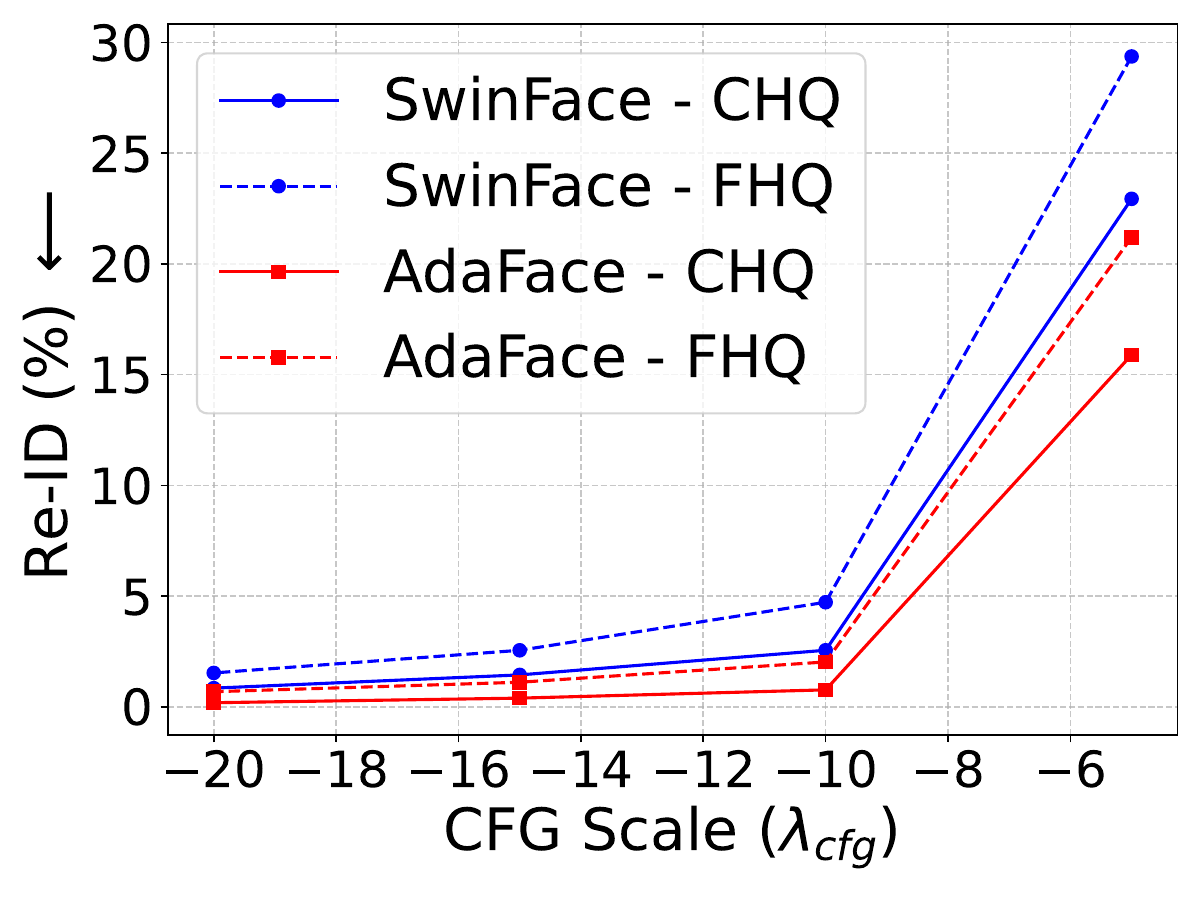}
    \caption{}
    \label{fig:reid-vs-cfg}
  \end{subfigure}
  \hfill
  \begin{subfigure}{0.48\linewidth}
    \includegraphics[width=1.0\linewidth]{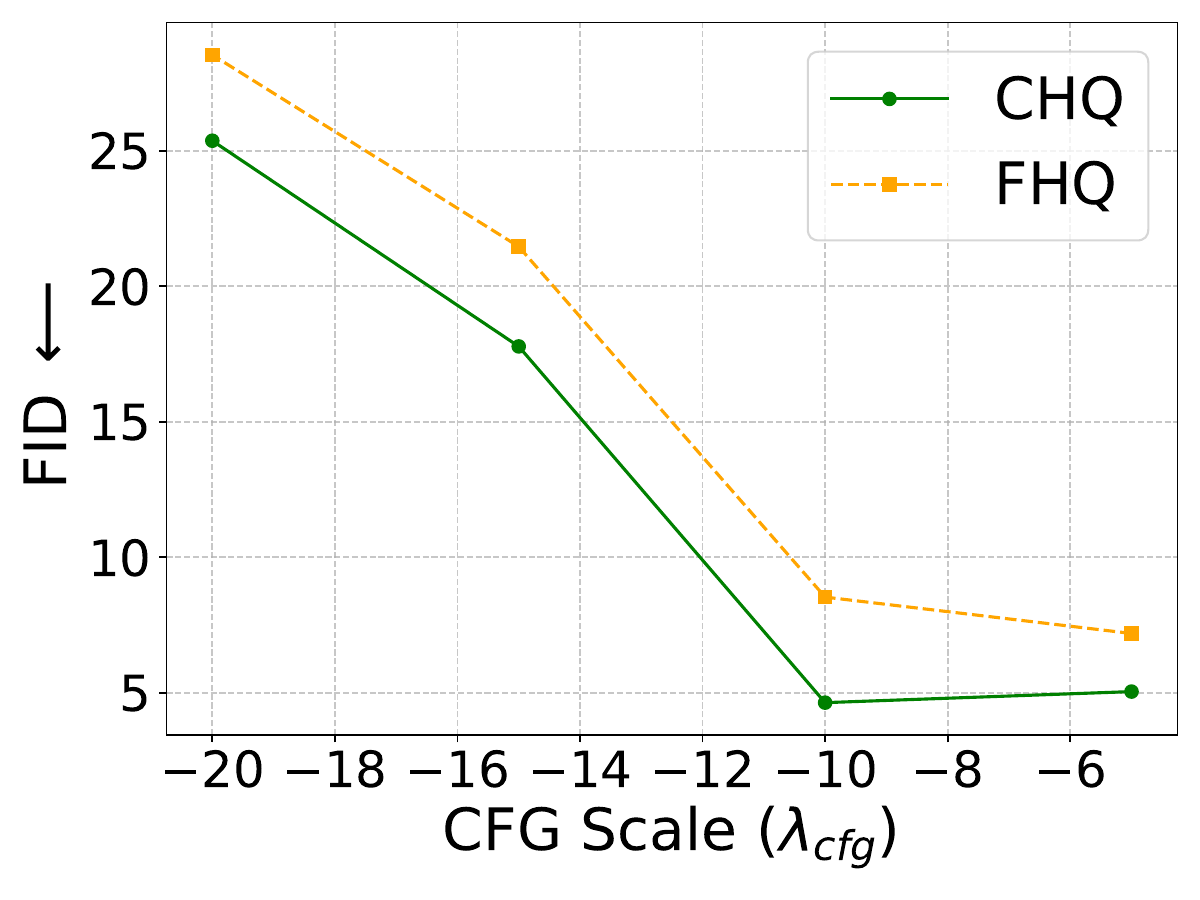}
    \caption{}
    \label{fig:fid-vs-cfg}
  \end{subfigure}
  \\
  \begin{subfigure}{0.48\linewidth}
    \includegraphics[width=1.0\linewidth]{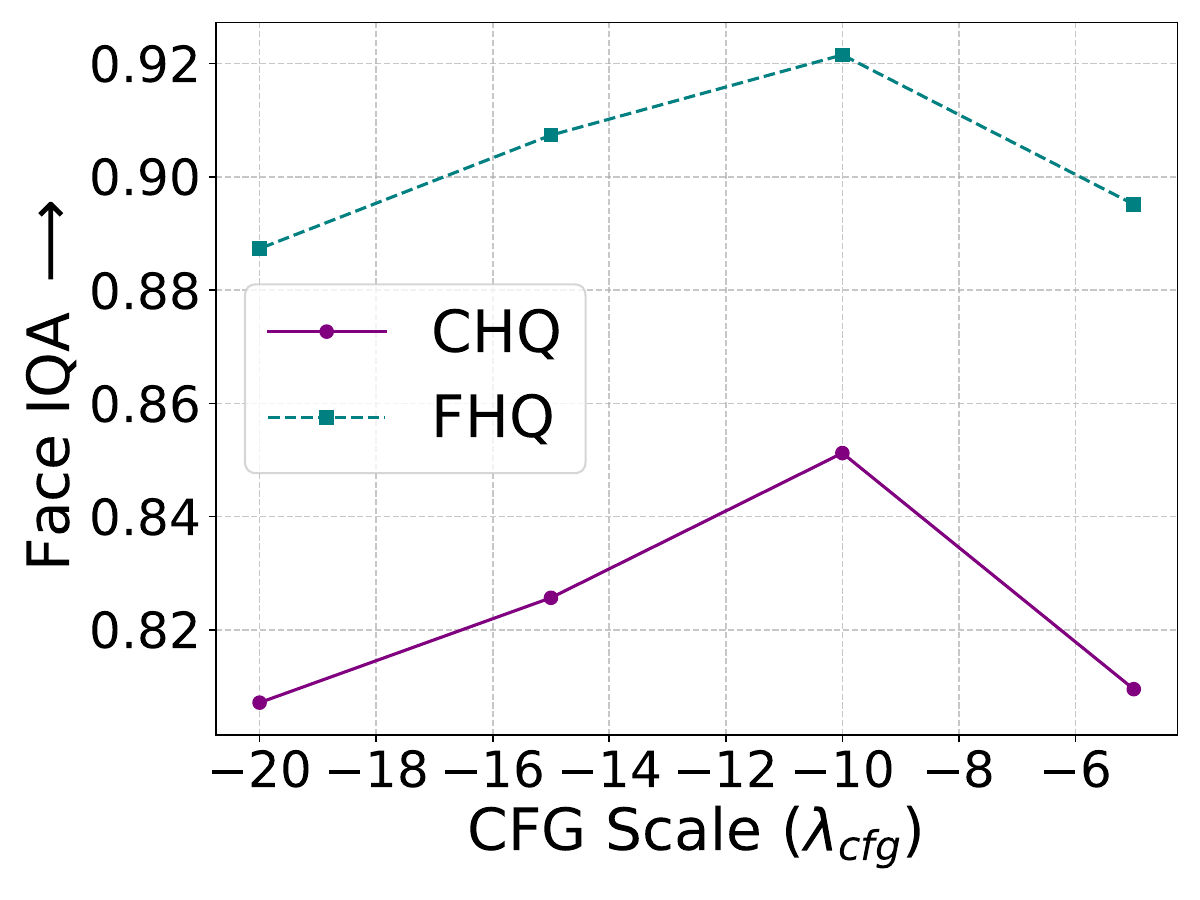}
    \caption{}
    \label{fig:iqa-vs-cfg}
  \end{subfigure}
  \hfill
  \begin{subfigure}{0.48\linewidth}
    \includegraphics[width=1.0\linewidth]{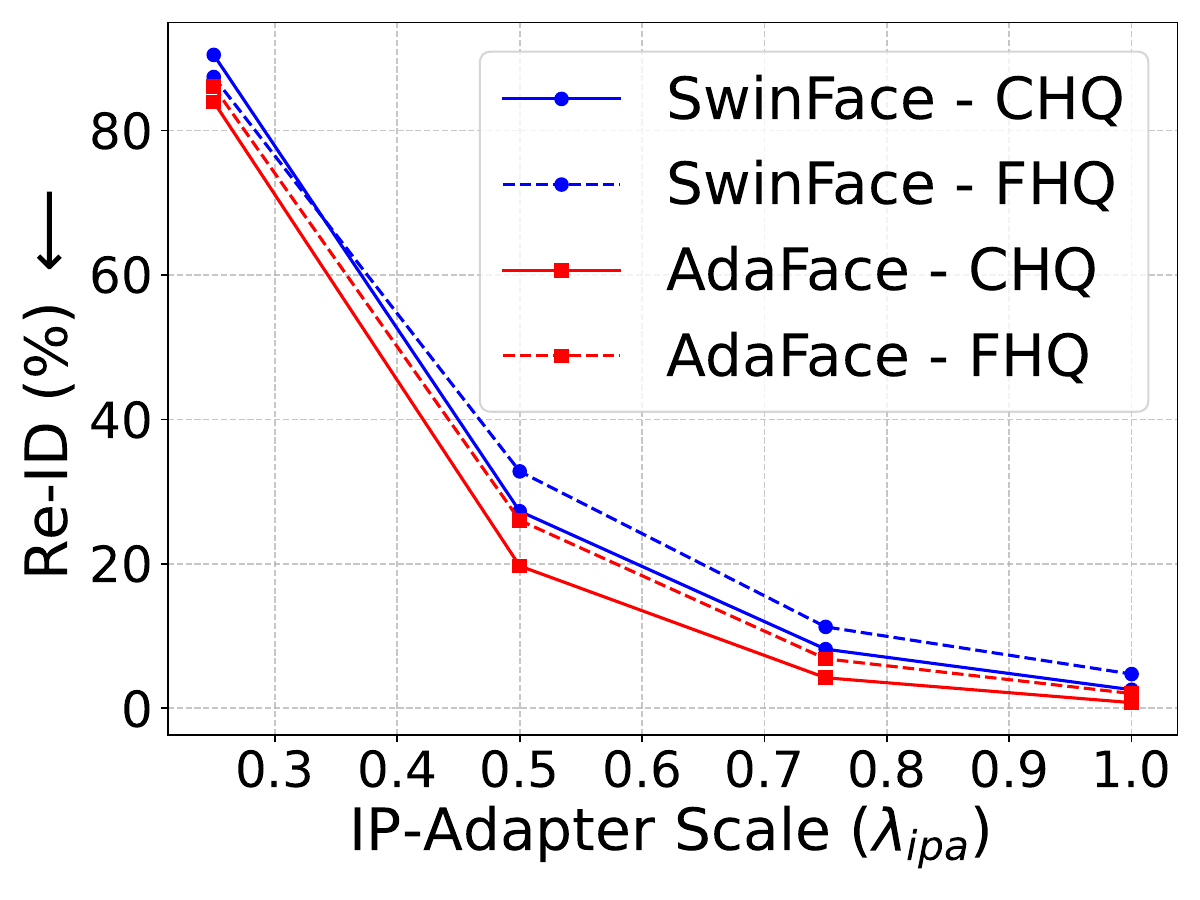}
    \caption{}
    \label{fig:reid-vs-ipadapter}
  \end{subfigure}
  \caption{Hyperparameter analysis of classifier-free guidance~\cite{ho2022classifier} scale and IP-Adapter~\cite{ye2023ip} scale on CHQ (CelebA-HQ~\cite{karras2017progressive}) and FHQ (FFHQ~\cite{karras2019style}).}
  \label{fig:quan-plot}
\end{figure}

\begin{table*}
  \centering
	\resizebox{\linewidth}{!}{
    \begin{tabular}{@{}l cc cc cc cc cc cc cc@{}}
      \toprule
      & \multicolumn{4}{c}{Re-ID (\%) $\downarrow$} & \multicolumn{6}{c}{Attribute preservation $\downarrow$} & \multicolumn{4}{c}{Image quality} \\
      \cmidrule(lr){2-5} \cmidrule(lr){6-11} \cmidrule(lr){12-15}
      & \multicolumn{2}{c}{SwinFace} & \multicolumn{2}{c}{AdaFace} & \multicolumn{2}{c}{Expression} & \multicolumn{2}{c}{Gaze} & \multicolumn{2}{c}{Pose} & \multicolumn{2}{c}{FID $\downarrow$} & \multicolumn{2}{c}{Face IQA $\uparrow$} \\
      \cmidrule(lr){2-3} \cmidrule(lr){4-5} \cmidrule(lr){6-7} \cmidrule(lr){8-9} \cmidrule(lr){10-11} \cmidrule(lr){12-13} \cmidrule(lr){14-15}
      & CHQ & FHQ & CHQ & FHQ & CHQ & FHQ & CHQ & FHQ & CHQ & FHQ & CHQ & FHQ & CHQ & FHQ \\
      \midrule
      Ours & 2.622 & 4.800 & 0.783 & 2.029 & \underline{9.119} & \underline{9.353} & \textbf{0.152} & \underline{0.177} & \textbf{0.050} & \underline{0.052} & \textbf{4.809} & \textbf{8.651} & \underline{0.856} & \textbf{0.921} \\
      NullFace~\cite{kung2025nullface} & \textbf{0.489} & \textbf{0.844} & \textbf{0.157} & \textbf{0.358} & 10.025 & 9.856 & \underline{0.165} & 0.187 & 0.055 & 0.058 & 8.426 & \underline{8.932} & 0.758 & 0.796 \\
      FAMS~\cite{Kung_2025_WACV} & 7.467 & 24.111 & 3.309 & 13.912 & 10.012 & \textbf{8.847} & \underline{0.165} & \textbf{0.176} & \underline{0.054} & \textbf{0.049} & 17.253 & 11.381 & 0.815 & 0.808 \\
      FALCO~\cite{barattin2023attribute} & \underline{1.889} & - & \underline{0.179} & - & 10.206 & - & 0.263 & - & 0.088 & - & 39.501 & - & \textbf{0.875} & - \\
      RiDDLE~\cite{li2023riddle} & - & \underline{2.044} & - & \underline{0.512} & - & 10.049 & - & 0.214 & - & 0.080 & - & 65.141 & - & 0.674 \\
      LDFA~\cite{klemp2023ldfa} & 19.578 & 21.000 & 11.495 & 11.369 & \textbf{8.653} & 10.392 & 0.265 & 0.346 & 0.093 & 0.114 & \underline{8.303} & 10.390 & 0.785 & \underline{0.857} \\
      DP2~\cite{hukkelaas2023deepprivacy2} & 3.889 & 6.644 & 0.901 & 1.927 & 9.931 & 10.141 & 0.263 & 0.295 & 0.163 & 0.163 & 17.544 & 18.809 & 0.599 & 0.629 \\
      \bottomrule
    \end{tabular}
  }
  \caption{Quantitative comparison of facial anonymization methods on CHQ (CelebA-HQ~\cite{karras2017progressive}) and FHQ (FFHQ~\cite{karras2019style}). The best result is marked in \textbf{bold}, and the second-best is \underline{underlined}.}
  \label{tab:baseline-quan-comp}
\end{table*}

\Cref{fig:quan-plot} presents a quantitative analysis of hyperparameters---classifier-free guidance~\cite{ho2022classifier} scale and IP-Adapter~\cite{ye2023ip} scale---evaluated on CelebA-HQ~\cite{karras2017progressive} and FFHQ~\cite{karras2019style} datasets.

\Cref{fig:reid-vs-cfg} illustrates how the classifier-free guidance~\cite{ho2022classifier} scale influences re-identification rates, computed using two state-of-the-art face recognition models: SwinFace~\cite{qin2023swinface} and AdaFace~\cite{kim2022adaface}. When the guidance scale approaches zero, the generated faces closely resemble the original inputs, leading to higher re-identification rates. Conversely, as the guidance scale becomes more negative, the generated faces diverge further from the originals, resulting in lower re-identification rates.

While reducing the guidance scale lowers re-identification rates, excessively negative values degrade image quality. As shown in \cref{fig:fid-vs-cfg,fig:iqa-vs-cfg}, overly negative guidance scales lead to higher Fréchet Inception Distance (FID)~\cite{heusel2017gans} and reduced scores from a face-specific image quality assessment (IQA) model~\cite{chen2024topiq}.

\Cref{fig:reid-vs-ipadapter} examines the impact of the IP-Adapter~\cite{ye2023ip} scale under a fixed negative guidance setting. This parameter controls the strength of identity embeddings. IP-Adapter~\cite{ye2023ip} scale of 0.0 disables identity conditioning, leading to poor anonymization and high re-identification rates. As the scale increases, the model increasingly leverages the identity embeddings, enabling more effective modification of identity-specific features and a drop in re-identification rates.

\subsection{Comparison with existing methods}

We compare our method against six state-of-the-art facial anonymization approaches: NullFace~\cite{kung2025nullface}, FAMS~\cite{Kung_2025_WACV}, FALCO~\cite{barattin2023attribute}, RiDDLE~\cite{li2023riddle}, LDFA~\cite{klemp2023ldfa}, and DP2~\cite{hukkelaas2023deepprivacy2}, using CelebA-HQ~\cite{karras2017progressive} and FFHQ~\cite{karras2019style} datasets. We evaluated 4,500 subjects from each dataset. Our method was implemented on Stable Diffusion XL (SDXL)~\cite{podell2023sdxl} for its superior image quality, with hyperparameters set to an IP-Adapter~\cite{ye2023ip} scale of 1.0 and a classifier-free guidance~\cite{ho2022classifier} scale of –10.0. On an A100 GPU, generating a $1024 \times 1024$ image took approximately 13 seconds.

Evaluation was conducted across three dimensions: identity removal, attribute preservation, and image quality. Identity removal was measured by re-identification rates using SwinFace~\cite{qin2023swinface} and AdaFace~\cite{kim2022adaface}. To avoid bias, a different face recognition model~\cite{deng2019arcface} was used to extract identity embeddings during generation. Attribute preservation was assessed via three metrics: expression difference using a 3D face reconstruction model~\cite{deng2019accurate}, pose difference using a head pose estimation network~\cite{ruiz2018fine}, and gaze difference using a gaze estimation model~\cite{abdelrahman2023l2cs}. For image quality, we adopted the approach from previous studies~\cite{maximov2020ciagan,barattin2023attribute,dall2022graph,helou2023vera} by calculating FID~\cite{heusel2017gans}. Additionally, we utilized a face-specific IQA model~\cite{chen2024topiq}.

As summarized in \cref{tab:baseline-quan-comp}, our method and NullFace~\cite{kung2025nullface} exhibit balanced performance without weaknesses. FAMS~\cite{Kung_2025_WACV} and LDFA~\cite{klemp2023ldfa} suffer from high re-identification rates, indicating poor anonymization. Despite leveraging StyleGAN2~\cite{karras2020analyzing}, known for producing photorealistic images, FALCO~\cite{barattin2023attribute} and RiDDLE~\cite{li2023riddle} exhibit highest FID~\cite{heusel2017gans} scores---due to their inability to preserve background consistency, resulting in generated images that deviate from the originals. DP2~\cite{hukkelaas2023deepprivacy2} struggles with pose preservation and image quality, likely because its inpainting strategy relies on pose estimation, which can be inaccurate, leading to generated faces misaligned with the original orientation. We also include privacy–utility trade-off plots in the supplementary material (\cref{sec:balanced-performance}) to visualize the relationship between identity removal, attribute preservation, and image quality across methods.

While NullFace~\cite{kung2025nullface} achieves the lowest re-identification rates, it underperforms in attribute preservation and image quality compared to our method. NullFace~\cite{kung2025nullface} samples anonymized identities in the embedding space of an identity encoder, whereas our approach operates in the latent space of diffusion models. A drawback of the former is that sampled embeddings may not always correspond to valid human faces, leading to lower Face IQA~\cite{chen2024topiq} scores.

\begin{figure}
	\centering
	\begin{tabularx}{\linewidth}{@{}X@{}X@{}X@{}X@{}X@{}X@{}X@{}}
    \multicolumn{7}{@{}c@{}}{\includegraphics[width=\linewidth]{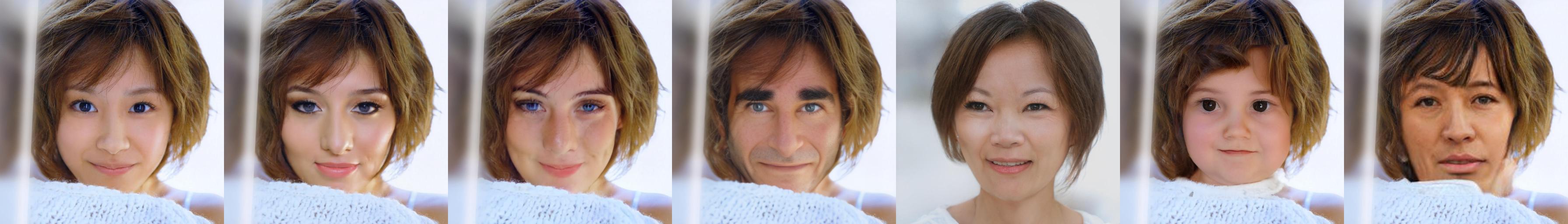}} \\
		\multicolumn{7}{@{}c@{}}{\includegraphics[width=\linewidth]{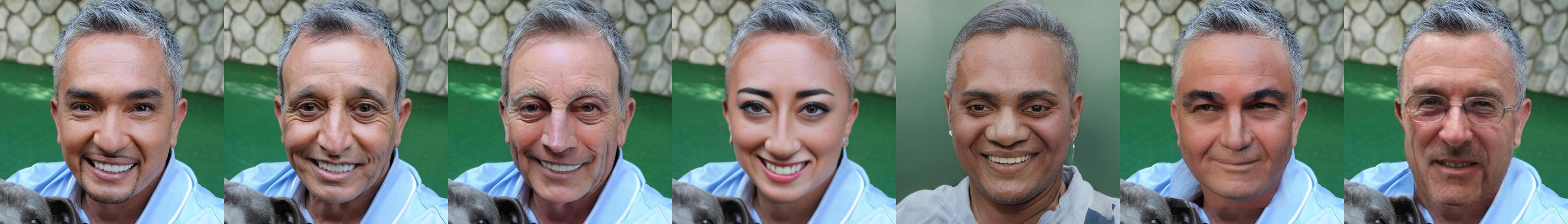}} \\
		\multicolumn{7}{@{}c@{}}{\includegraphics[width=\linewidth]{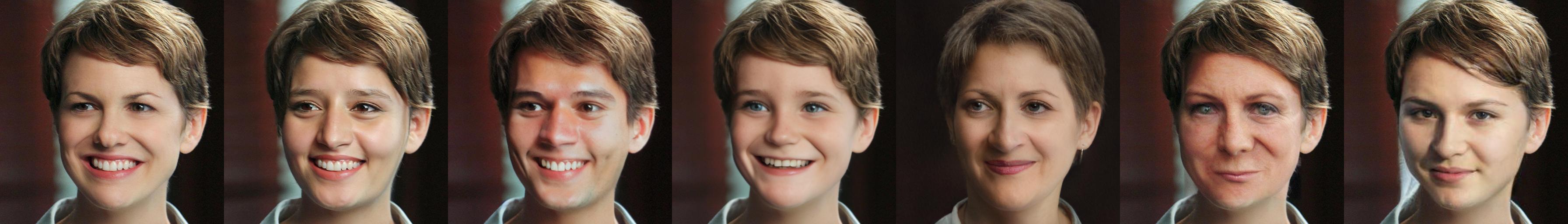}} \\
		\multicolumn{7}{@{}c@{}}{\includegraphics[width=\linewidth]{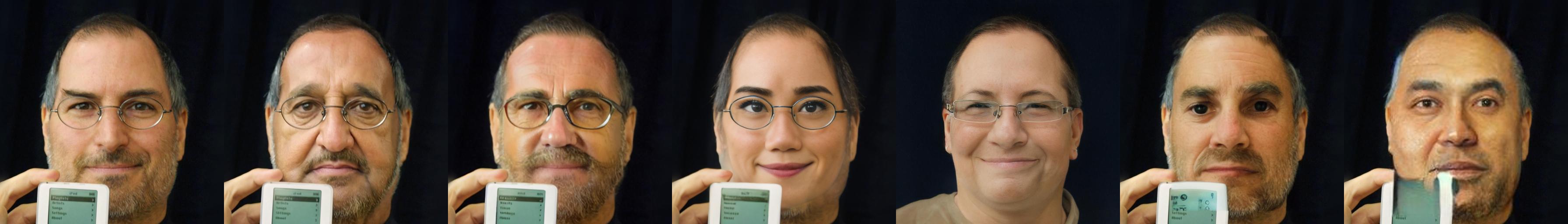}} \\
    \centering \scriptsize Input & \centering \scriptsize Ours & \centering \scriptsize \shortstack{NullFace \\ ~\cite{kung2025nullface}} & \centering \scriptsize \shortstack{FAMS \\ ~\cite{Kung_2025_WACV}} & \centering \scriptsize \shortstack{FALCO \\ ~\cite{barattin2023attribute}} & \centering \scriptsize \shortstack{LDFA \\ ~\cite{klemp2023ldfa}} & \centering \scriptsize \shortstack{DP2 \\ ~\cite{hukkelaas2023deepprivacy2}} \\
	\end{tabularx}
	\caption{Qualitative comparison of anonymization results on CelebA-HQ~\cite{karras2017progressive}.}
	\label{fig:cele-qual-comp}
\end{figure}

\begin{figure}
	\centering
	\begin{tabularx}{\linewidth}{@{}X@{}X@{}X@{}X@{}X@{}X@{}X@{}}
		\multicolumn{7}{@{}c@{}}{\includegraphics[width=\linewidth]{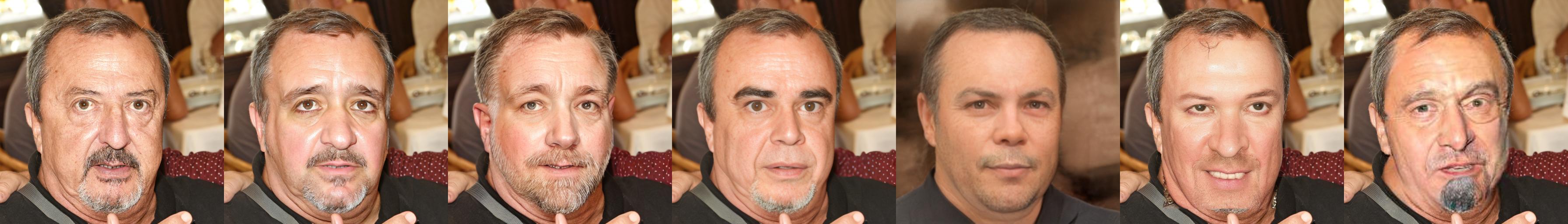}} \\
		\multicolumn{7}{@{}c@{}}{\includegraphics[width=\linewidth]{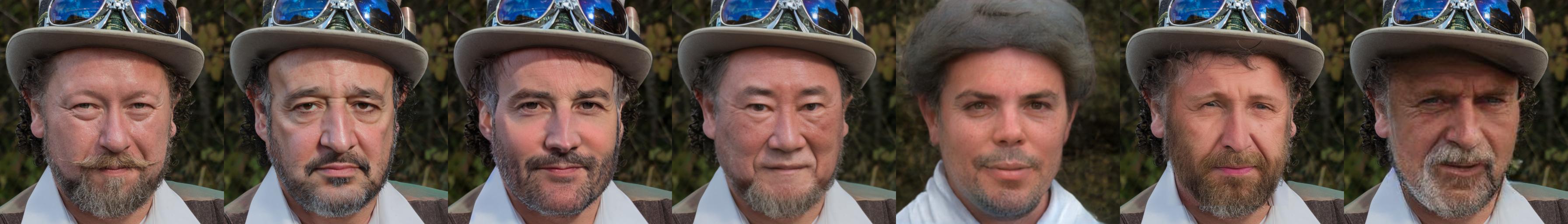}} \\
		\multicolumn{7}{@{}c@{}}{\includegraphics[width=\linewidth]{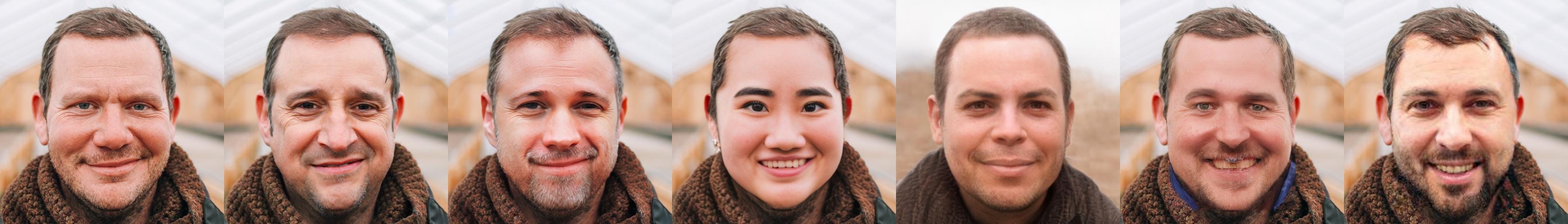}} \\
		\multicolumn{7}{@{}c@{}}{\includegraphics[width=\linewidth]{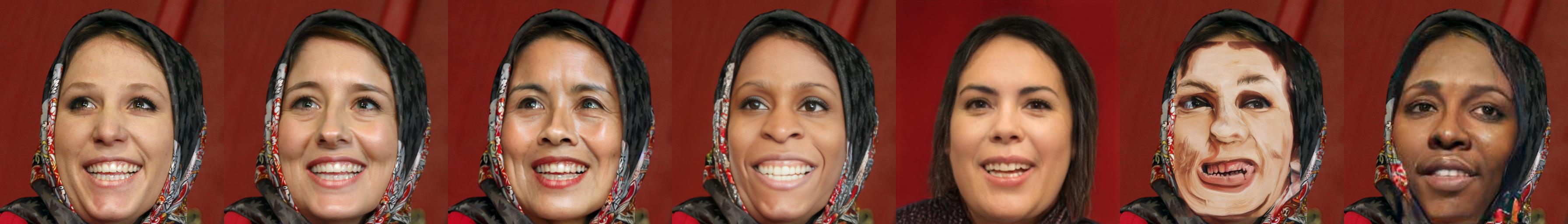}} \\
    \centering \scriptsize Input & \centering \scriptsize Ours & \centering \scriptsize \shortstack{NullFace \\ ~\cite{kung2025nullface}} & \centering \scriptsize \shortstack{FAMS \\ ~\cite{Kung_2025_WACV}} & \centering \scriptsize \shortstack{RiDDLE \\ ~\cite{li2023riddle}} & \centering \scriptsize \shortstack{LDFA \\ ~\cite{klemp2023ldfa}} & \centering \scriptsize \shortstack{DP2 \\ ~\cite{hukkelaas2023deepprivacy2}} \\
	\end{tabularx}
	\caption{Qualitative comparison of anonymization results on FFHQ~\cite{karras2019style}.}
	\label{fig:ffhq-qual-comp}
\end{figure}

\Cref{fig:cele-qual-comp,fig:ffhq-qual-comp} present qualitative comparisons on CelebA-HQ~\cite{karras2017progressive} and FFHQ~\cite{karras2019style}, respectively. Additional examples are provided in the supplementary material (\cref{sec:additional-qualitative}).

\subsection{Attribute-controllable anonymization}

An advantage of our method is its ability to control high-level facial attributes in generated images. To demonstrate this capability, we conduct experiments on controlling three types of basic demographic information---age, sex, and race---so that the generated anonymized faces match those of the original images.

Among existing methods, DP2~\cite{hukkelaas2023deepprivacy2} is the only one that supports attribute-guided anonymization. It does this by adapting StyleMC~\cite{kocasari2022stylemc}, which manipulates images along semantically meaningful directions in the GAN~\cite{goodfellow2014generative} latent space using a CLIP-based~\cite{radford2021learning} loss guided by textual prompts. While enabling attribute control improves DP2~\cite{hukkelaas2023deepprivacy2}'s accuracy in matching target attributes, its overall performance remains below that of our method.

\begin{threeparttable}
  \centering

  {\tablefont
    \begin{adjustbox}{center}
      \begin{tabular}{@{}l cc cc cc@{}}
        \toprule
        & \multicolumn{2}{c}{Age $\downarrow$} & \multicolumn{2}{c}{Sex (\%) $\uparrow$} & \multicolumn{2}{c}{Race (\%) $\uparrow$} \\
        \cmidrule(lr){2-3} \cmidrule(lr){4-5} \cmidrule(lr){6-7}
        & CHQ & FHQ & CHQ & FHQ & CHQ & FHQ \\
        \midrule
        Ours\tnote{*} & \textbf{3.744} & \textbf{4.314} & \textbf{99.485} & \textbf{98.638} & \textbf{87.016} & \underline{76.548} \\
        Ours\tnote{\textdagger} & 4.931 & 5.854 & 92.663 & 82.720 & 60.882 & 49.479 \\
        NullFace~\cite{kung2025nullface} & 5.461 & 6.017 & 87.113 & 77.023 & 62.784 & 56.048 \\
        FAMS~\cite{Kung_2025_WACV} & 6.293 & 6.598 & 29.802 & 58.704 & 32.749 & 44.621 \\
        FALCO~\cite{barattin2023attribute} & 5.016 & - & 89.138 & - & 66.014 & - \\
        RiDDLE~\cite{li2023riddle} & - & 6.144 & - & 73.437 & - & 54.039 \\
        LDFA~\cite{klemp2023ldfa} & 4.426 & 5.888  & 88.713 & 77.401 & 67.572 & 59.238 \\
        DP2~\cite{hukkelaas2023deepprivacy2}\tnote{*} & \underline{3.890} & \underline{4.468} & \underline{98.124} & \underline{96.674} & \underline{82.433} & \textbf{83.033} \\
        DP2~\cite{hukkelaas2023deepprivacy2}\tnote{\textdagger} & 4.822 & 5.627 & 87.523 & 79.750 & 63.951 & 58.865 \\
        \bottomrule
      \end{tabular}
    \end{adjustbox}

    \begin{tablenotes}[para]
      \item[*] w/ attribute control
      \item[\textdagger] w/o attribute control
    \end{tablenotes}
  }
  \caption{Accuracy of preserving high-level facial attributes (age, sex, race) in anonymized faces.}
  \label{tab:baseline-age-gender-race-quan-comp}
\end{threeparttable}

\begin{figure*}
  \centering
  \begin{subfigure}{0.31\linewidth}

		\begin{tabular}{
			*{3}{>{\centering\arraybackslash}m{\dimexpr0.33\linewidth-2\tabcolsep}}
			}
			\small Input & \multicolumn{2}{@{}c@{}}{\small Keep/Change age} \\
			\cmidrule(lr){1-1} \cmidrule(lr){2-3}
			\multicolumn{3}{@{}c@{}}{\includegraphics[width=1.0\linewidth]{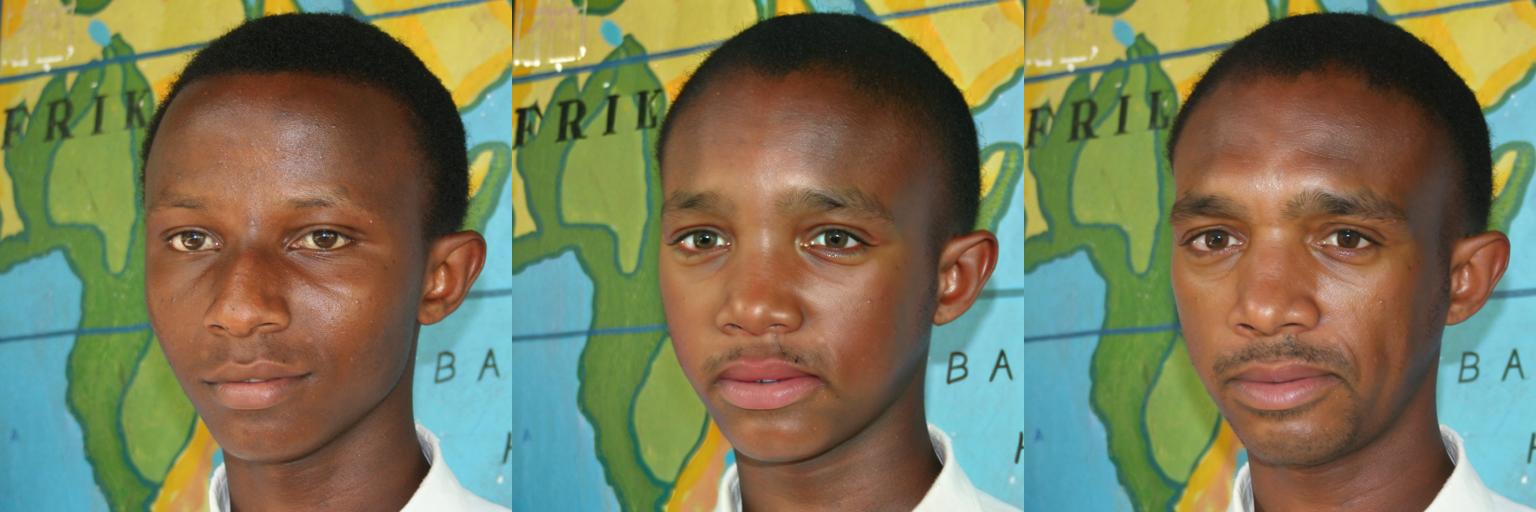}} \\
			\multicolumn{3}{@{}c@{}}{\includegraphics[width=1.0\linewidth]{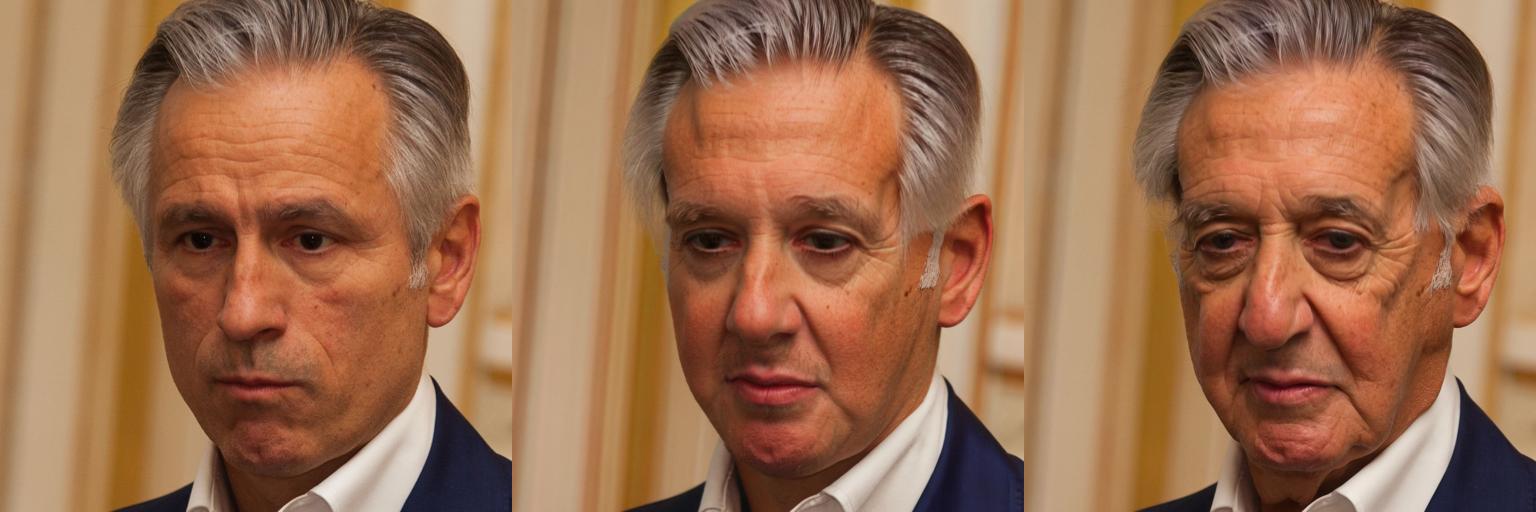}} \\
		\end{tabular}

    \caption{}
  \end{subfigure}
  \quad
  \begin{subfigure}{0.31\linewidth}

		\begin{tabular}{
			*{3}{>{\centering\arraybackslash}m{\dimexpr0.33\linewidth-2\tabcolsep}}
			}
			\small Input & \multicolumn{2}{@{}c@{}}{\small Keep/Change sex} \\
			\cmidrule(lr){1-1} \cmidrule(lr){2-3}
			\multicolumn{3}{@{}c@{}}{\includegraphics[width=1.0\linewidth]{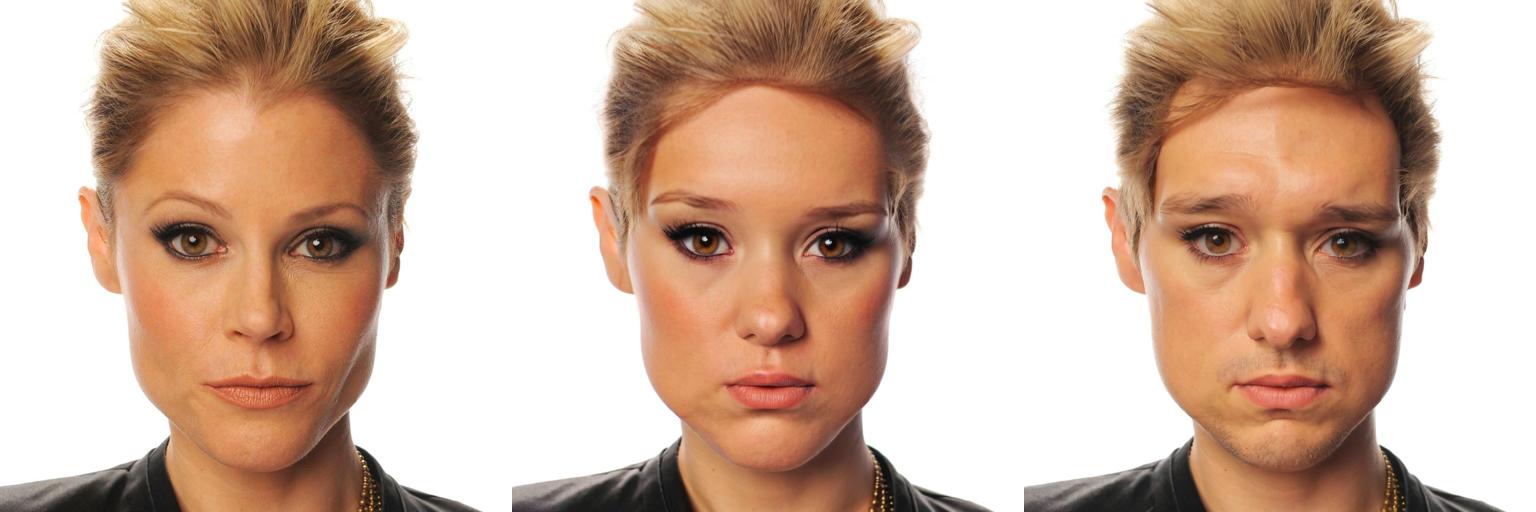}} \\
			\multicolumn{3}{@{}c@{}}{\includegraphics[width=1.0\linewidth]{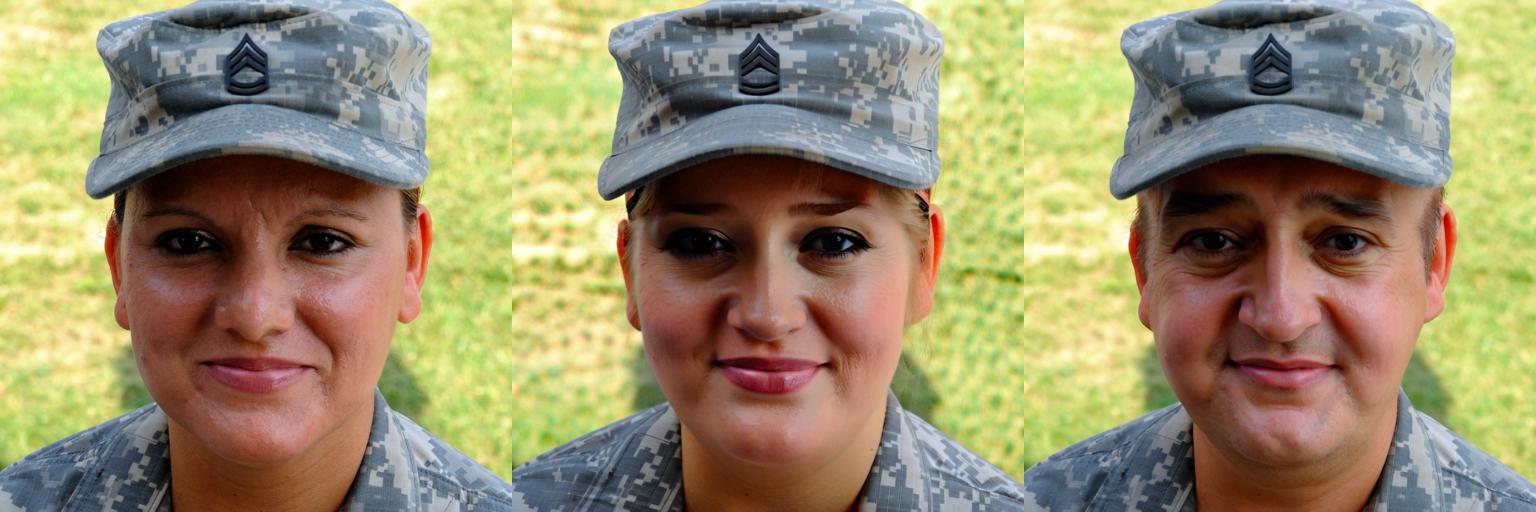}} \\
		\end{tabular}

    \caption{}
  \end{subfigure}
  \quad
  \begin{subfigure}{0.31\linewidth}

		\begin{tabular}{
			*{3}{>{\centering\arraybackslash}m{\dimexpr0.33\linewidth-2\tabcolsep}}
			}
			\small Input & \multicolumn{2}{@{}c@{}}{\small Keep/Change race} \\
			\cmidrule(lr){1-1} \cmidrule(lr){2-3}
			\multicolumn{3}{@{}c@{}}{\includegraphics[width=1.0\linewidth]{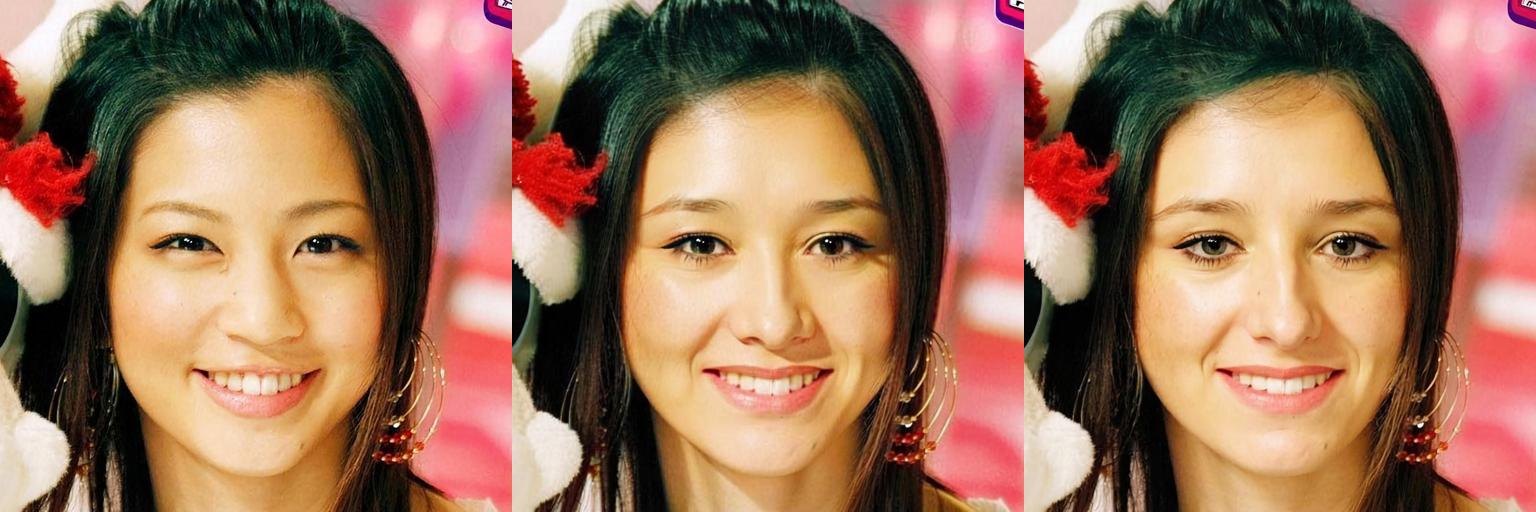}} \\
			\multicolumn{3}{@{}c@{}}{\includegraphics[width=1.0\linewidth]{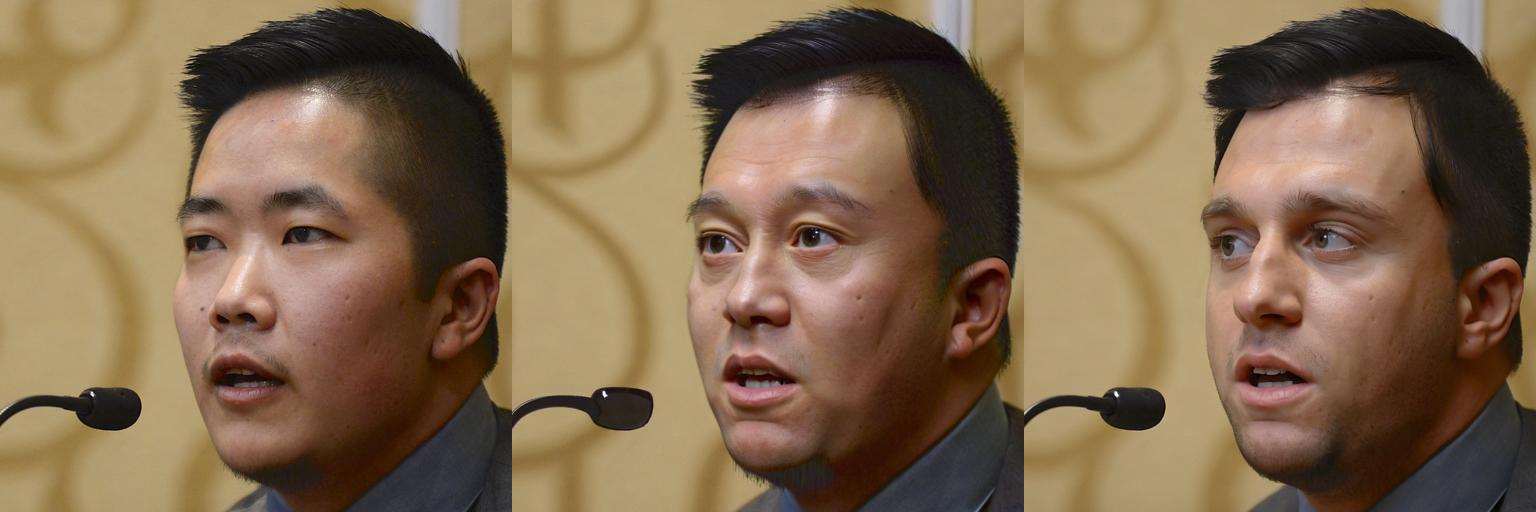}} \\
		\end{tabular}

    \caption{}
  \end{subfigure}
  \caption{Qualitative comparison demonstrating the effect of attribute control on anonymized faces. Examples show the difference between preserving and altering age, sex, and race attributes.}
  \label{fig:age-gender-race-qual}
\end{figure*}

In \cref{tab:baseline-age-gender-race-quan-comp}, our attribute-controlled anonymization achieves the highest accuracy in preserving all three attributes on CelebA-HQ~\cite{karras2017progressive}, and two of the three attributes on FFHQ~\cite{karras2019style}. For the race attribute on FFHQ~\cite{karras2019style}, DP2~\cite{hukkelaas2023deepprivacy2} with attribute control slightly outperforms our method. \Cref{fig:age-gender-race-qual} presents qualitative examples that illustrate the effectiveness of attribute control, highlighting differences between preserving and altering these facial features.

\subsection{Ablation study}

\begin{table*}
  \centering
	\resizebox{\linewidth}{!}{
    \begin{tabular}{@{}l cc cc cc cc cc cc cc@{}}
      \toprule
      & \multicolumn{4}{c}{Re-ID (\%) $\downarrow$} & \multicolumn{6}{c}{Attribute preservation $\downarrow$} & \multicolumn{4}{c}{Image quality} \\
      \cmidrule(lr){2-5} \cmidrule(lr){6-11} \cmidrule(lr){12-15}
      & \multicolumn{2}{c}{SwinFace} & \multicolumn{2}{c}{AdaFace} & \multicolumn{2}{c}{Expression} & \multicolumn{2}{c}{Gaze} & \multicolumn{2}{c}{Pose} & \multicolumn{2}{c}{FID $\downarrow$} & \multicolumn{2}{c}{Face IQA $\uparrow$} \\
      \cmidrule(lr){2-3} \cmidrule(lr){4-5} \cmidrule(lr){6-7} \cmidrule(lr){8-9} \cmidrule(lr){10-11} \cmidrule(lr){12-13} \cmidrule(lr){14-15}
      & CHQ & FHQ & CHQ & FHQ & CHQ & FHQ & CHQ & FHQ & CHQ & FHQ & CHQ & FHQ & CHQ & FHQ \\
      \midrule
      Ours & \underline{2.622} & \underline{4.800} & \underline{0.783} & \underline{2.029} & \textbf{9.119} & \textbf{9.353} & \textbf{0.152} & \textbf{0.177} & \textbf{0.050} & \textbf{0.052} & \textbf{4.809} & \textbf{8.651} & \underline{0.856} & \underline{0.921} \\
      DDIM~\cite{song2020denoising} & 36.889 & 39.533 & 36.206 & 37.698 & \underline{9.319} & \underline{10.275} & 0.454 & 0.458 & 0.154 & 0.171 & 47.194 & 38.873 & 0.708 & 0.733 \\
      InstantID~\cite{wang2024instantid} & \textbf{0.489} & \textbf{0.911} & \textbf{0.202} & \textbf{0.781} & 11.532 & 11.802 & \underline{0.185} & \underline{0.204} & \underline{0.057} & \underline{0.054} & \underline{30.334} & \underline{18.347} & \textbf{0.889} & \textbf{0.947} \\
      \bottomrule
    \end{tabular}
  }
  \caption{Ablation study comparing our method with alternatives using DDIM inversion~\cite{song2020denoising} and InstantID~\cite{wang2024instantid} as the generation backbone.}
  \label{tab:ablation-quan-comp}
\end{table*}

To assess key design choices, we conducted an ablation study with two alternative implementations. First, we replaced our DDPM~\cite{ho2020denoising} inversion with DDIM~\cite{song2020denoising}. Second, we substituted the SDXL~\cite{podell2023sdxl} model with InstantID~\cite{wang2024instantid}, a generation model designed for identity conditioning.

Quantitative results are in \cref{tab:ablation-quan-comp}. The DDIM inversion~\cite{song2020denoising} underperforms across all metrics. This is primarily due to DDIM's inability to reconstruct the original image~\cite{mokady2023null,huberman2024edit}, leading to degraded attribute preservation and image quality. Furthermore, DDIM~\cite{song2020denoising} exhibits poor performance to large adjustments in the classifier-free guidance~\cite{ho2022classifier} scale, where excessive changes introduce artifacts and cause image failures. Consequently, this implementation suffers from higher re-identification rates due to its limited capacity to support aggressive anonymization.

The InstantID-based implementation achieves lower re-identification rates than SDXL~\cite{podell2023sdxl}. However, this results in reduced attribute preservation and higher FID~\cite{heusel2017gans}. This degradation is due to the ID embedding in InstantID, which entangles identity features with other facial attributes~\cite{wang2024instantid}, making it difficult to preserve details unrelated to identity.

\begin{figure}
	\centering
	\begin{tabularx}{.75\linewidth}{@{}X@{}X@{}X@{}X@{}}
    \multicolumn{4}{@{}c@{}}{\includegraphics[width=.75\linewidth]{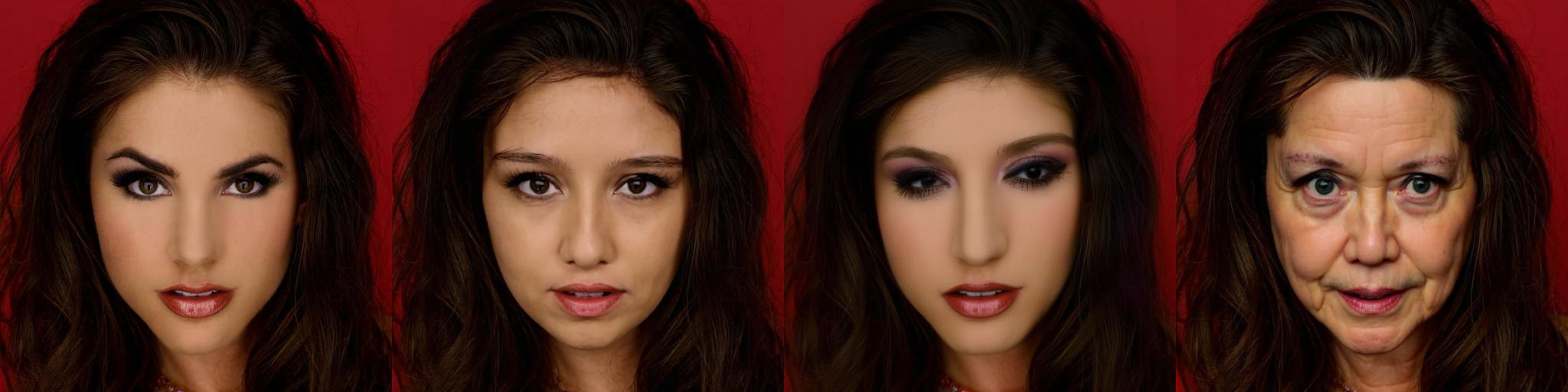}} \\
		\multicolumn{4}{@{}c@{}}{\includegraphics[width=.75\linewidth]{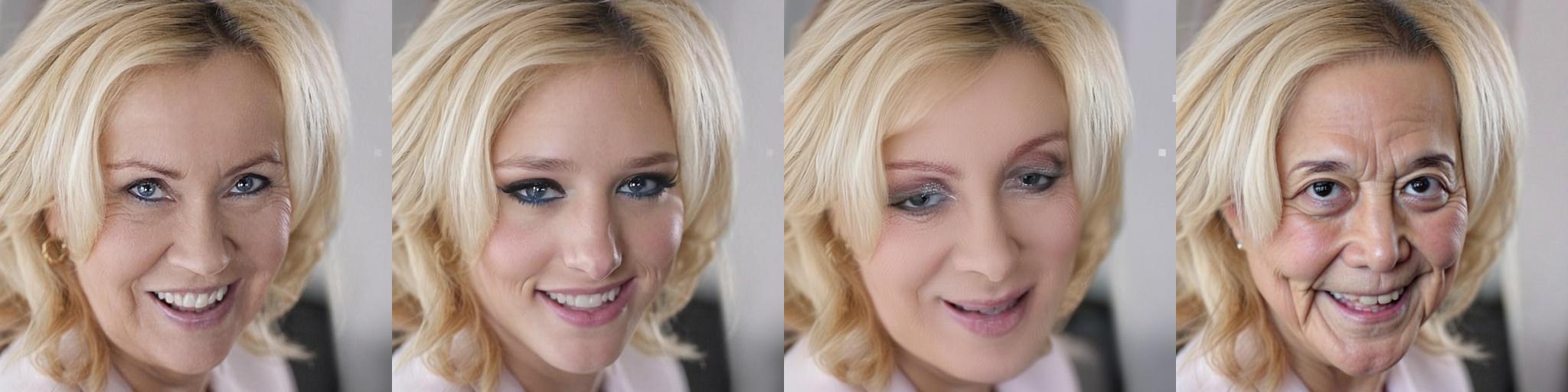}} \\
		\multicolumn{4}{@{}c@{}}{\includegraphics[width=.75\linewidth]{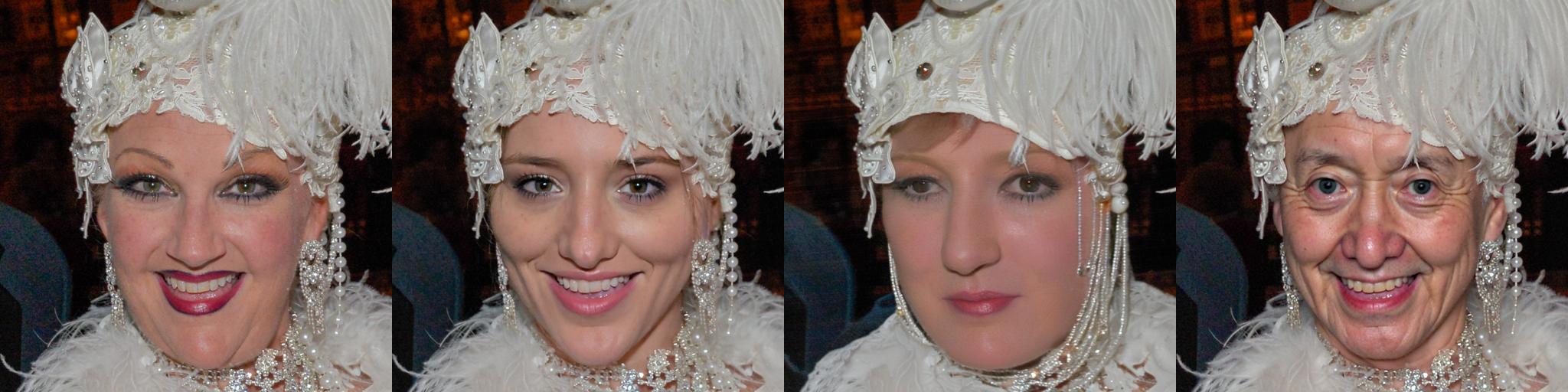}} \\
		\multicolumn{4}{@{}c@{}}{\includegraphics[width=.75\linewidth]{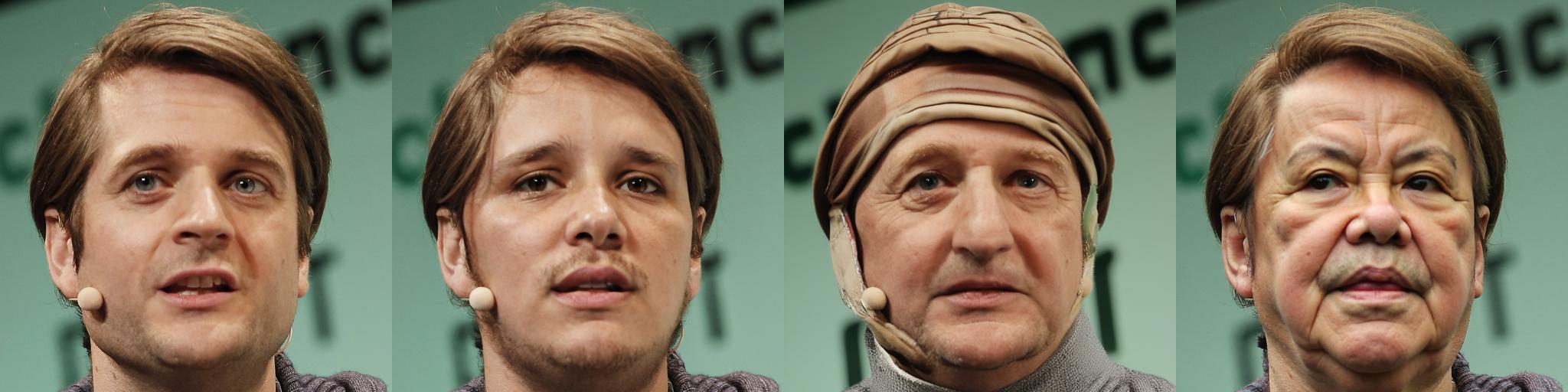}} \\
    \small \centering Input & \small \centering Ours & \small \centering \shortstack{DDIM \\ ~\cite{song2020denoising}} & \small \centering \shortstack{InstantID \\ ~\cite{wang2024instantid}} \\
	\end{tabularx}
	\caption{Qualitative results from the ablation study. Our method maintains both realism and accurate attribute preservation.}
	\label{fig:ablation-qual-comp}
\end{figure}

Qualitative comparisons in \cref{fig:ablation-qual-comp} further support these findings. The DDIM-based model fails to preserve facial attributes such as expression, gaze, and pose, and introduces artifacts. Meanwhile, the InstantID-based model tends to produce facial outputs appearing oversaturated and lacking realism.

We also evaluated the attribute control of these alternative implementations, results in \cref{tab:ablation-age-gender-race-quan-comp}. Neither approach matches our SDXL-based implementation in preserving high-level facial attributes.

\begin{threeparttable}
  \centering

  {\scriptsize
    \begin{adjustbox}{center}
      \begin{tabular}{@{}l cc cc cc@{}}
        \toprule
        & \multicolumn{2}{c}{Age $\downarrow$} & \multicolumn{2}{c}{Sex (\%) $\uparrow$} & \multicolumn{2}{c}{Race (\%) $\uparrow$} \\
        \cmidrule(lr){2-3} \cmidrule(lr){4-5} \cmidrule(lr){6-7}
        & CHQ & FHQ & CHQ & FHQ & CHQ & FHQ \\
        \midrule
        Ours\tnote{*} & \textbf{3.744} & \textbf{4.314} & \textbf{99.485} & \textbf{98.638} & \textbf{87.016} & \textbf{76.548} \\
        Ours\tnote{\textdagger} & 4.931 & 5.854 & \underline{92.663} & \underline{82.720} & 60.882 & 49.479 \\
        DDIM~\cite{song2020denoising} & \underline{4.257} & \underline{5.413} & 78.701 & 71.016 & \underline{62.285} & \underline{55.889} \\
        InstantID~\cite{wang2024instantid} & 16.009 & 15.160 & 83.038 & 78.038 & 19.600 & 28.517 \\
        \bottomrule
      \end{tabular}
    \end{adjustbox}

    \begin{tablenotes}[para]
      \item[*] w/ attribute control
      \item[\textdagger] w/o attribute control
    \end{tablenotes}
  }
  \caption{Attribute control accuracy (age, sex, race) for alternative implementations. Neither DDIM inversion~\cite{song2020denoising} nor InstantID~\cite{wang2024instantid} achieves attribute consistency comparable to our SDXL-based method.}
  \label{tab:ablation-age-gender-race-quan-comp}
\end{threeparttable}

\section{Conclusion}

We proposed a reverse personalization framework for face anonymization that removes identity-specific features while preserving non-identity attributes. Our approach does not require the diffusion model to have prior exposure to the subject or any model fine-tuning, making it broadly applicable to arbitrary individuals. In addition to offering flexible  anonymization, our method strikes an optimal balance among identity obfuscation, attribute preservation, and image quality.

\begin{ack}
This work was supported by the EU Horizon project ``ELIAS - European Lighthouse of AI for Sustainability'' (No. 101120237) and the FIS project GUIDANCE (Debugging Computer Vision Models via Controlled Cross-modal Generation) (No. FIS2023-03251). We acknowledge ISCRA for awarding this project access to the LEONARDO supercomputer, owned by the EuroHPC Joint Undertaking, hosted by CINECA (Italy), and thank the Finnish Foundation for Technology Promotion.
\end{ack}

{
    \small
    \bibliographystyle{ieeenat_fullname}
    \bibliography{main}
}

\clearpage
\setcounter{page}{1}
\maketitlesupplementary

\section{Balanced performance across privacy and utility}
\label{sec:balanced-performance}

\begin{figure}[b]
  \centering
  \begin{subfigure}{0.48\linewidth}
    \includegraphics[width=1.0\linewidth]{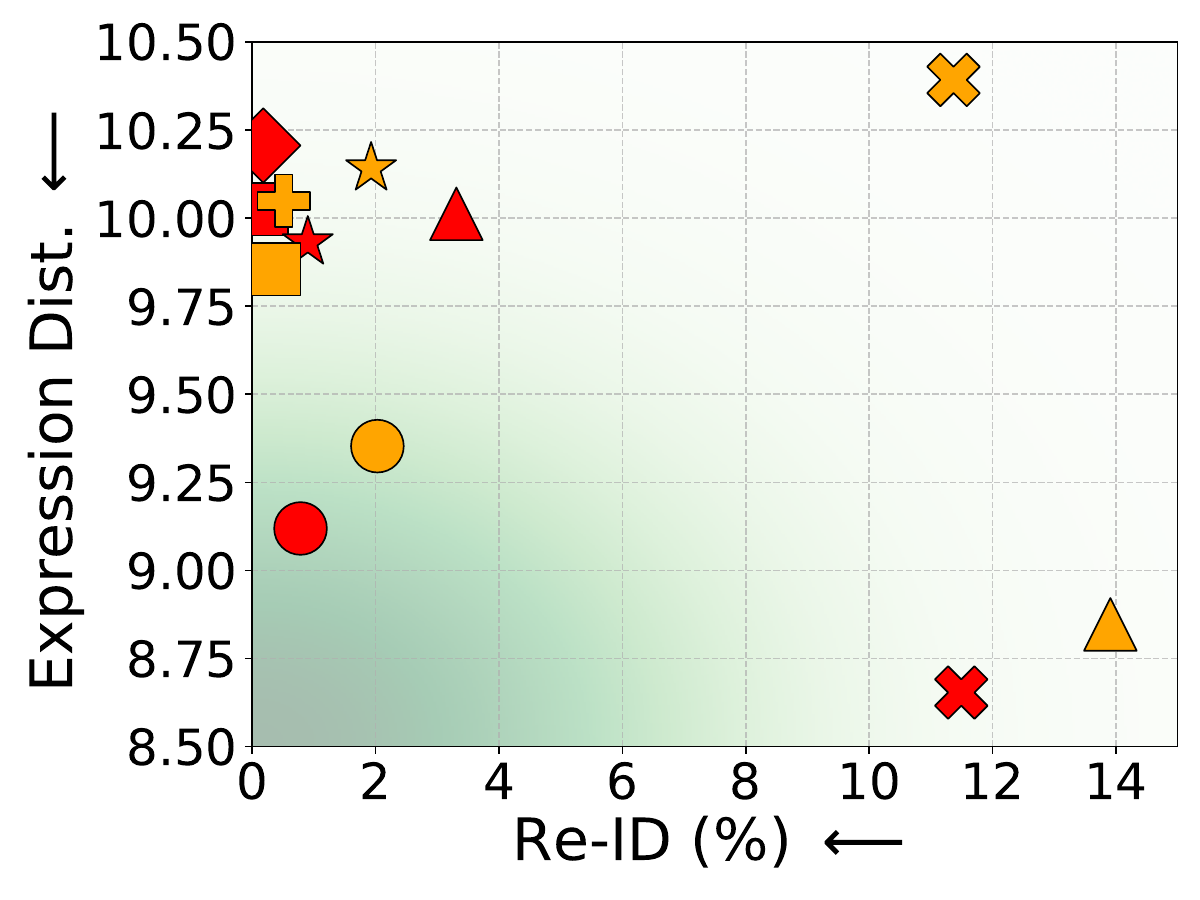}
    \caption{}
    \label{fig:reid_vs_expression_baselines_adaface}
  \end{subfigure}
  \hfill
  \begin{subfigure}{0.48\linewidth}
    \includegraphics[width=1.0\linewidth]{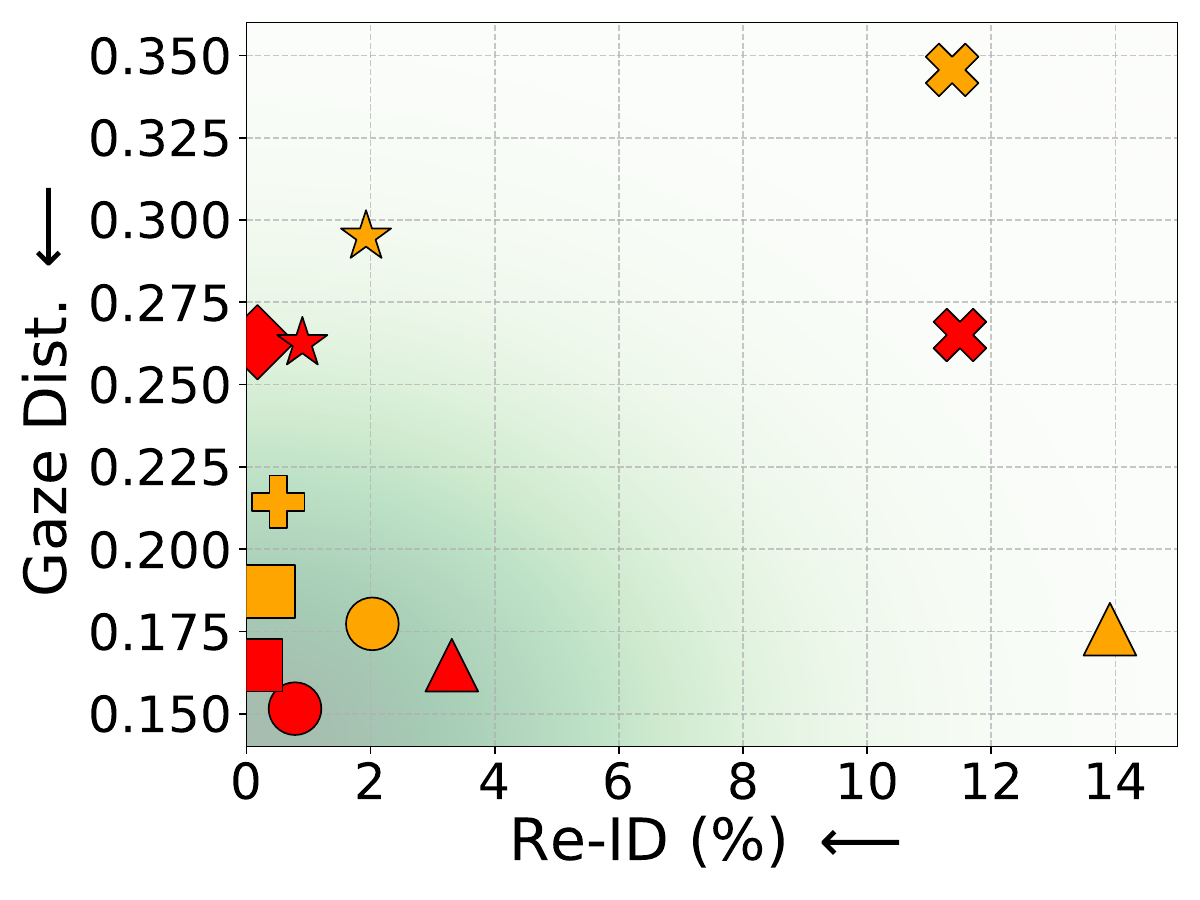}
    \caption{}
    \label{fig:reid_vs_gaze_baselines_adaface}
  \end{subfigure}
  \\
  \begin{subfigure}{0.48\linewidth}
    \includegraphics[width=1.0\linewidth]{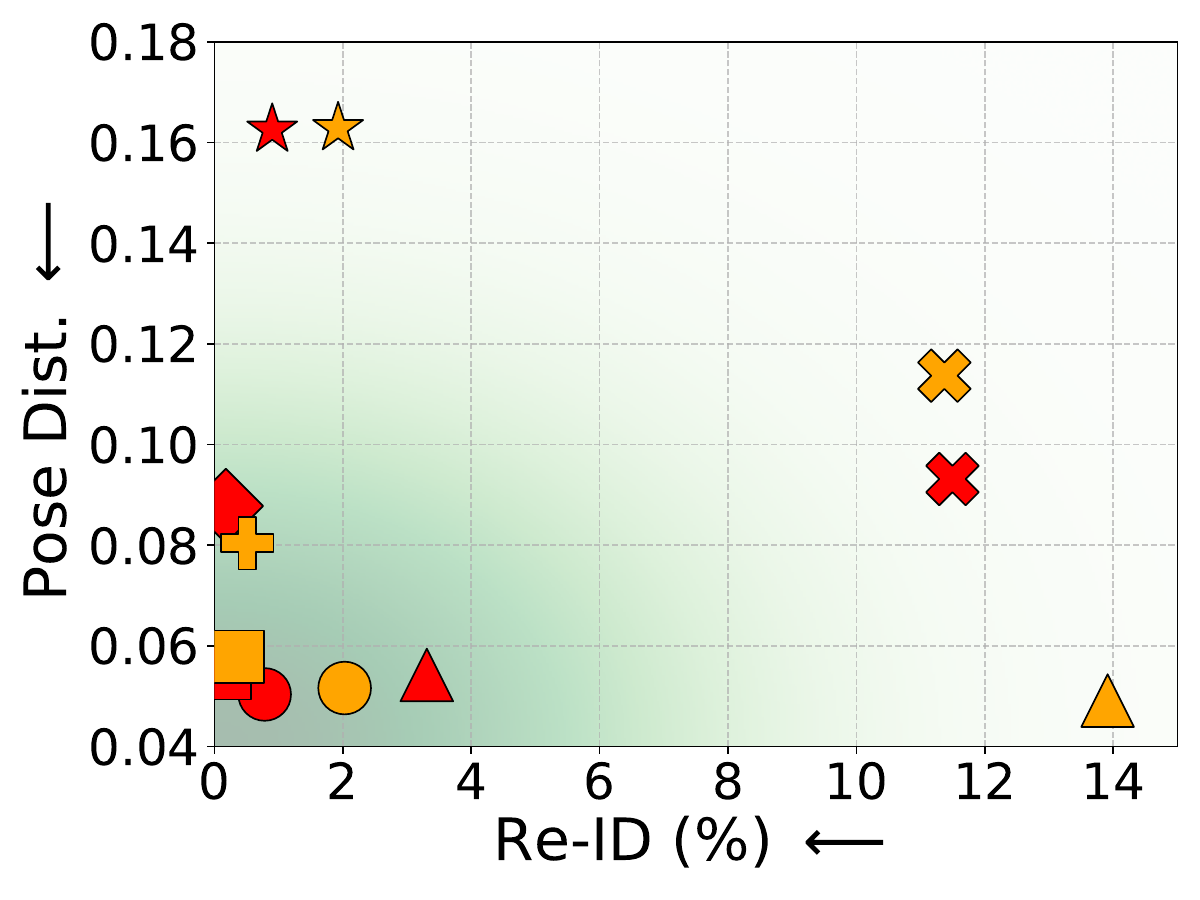}
    \caption{}
    \label{fig:reid_vs_pose_baselines_adaface}
  \end{subfigure}
  \hfill
  \begin{subfigure}{0.48\linewidth}
    \includegraphics[width=1.0\linewidth]{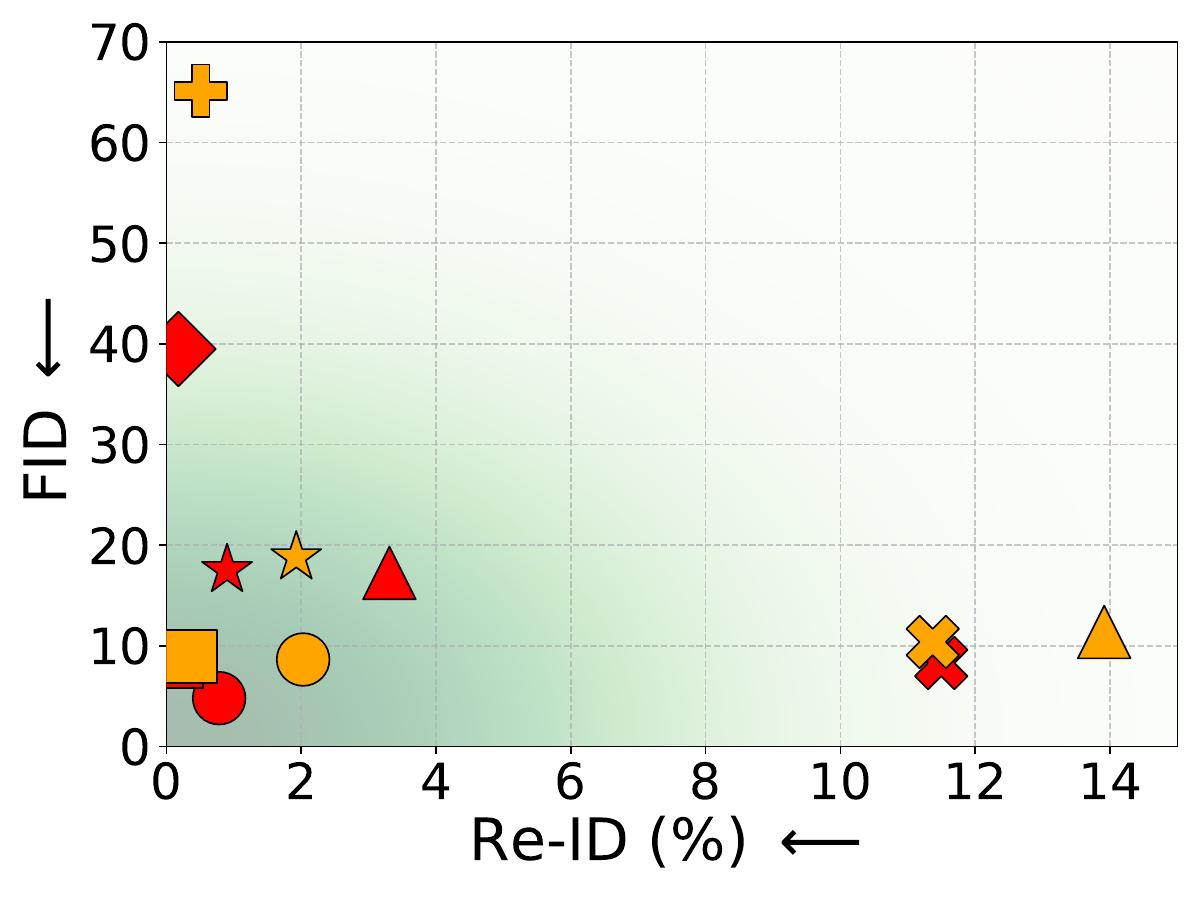}
    \caption{}
    \label{fig:reid_vs_fid_baselines_adaface}
  \end{subfigure}
  \\
  \begin{subfigure}{0.48\linewidth}
    \includegraphics[width=1.0\linewidth]{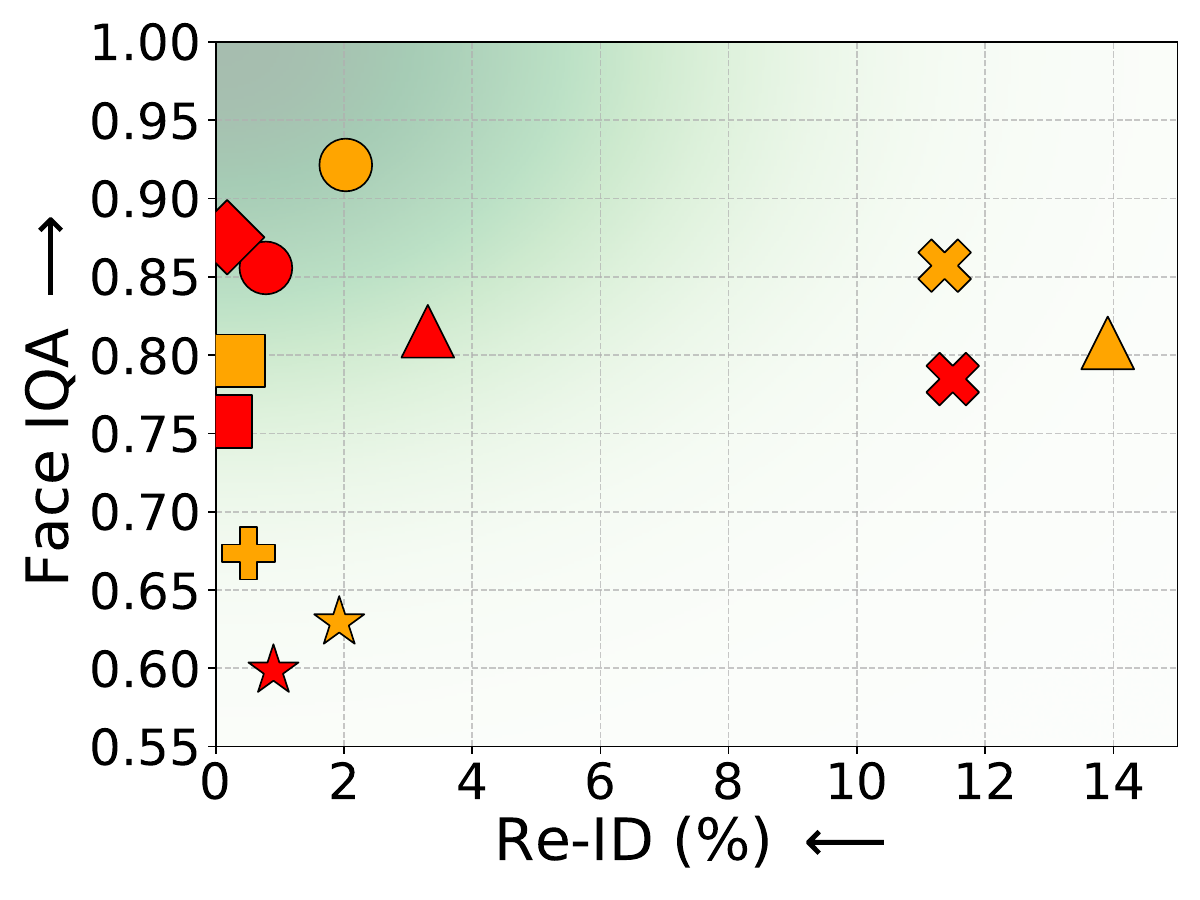}
    \caption{}
    \label{fig:reid_vs_faceiqa_baselines_adaface}
  \end{subfigure}
  \hfill
  \begin{subfigure}{0.48\linewidth}
    \includegraphics[width=1.0\linewidth]{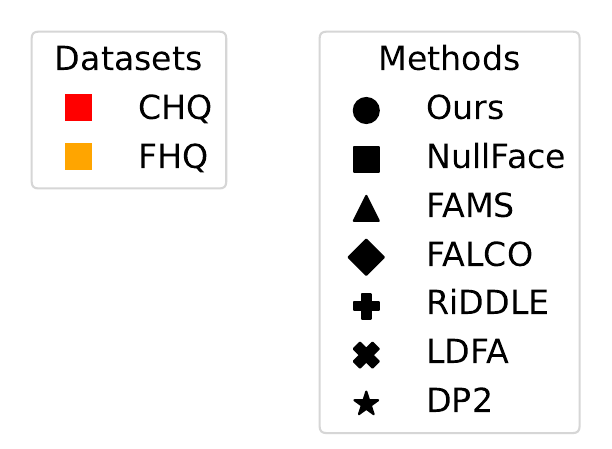}
    \caption{}
    \label{fig:baselines_legends}
  \end{subfigure}
  \caption{Privacy–utility trade-off plots. The gradient background highlights the optimal region (darker green) for balancing privacy and utility.}
  \label{fig:privacy–utility-baselines}
\end{figure}

In \cref{fig:privacy–utility-baselines}, we show five privacy-utility trade-off plots on CelebA-HQ~\cite{karras2017progressive} (CHQ) and FFHQ~\cite{karras2019style} (FHQ). The x-axis shows the re-identification rate, computed using AdaFace~\cite{kim2022adaface}. The y-axis shows utility, with lower values indicating better performance for expression, gaze, pose, and  FID~\cite{heusel2017gans}, and higher values for Face IQA~\cite{chen2024topiq}. Each method (NullFace~\cite{kung2025nullface}, FAMS~\cite{Kung_2025_WACV}, FALCO~\cite{barattin2023attribute}, RiDDLE~\cite{li2023riddle}, LDFA~\cite{klemp2023ldfa}, DP2~\cite{hukkelaas2023deepprivacy2}) appears as a point. A gradient background highlights the lower-left (or upper-left for Face IQA~\cite{chen2024topiq}) corner as the  optimal balance between privacy and utility. This visualization shows that our method consistently lies closest to the ``sweet spot,'' while others compromise either privacy or utility.

\section{Privacy-utility trade-off across guidance scales}
\label{sec:privacy–utility}

\begin{figure}[b]
  \centering
  \begin{subfigure}{0.48\linewidth}
    \includegraphics[width=1.0\linewidth]{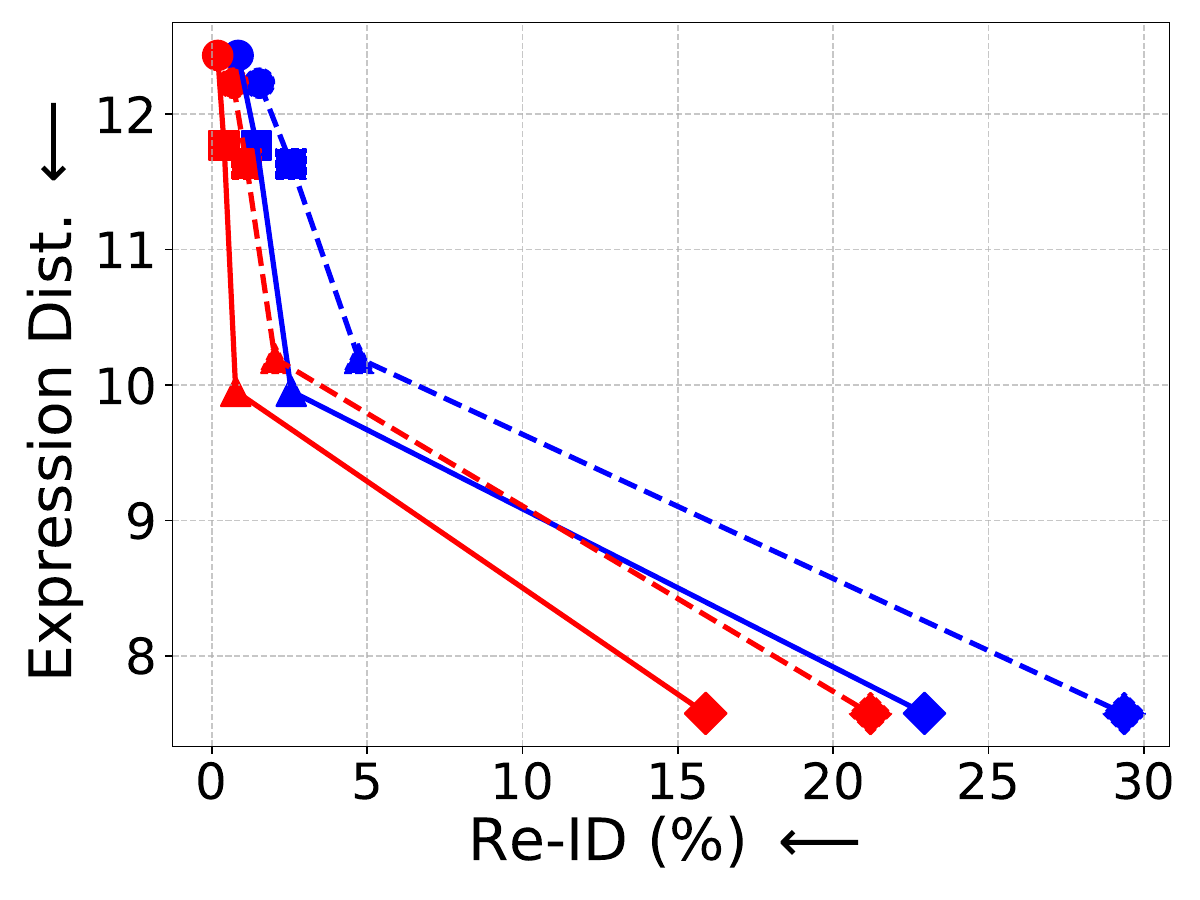}
    \caption{}
    \label{fig:reid_vs_expression}
  \end{subfigure}
  \hfill
  \begin{subfigure}{0.48\linewidth}
    \includegraphics[width=1.0\linewidth]{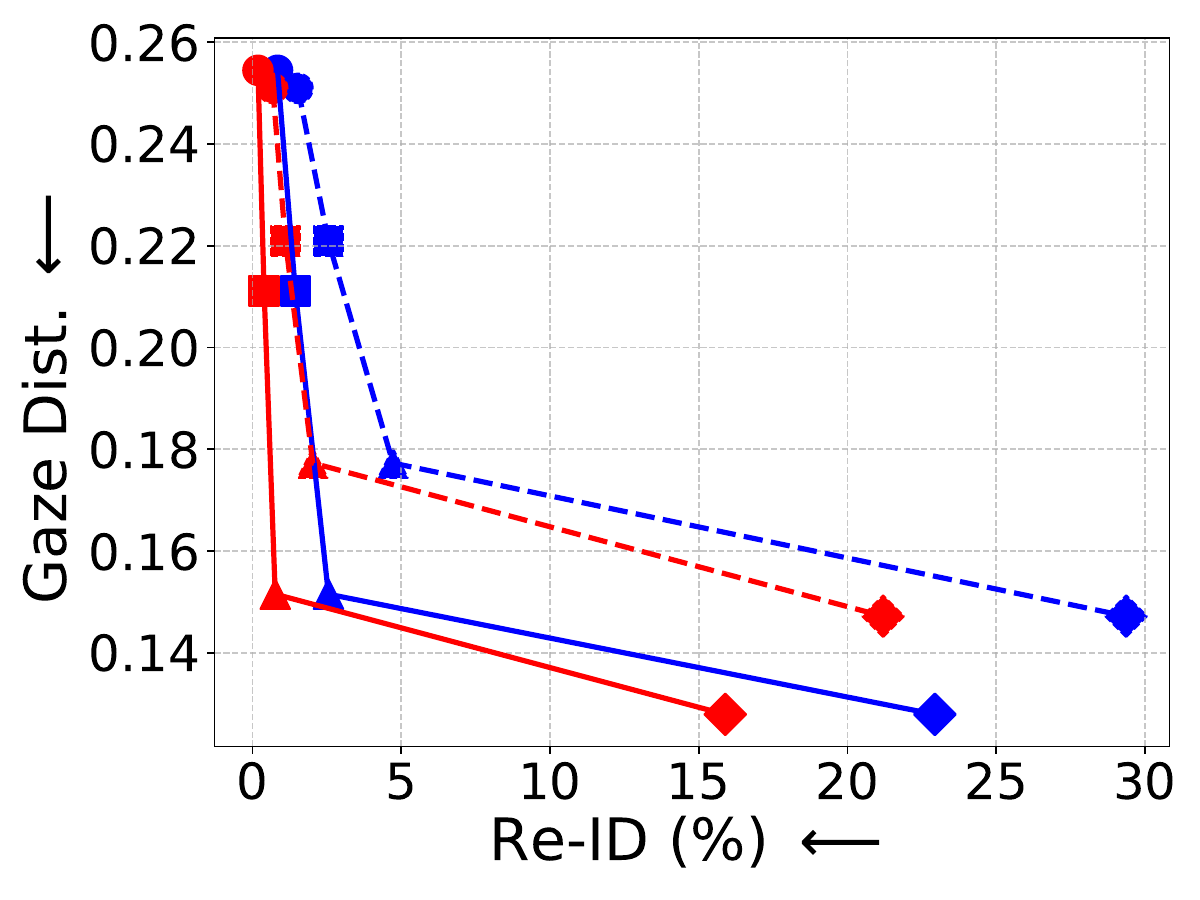}
    \caption{}
    \label{fig:reid_vs_gaze}
  \end{subfigure}
  \\
  \begin{subfigure}{0.48\linewidth}
    \includegraphics[width=1.0\linewidth]{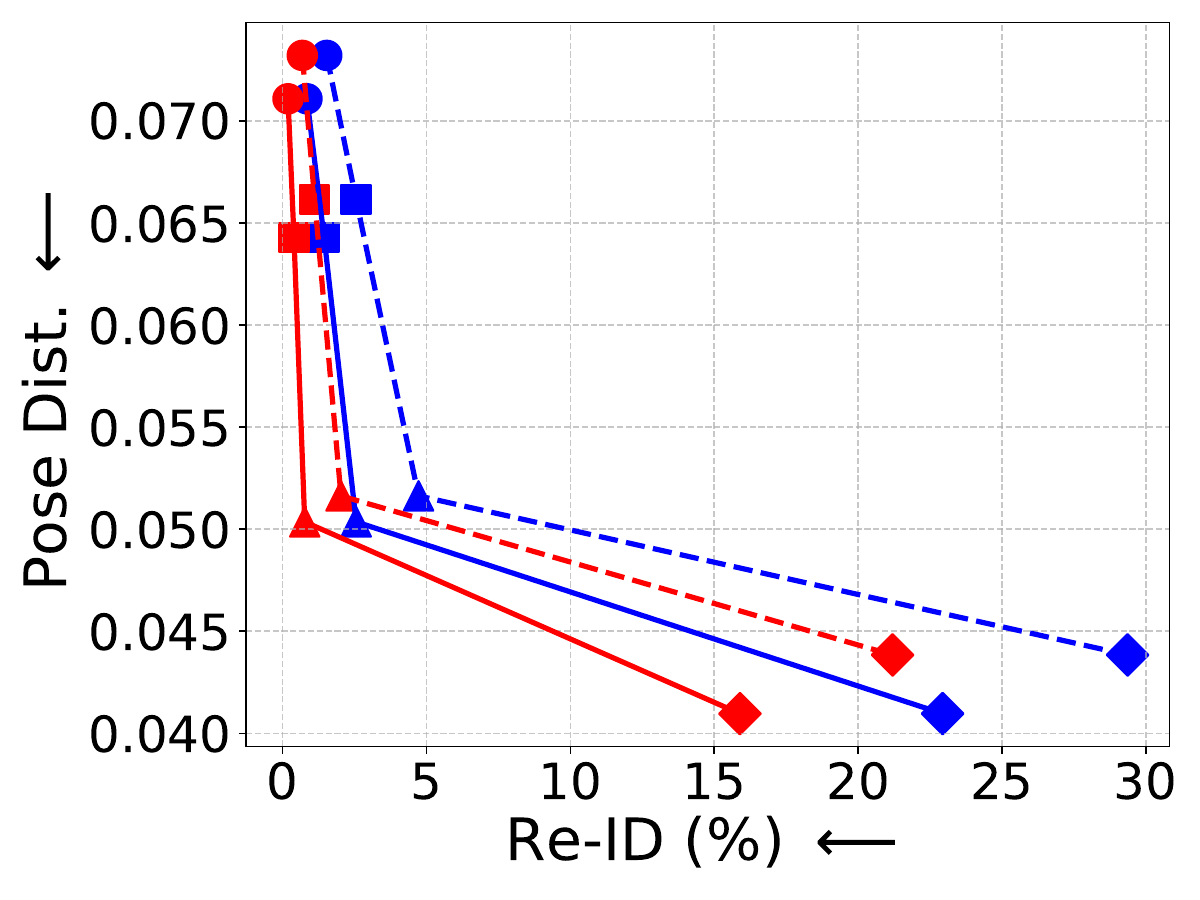}
    \caption{}
    \label{fig:reid_vs_pose}
  \end{subfigure}
  \hfill
  \begin{subfigure}{0.48\linewidth}
    \includegraphics[width=1.0\linewidth]{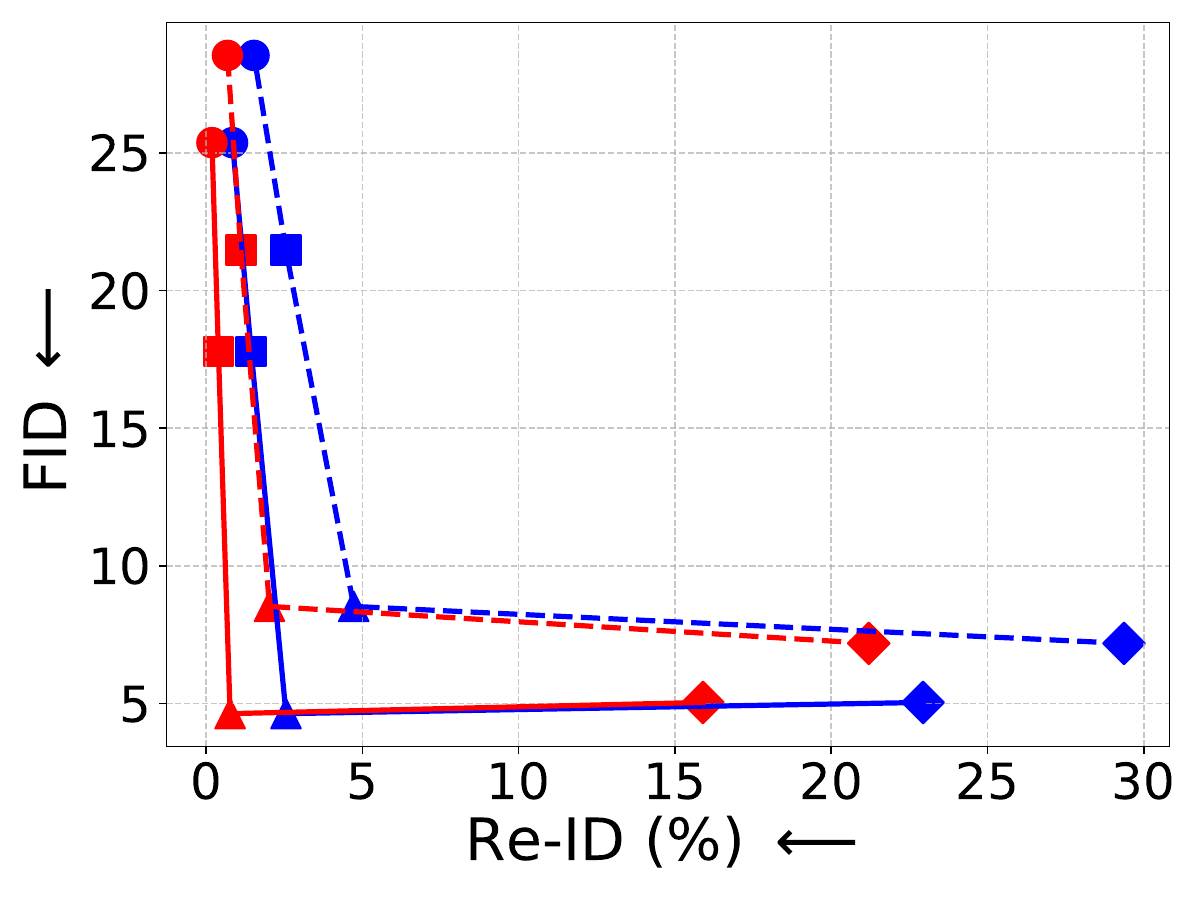}
    \caption{}
    \label{fig:reid-vs-fid}
  \end{subfigure}
  \\
  \begin{subfigure}{0.48\linewidth}
    \includegraphics[width=1.0\linewidth]{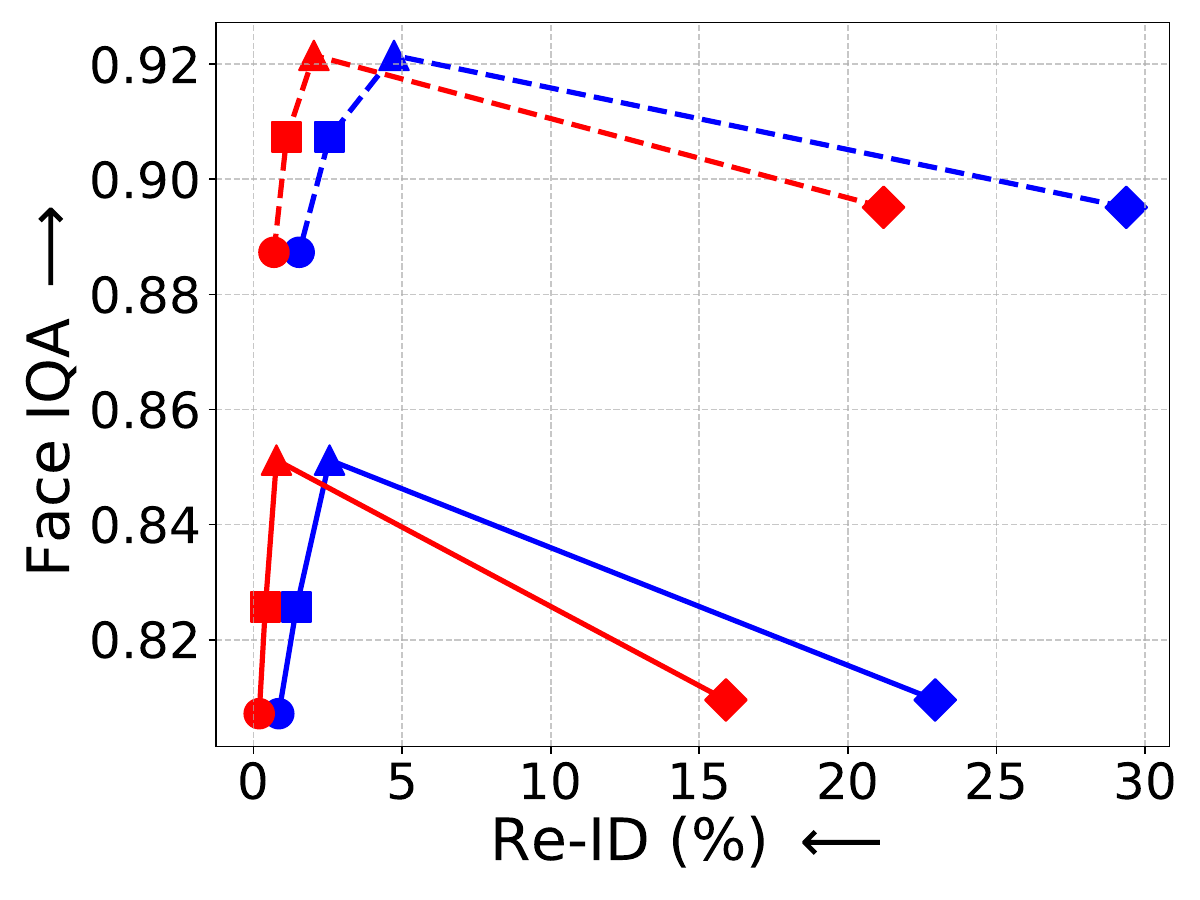}
    \caption{}
    \label{fig:reid-vs-faceiqa}
  \end{subfigure}
  \hfill
  \begin{subfigure}{0.48\linewidth}
    \includegraphics[width=1.0\linewidth]{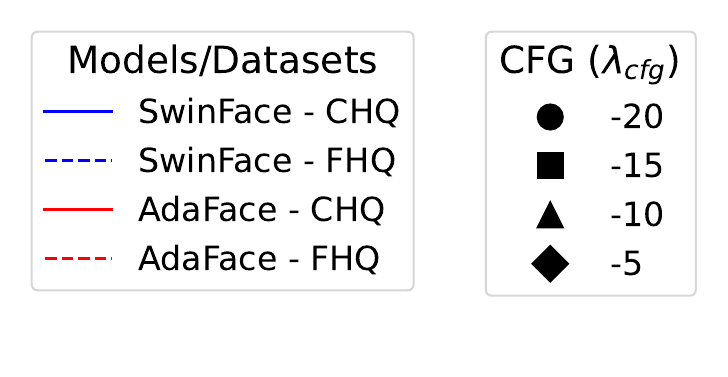}
    \caption{}
    \label{fig:legends}
  \end{subfigure}
  \caption{Privacy–utility trade-off plots showing that increasingly negative guidance scales lower re-identification rates but also degrade image quality and distort non-identity attributes (expression, gaze, pose).}
  \label{fig:privacy–utility-trade-off}
\end{figure}

\Cref{fig:privacy–utility-trade-off} shows the effect of negative classifier-free guidance~\cite{ho2022classifier} scales. Re-identification rates, computed with AdaFace~\cite{kim2022adaface} and SwinFace~\cite{qin2023swinface}, are compared to utility measures: expression distance, gaze distance, pose distance, FID~\cite{heusel2017gans}, and Face IQA~\cite{chen2024topiq}. Points represent different guidance scales (-20, -15, -10, -5). More negative scales reduce re-identification rates but increase attribute distances (expression, gaze, pose) and degrade image quality (higher  FID~\cite{heusel2017gans}, lower Face IQA~\cite{chen2024topiq}). This trade-off occurs because excessively negative guidance values impose the synthetic identity too strongly, altering non-identity attributes that should remain unchanged. Consistent with prior findings~\cite{sadat2024eliminating}, excessive classifier-free guidance~\cite{ho2022classifier} leads to overall image degradation.

\section{Challenging cases and limitations}

\begin{figure}[ht]
	\centering
	\begin{tabularx}{\linewidth}{@{}X@{}X@{}X@{}X@{}X@{}X@{}X@{}}
		\multicolumn{7}{@{}c@{}}{\includegraphics[width=\linewidth]{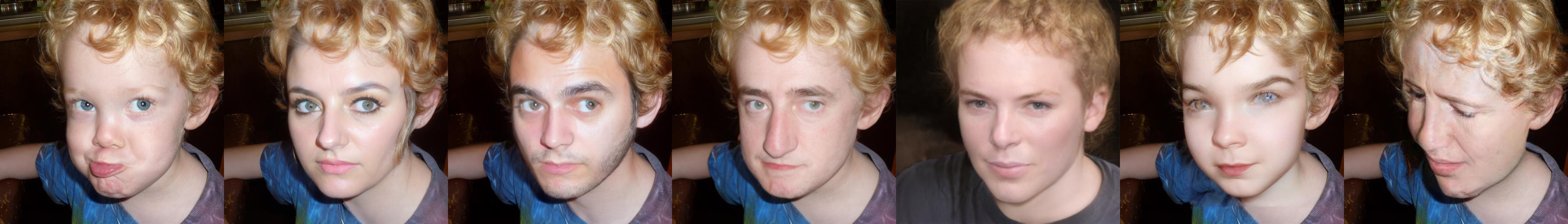}} \\
		\multicolumn{7}{@{}c@{}}{\includegraphics[width=\linewidth]{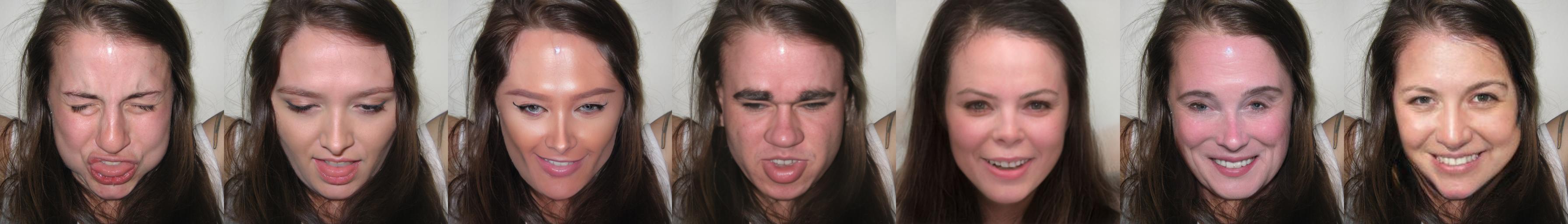}} \\
    \centering \scriptsize Input & \centering \scriptsize Ours & \centering \scriptsize \shortstack{NullFace \\ ~\cite{kung2025nullface}} & \centering \scriptsize \shortstack{FAMS \\ ~\cite{Kung_2025_WACV}} & \centering \scriptsize \shortstack{RiDDLE \\ ~\cite{li2023riddle}} & \centering \scriptsize \shortstack{LDFA \\ ~\cite{klemp2023ldfa}} & \centering \scriptsize \shortstack{DP2 \\ ~\cite{hukkelaas2023deepprivacy2}} \\
	\end{tabularx}
  \caption{Examples where our method and baselines fail to preserve rare or extreme facial expressions.}
	\label{fig:limitations}
\end{figure}

While prior methods struggle with extreme face angles or occlusions~\cite{hukkelaas2019deepprivacy,yang2024g}, our approach performs reliably in these cases. The greater challenge occurs with extreme or uncommon expressions (see \cref{fig:limitations}). In such cases, both our method and baselines struggle to preserve the original expression, likely due to limited training data for such expressions.

Image quality is also bounded by the underlying diffusion model (SDXL~\cite{podell2023sdxl} in our experiments). Using more advanced diffusion models could further improve quality.

Our framework currently anonymizes single images. When applied to videos, it lacks temporal consistency. Extending the method to video anonymization using video diffusion models~\cite{blattmann2023stable} is a promising direction for future work.

\section{Identity recovery test}

A potential concern is whether the original identity can be recovered from an anonymized image by applying a negative classifier-free guidance~\cite{ho2022classifier} scale to reverse anonymization. We argue that this is not feasible, as the outcome of our method depends on the interaction of multiple components---including the model architecture, inversion process, and hyperparameter settings---which prevent reversibility.

\begin{figure}[ht]
	\centering
	\resizebox{0.85\linewidth}{!}{
		\begin{tabular}{
			*{3}{m{\dimexpr.25\linewidth-2\tabcolsep}}
			}
      \multicolumn{1}{c}{\small Input} & \multicolumn{1}{c}{\small Anonymized} & \multicolumn{1}{c}{\small Recovery test} \\
			\multicolumn{1}{c}{\includegraphics[width={\dimexpr.25\linewidth}]{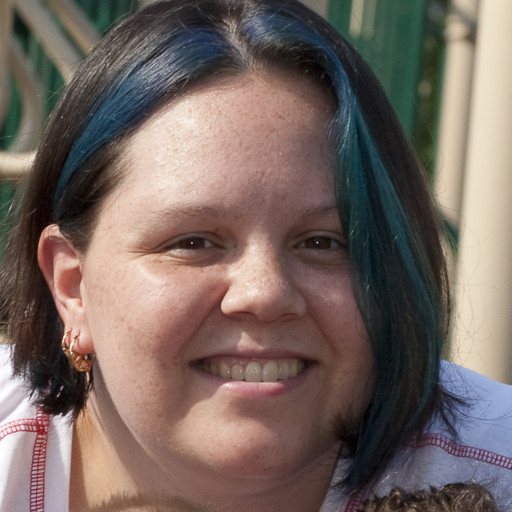}} &
      \multicolumn{1}{c}{\includegraphics[width={\dimexpr.25\linewidth}]{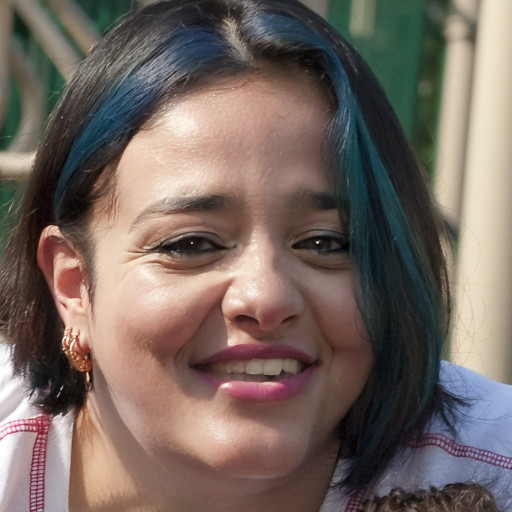}} &
      \multicolumn{1}{c}{\includegraphics[width={\dimexpr.25\linewidth}]{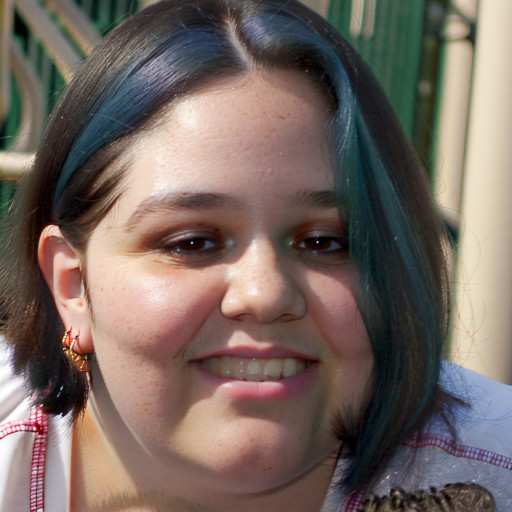}} \\
			\multicolumn{1}{c}{\includegraphics[width={\dimexpr.25\linewidth}]{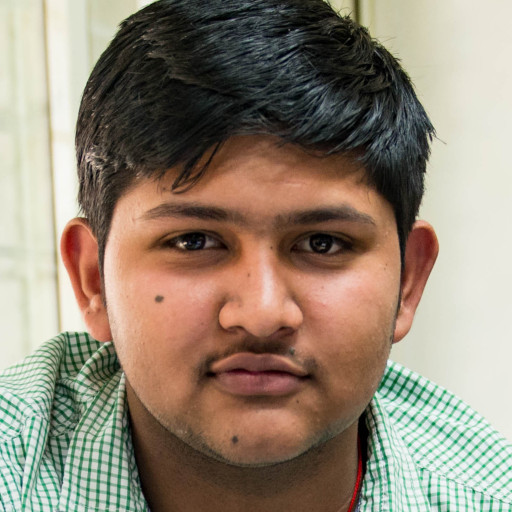}} &
      \multicolumn{1}{c}{\includegraphics[width={\dimexpr.25\linewidth}]{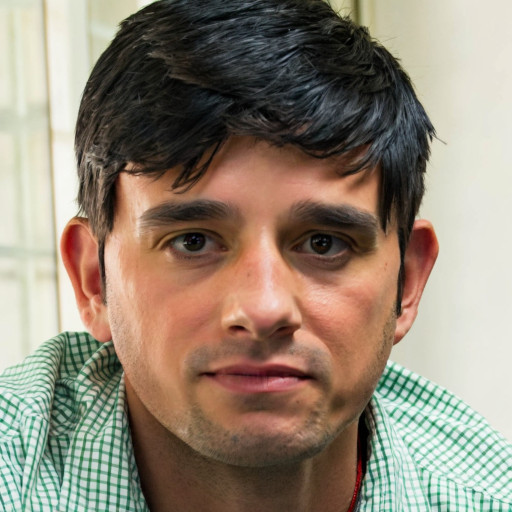}} &
      \multicolumn{1}{c}{\includegraphics[width={\dimexpr.25\linewidth}]{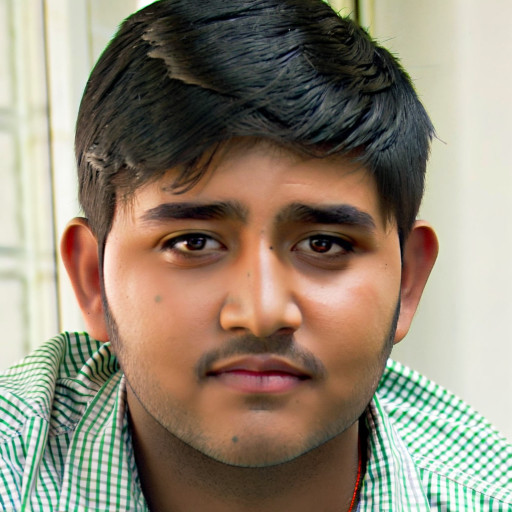}} \\
		\end{tabular}
	}
  \caption{Identity recovery attempts using negative classifier-free guidance~\cite{ho2022classifier} show that recovered images differ visually from the original inputs, demonstrating that the original identity cannot be restored.}
	\label{fig:recovery}
\end{figure}

We visualize recovery attempts in \cref{fig:recovery}. The recovered images remain visibly different from the original inputs, aligning with the quantitative findings in \cref{tab:recovery}, where re-identification rates remain low and comparable to anonymized images. These results demonstrate that the original identity cannot be restored through this approach.

\begin{table}[ht]
  \centering
	\resizebox{\linewidth}{!}{
    \begin{tabular}{@{}l cc cc@{}}
      \toprule
      & \multicolumn{4}{c}{Re-ID (\%) $\downarrow$} \\
      \cmidrule(lr){2-5}
      & \multicolumn{2}{c}{SwinFace} & \multicolumn{2}{c}{AdaFace} \\
      \cmidrule(lr){2-3} \cmidrule(lr){4-5}
      & CelebA-HQ & FFHQ & CelebA-HQ & FFHQ \\
      \midrule
      Anonymized & 2.556 & 4.724 & 0.768 & 2.028 \\
      Attempt to recover & 2.284 & 4.117 & 0.566 & 1.430 \\
      \bottomrule
    \end{tabular}
  }
  \caption{Identity recovery test by applying negative classifier-free guidance~\cite{ho2022classifier} to anonymized images. The low re-identification rates indicate that the original identity cannot be recovered through this method.}
  \label{tab:recovery}
\end{table}

\section{Societal risks of AI-generated human faces}

AI-generated human faces pose societal risks. Now nearly indistinguishable from real ones, such faces enable convincing fake identities on social media, fostering manipulation, fraud, and disinformation~\cite{ferrara2024genai}. These technologies also risk amplifying cultural, racial, and gender biases~\cite{aldahoul2025ai}. When trained on imbalanced datasets, face-generation models often reinforce societal prejudices, producing skewed and exclusionary representations. AI-generated faces are also used for sexual exploitation, such as creating non-consensual deepfake pornography~\cite{umbach2024non} that targets and violates victims without consent. While AI-generated faces have promising applications, their misuse highlights the urgent need for coordinated action by policymakers and technologists to address these harms.

\section{Additional qualitative results}
\label{sec:additional-qualitative}

\begin{figure*}
  \centering
  \begin{tabularx}{\linewidth}{@{}X@{}X@{}X@{}X@{}X@{}X@{}X@{}}
    \multicolumn{7}{@{}c@{}}{\includegraphics[width=\linewidth]{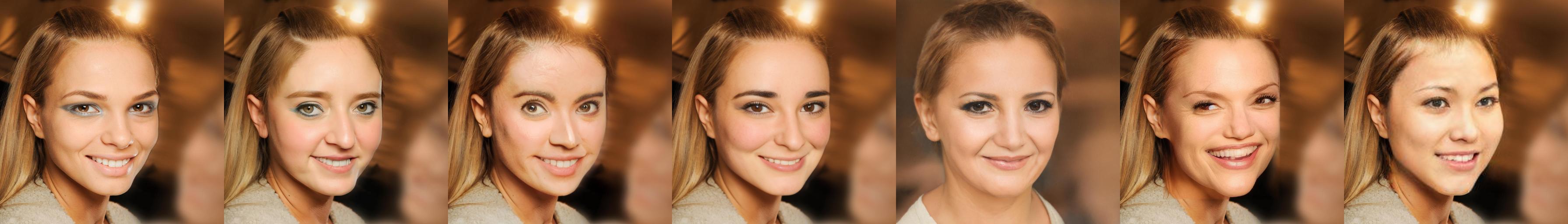}} \\
    \multicolumn{7}{@{}c@{}}{\includegraphics[width=\linewidth]{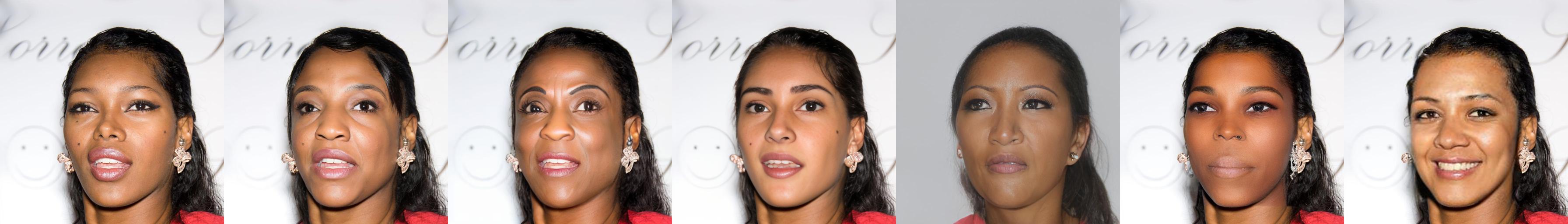}} \\
    \multicolumn{7}{@{}c@{}}{\includegraphics[width=\linewidth]{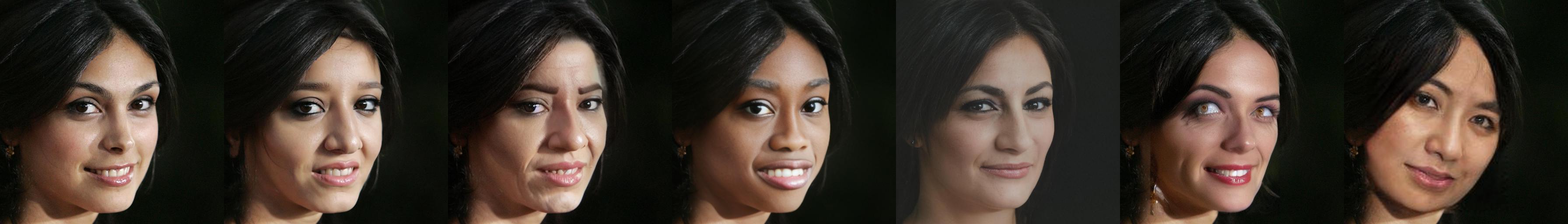}} \\
    \multicolumn{7}{@{}c@{}}{\includegraphics[width=\linewidth]{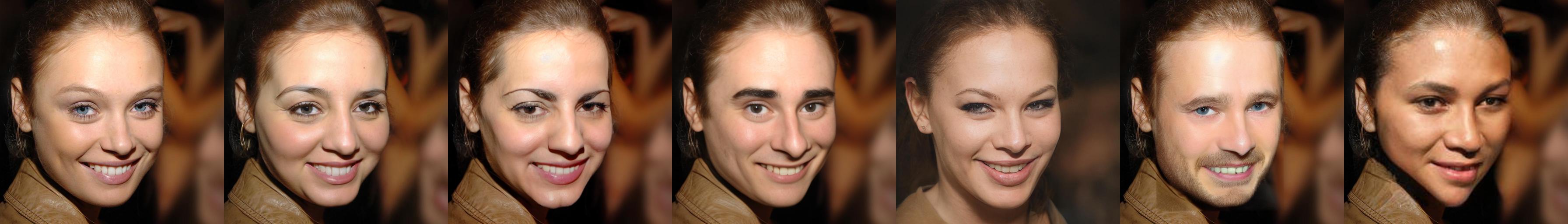}} \\
    \multicolumn{7}{@{}c@{}}{\includegraphics[width=\linewidth]{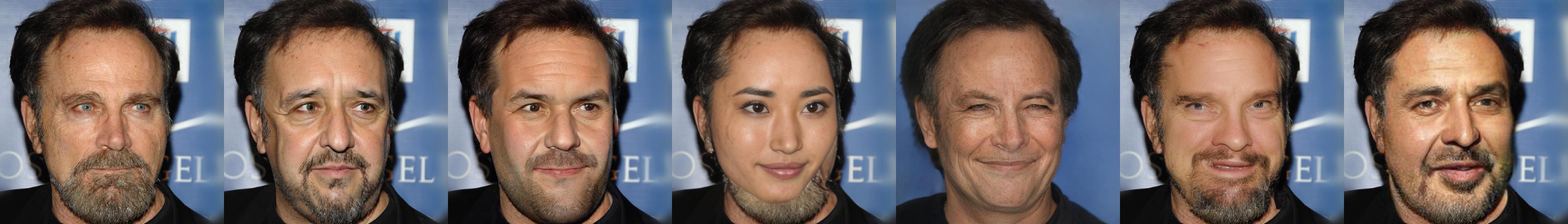}} \\
    \multicolumn{7}{@{}c@{}}{\includegraphics[width=\linewidth]{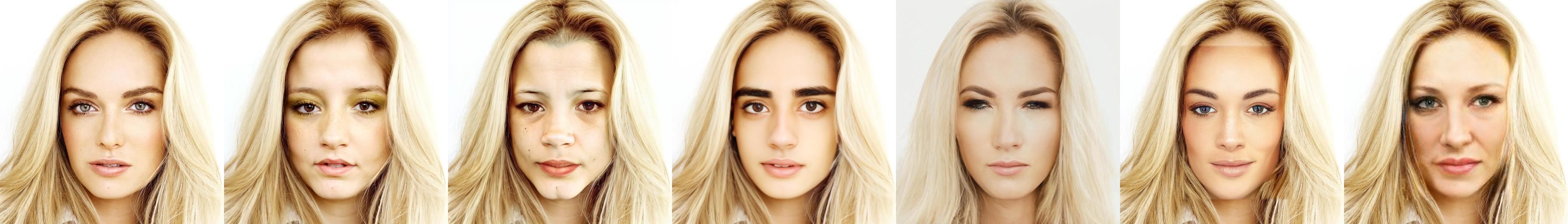}} \\
    \multicolumn{7}{@{}c@{}}{\includegraphics[width=\linewidth]{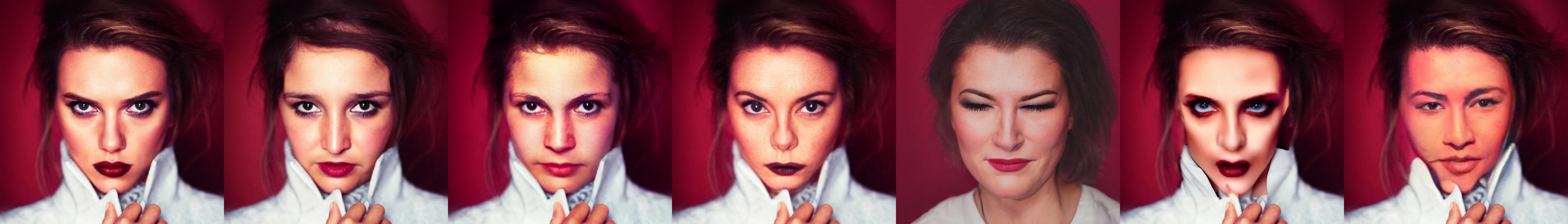}} \\
    \multicolumn{7}{@{}c@{}}{\includegraphics[width=\linewidth]{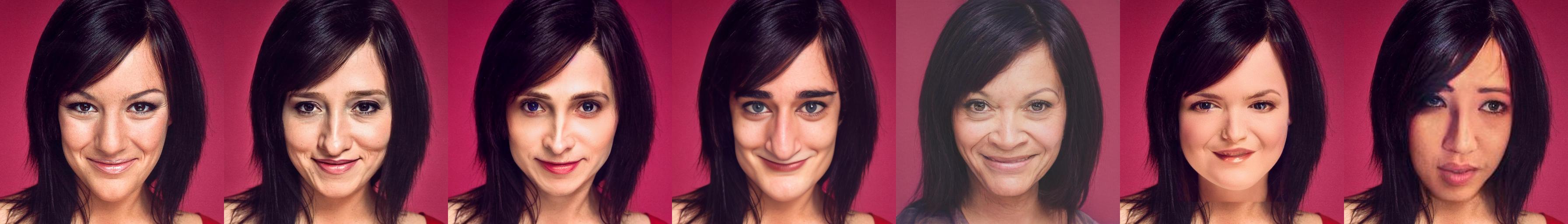}} \\
    \centering Input & \centering Ours & \centering NullFace~\cite{kung2025nullface} & \centering FAMS~\cite{Kung_2025_WACV} & \centering FALCO~\cite{barattin2023attribute} & \centering LDFA~\cite{klemp2023ldfa} & \centering DP2~\cite{hukkelaas2023deepprivacy2} \\
  \end{tabularx}
  \caption{Qualitative comparison of anonymization results on CelebA-HQ~\cite{karras2017progressive}.}
  \label{fig:comp_cele_plus_0}
\end{figure*}

\begin{figure*}
  \centering
  \begin{tabularx}{\linewidth}{@{}X@{}X@{}X@{}X@{}X@{}X@{}X@{}}
    \multicolumn{7}{@{}c@{}}{\includegraphics[width=\linewidth]{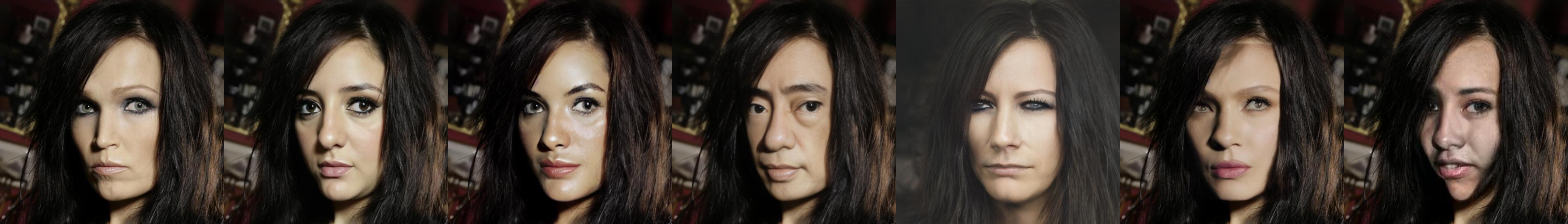}} \\
    \multicolumn{7}{@{}c@{}}{\includegraphics[width=\linewidth]{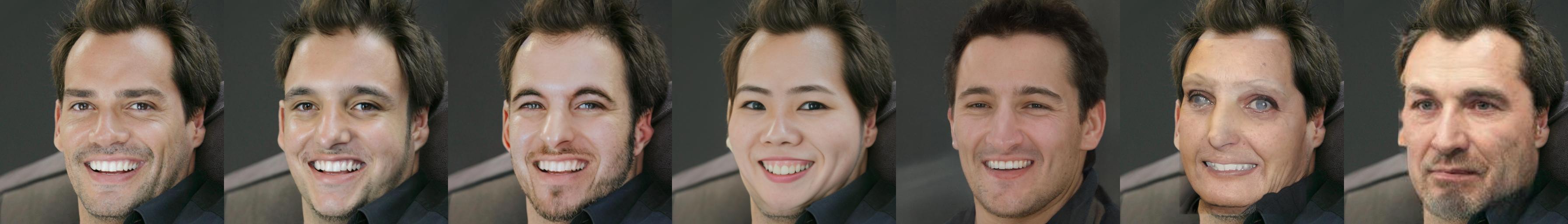}} \\
    \multicolumn{7}{@{}c@{}}{\includegraphics[width=\linewidth]{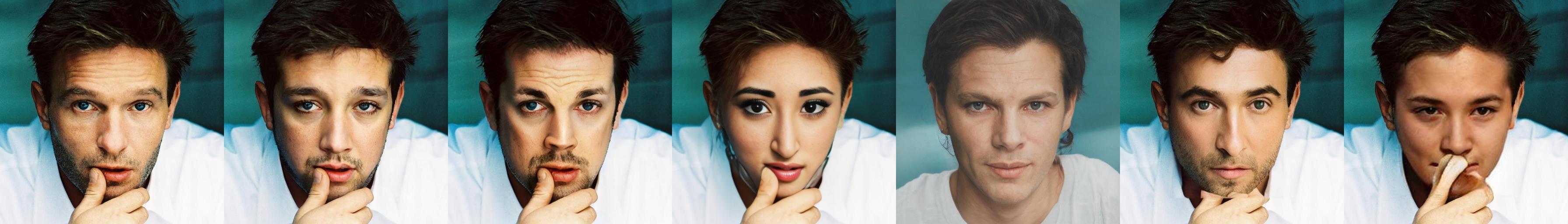}} \\
    \multicolumn{7}{@{}c@{}}{\includegraphics[width=\linewidth]{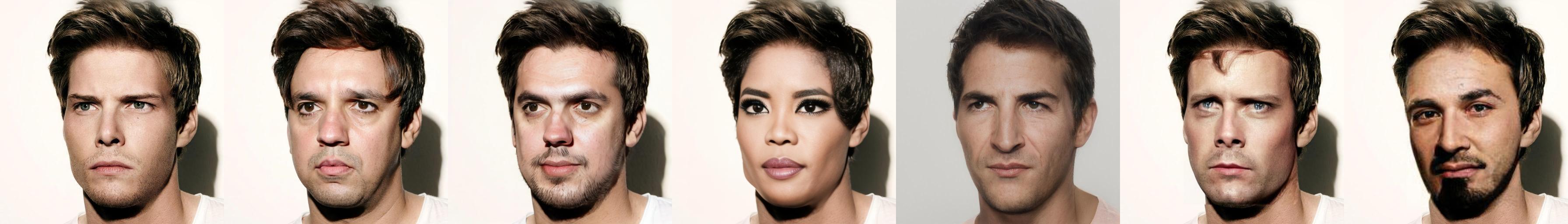}} \\
    \multicolumn{7}{@{}c@{}}{\includegraphics[width=\linewidth]{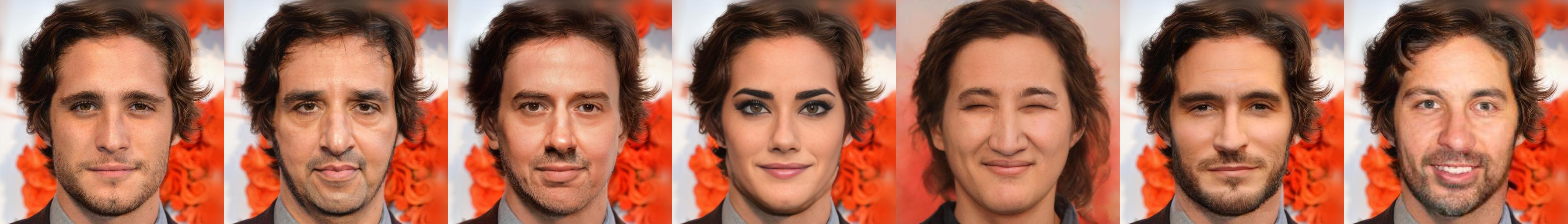}} \\
    \multicolumn{7}{@{}c@{}}{\includegraphics[width=\linewidth]{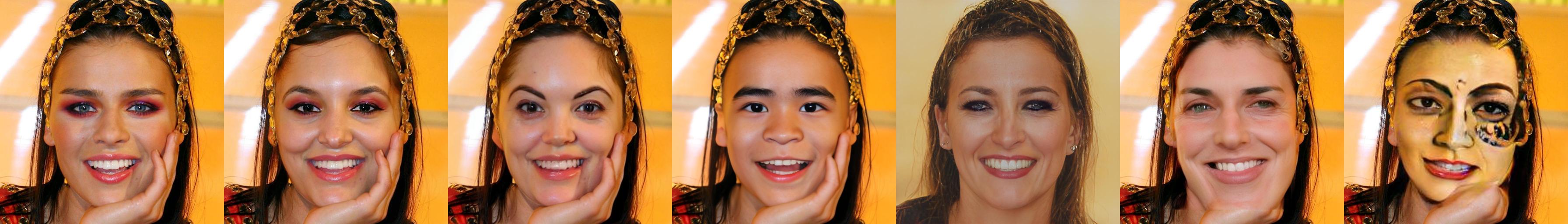}} \\
    \multicolumn{7}{@{}c@{}}{\includegraphics[width=\linewidth]{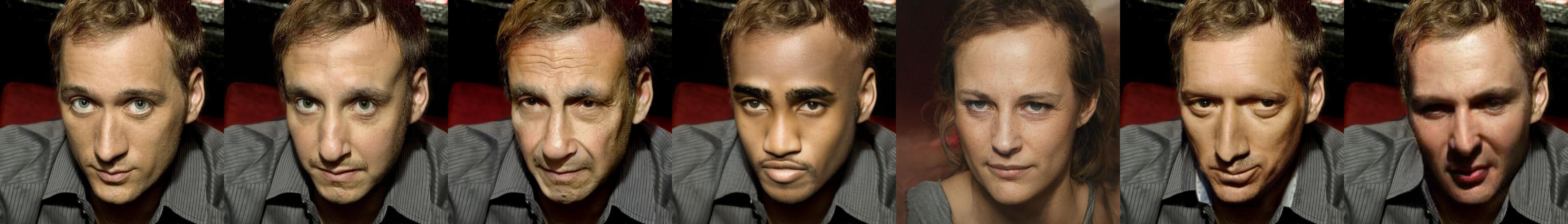}} \\
    \multicolumn{7}{@{}c@{}}{\includegraphics[width=\linewidth]{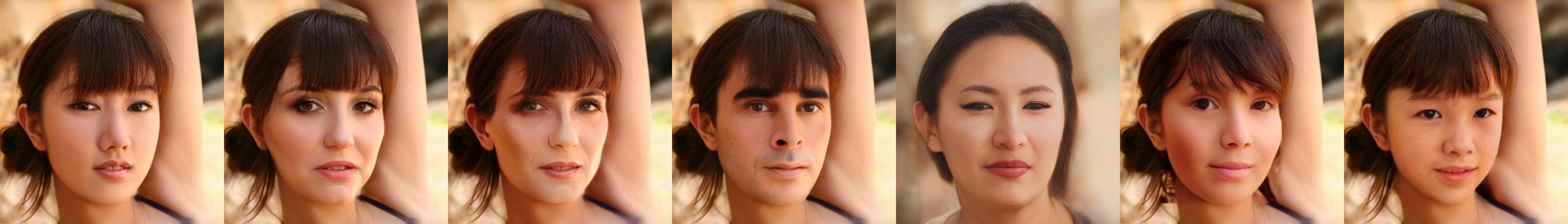}} \\
    \centering Input & \centering Ours & \centering NullFace~\cite{kung2025nullface} & \centering FAMS~\cite{Kung_2025_WACV} & \centering FALCO~\cite{barattin2023attribute} & \centering LDFA~\cite{klemp2023ldfa} & \centering DP2~\cite{hukkelaas2023deepprivacy2} \\
  \end{tabularx}
  \caption{Qualitative comparison of anonymization results on CelebA-HQ~\cite{karras2017progressive}.}
  \label{fig:comp_cele_plus_1}
\end{figure*}

\begin{figure*}
  \centering
  \begin{tabularx}{\linewidth}{@{}X@{}X@{}X@{}X@{}X@{}X@{}X@{}}
    \multicolumn{7}{@{}c@{}}{\includegraphics[width=\linewidth]{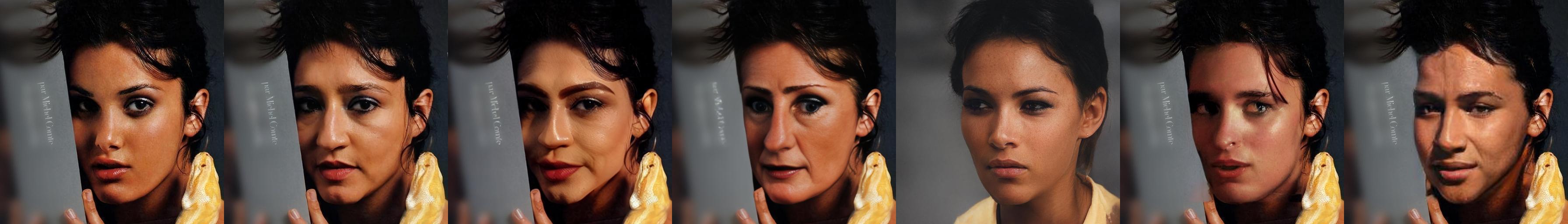}} \\
    \multicolumn{7}{@{}c@{}}{\includegraphics[width=\linewidth]{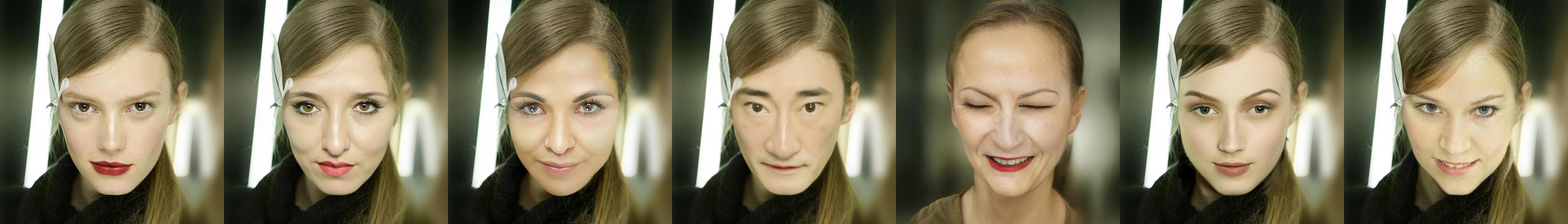}} \\
    \multicolumn{7}{@{}c@{}}{\includegraphics[width=\linewidth]{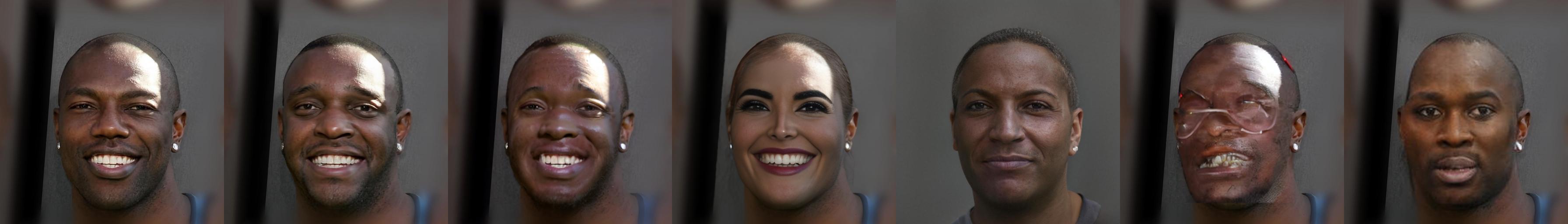}} \\
    \multicolumn{7}{@{}c@{}}{\includegraphics[width=\linewidth]{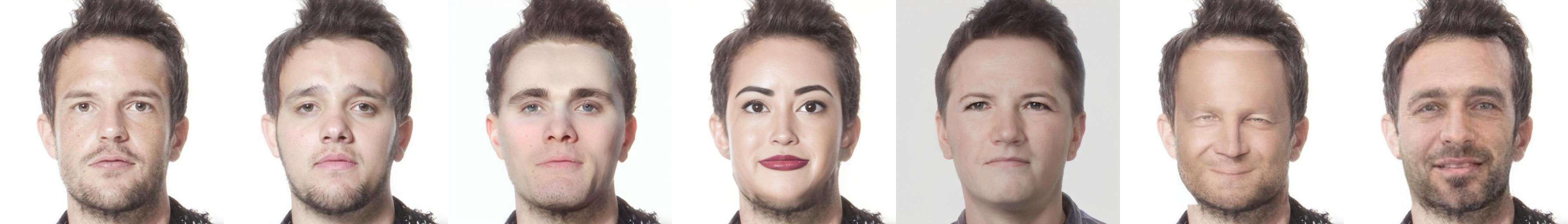}} \\
    \multicolumn{7}{@{}c@{}}{\includegraphics[width=\linewidth]{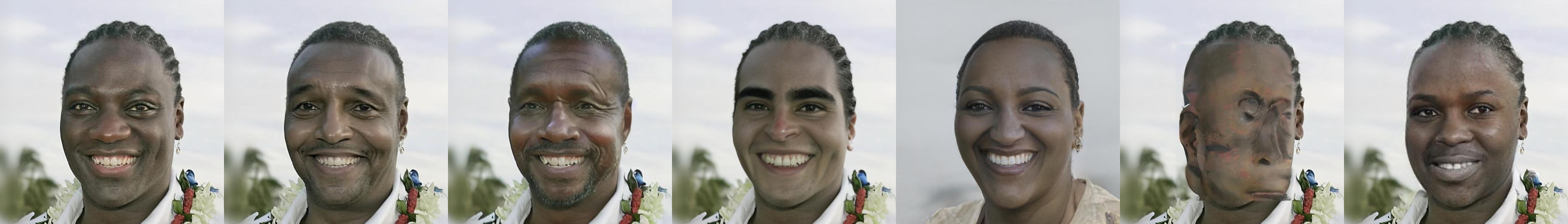}} \\
    \multicolumn{7}{@{}c@{}}{\includegraphics[width=\linewidth]{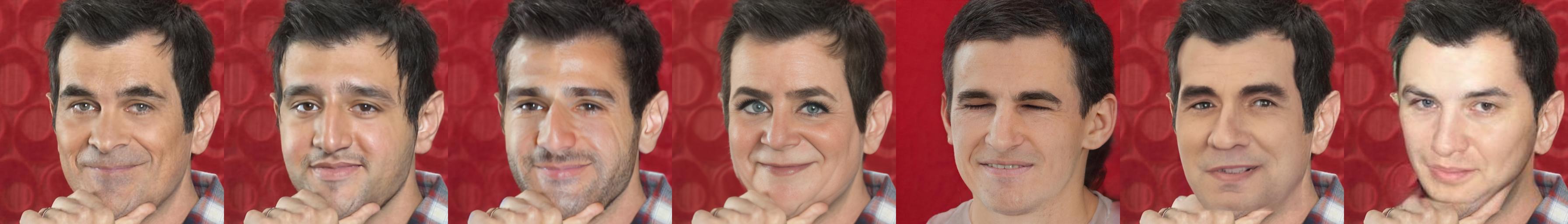}} \\
    \multicolumn{7}{@{}c@{}}{\includegraphics[width=\linewidth]{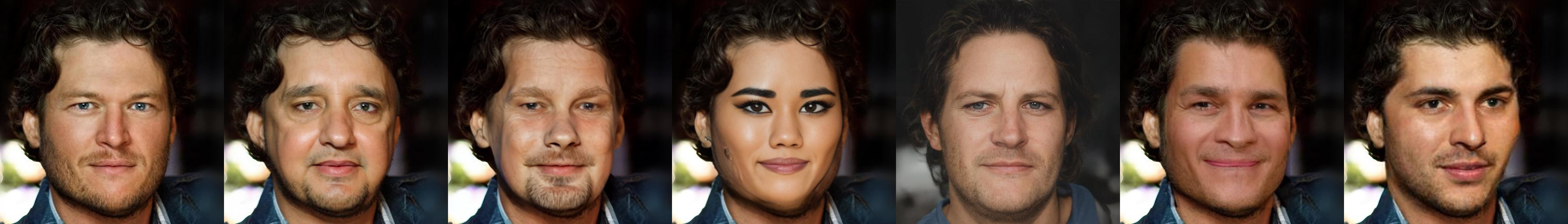}} \\
    \multicolumn{7}{@{}c@{}}{\includegraphics[width=\linewidth]{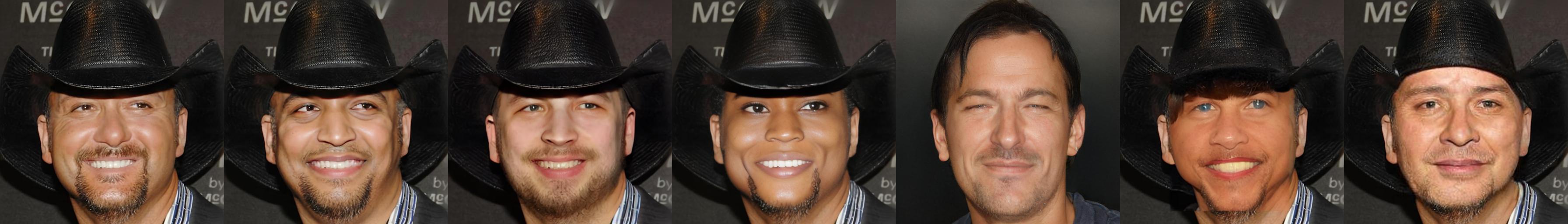}} \\
    \centering Input & \centering Ours & \centering NullFace~\cite{kung2025nullface} & \centering FAMS~\cite{Kung_2025_WACV} & \centering FALCO~\cite{barattin2023attribute} & \centering LDFA~\cite{klemp2023ldfa} & \centering DP2~\cite{hukkelaas2023deepprivacy2} \\
  \end{tabularx}
  \caption{Qualitative comparison of anonymization results on CelebA-HQ~\cite{karras2017progressive}.}
  \label{fig:comp_cele_plus_2}
\end{figure*}

\begin{figure*}
  \centering
  \begin{tabularx}{\linewidth}{@{}X@{}X@{}X@{}X@{}X@{}X@{}X@{}}
    \multicolumn{7}{@{}c@{}}{\includegraphics[width=\linewidth]{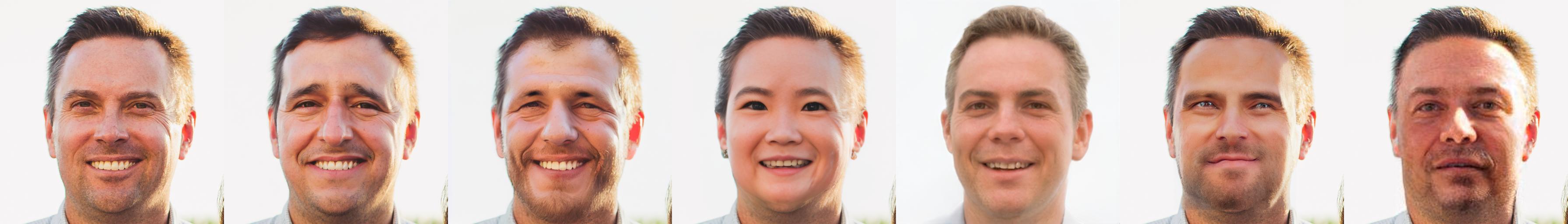}} \\
    \multicolumn{7}{@{}c@{}}{\includegraphics[width=\linewidth]{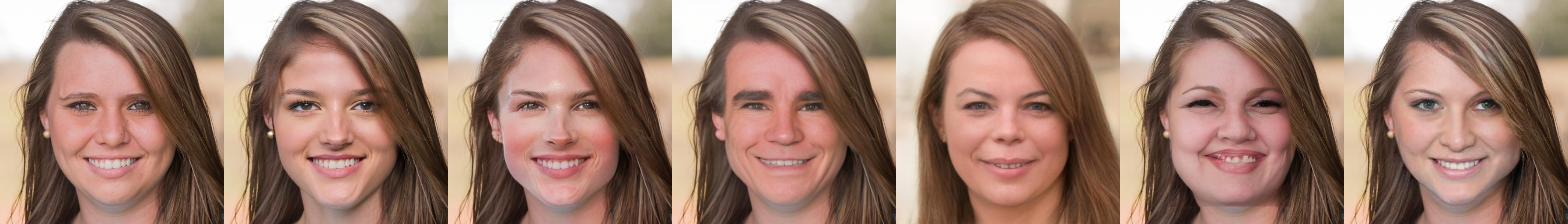}} \\
    \multicolumn{7}{@{}c@{}}{\includegraphics[width=\linewidth]{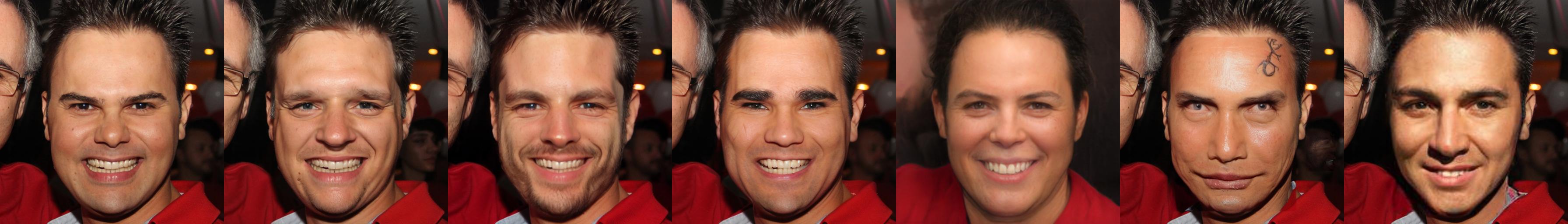}} \\
    \multicolumn{7}{@{}c@{}}{\includegraphics[width=\linewidth]{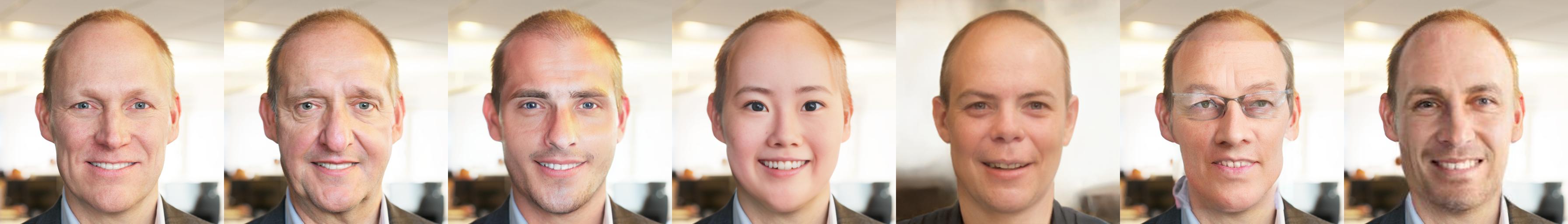}} \\
    \multicolumn{7}{@{}c@{}}{\includegraphics[width=\linewidth]{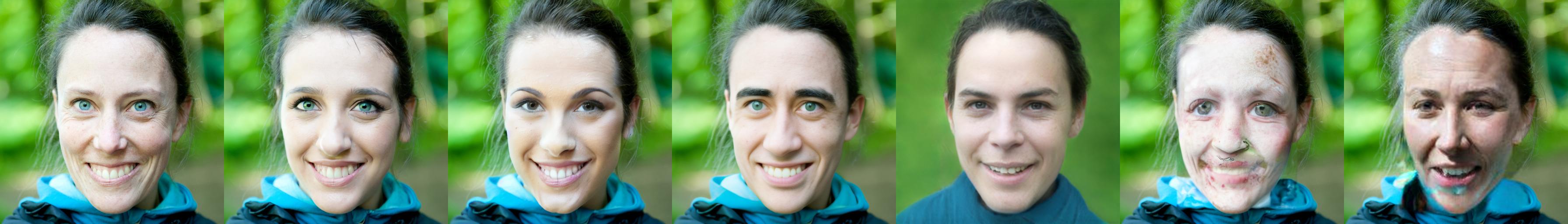}} \\
    \multicolumn{7}{@{}c@{}}{\includegraphics[width=\linewidth]{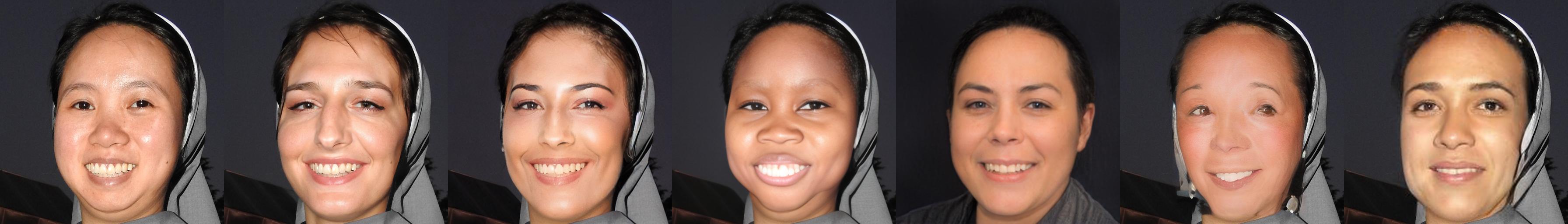}} \\
    \multicolumn{7}{@{}c@{}}{\includegraphics[width=\linewidth]{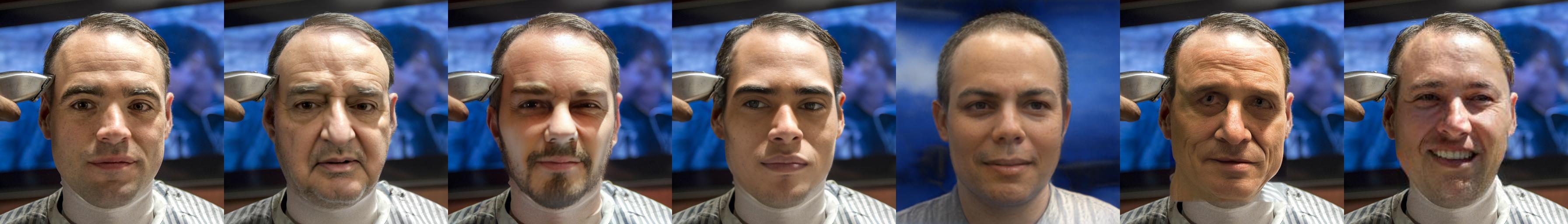}} \\
    \multicolumn{7}{@{}c@{}}{\includegraphics[width=\linewidth]{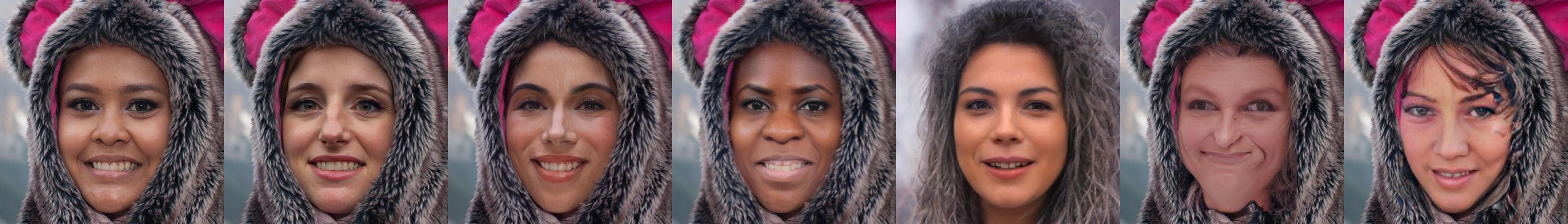}} \\
    \centering Input & \centering Ours & \centering NullFace~\cite{kung2025nullface} & \centering FAMS~\cite{Kung_2025_WACV} & \centering RiDDLE~\cite{li2023riddle} & \centering LDFA~\cite{klemp2023ldfa} & \centering DP2~\cite{hukkelaas2023deepprivacy2} \\
  \end{tabularx}
  \caption{Qualitative comparison of anonymization results on FFHQ~\cite{karras2019style}.}
  \label{fig:comp_ffhq_plus_0}
\end{figure*}

\begin{figure*}
  \centering
  \begin{tabularx}{\linewidth}{@{}X@{}X@{}X@{}X@{}X@{}X@{}X@{}}
    \multicolumn{7}{@{}c@{}}{\includegraphics[width=\linewidth]{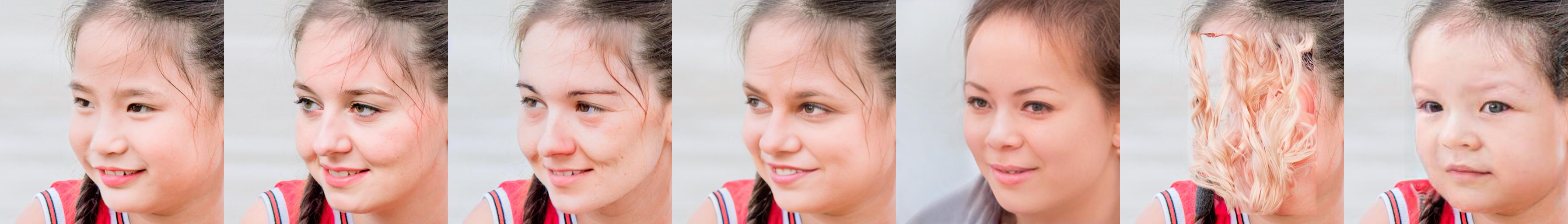}} \\
    \multicolumn{7}{@{}c@{}}{\includegraphics[width=\linewidth]{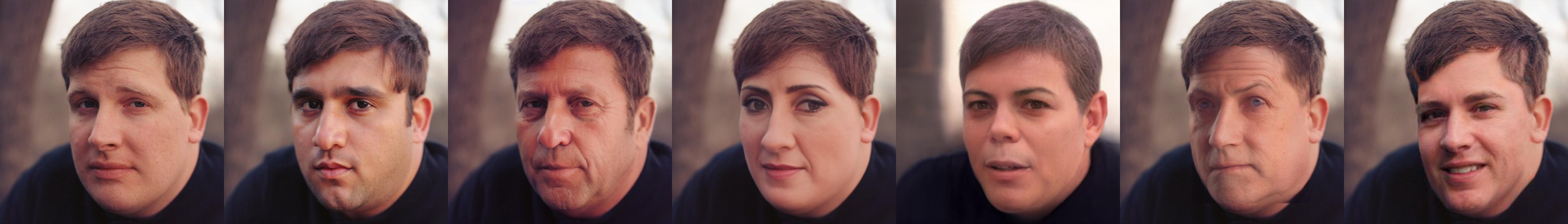}} \\
    \multicolumn{7}{@{}c@{}}{\includegraphics[width=\linewidth]{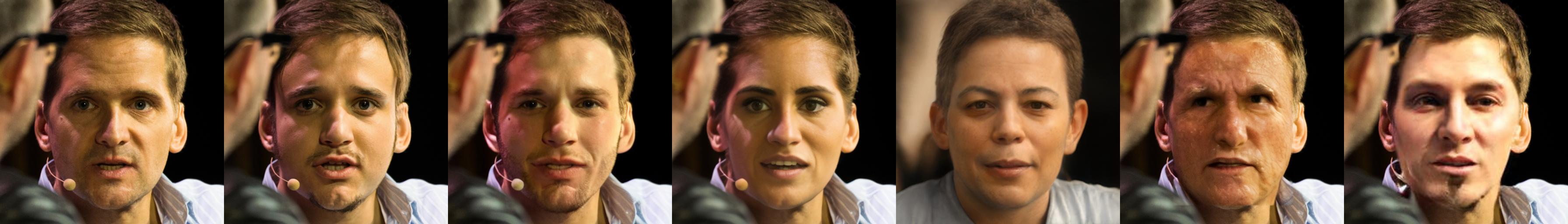}} \\
    \multicolumn{7}{@{}c@{}}{\includegraphics[width=\linewidth]{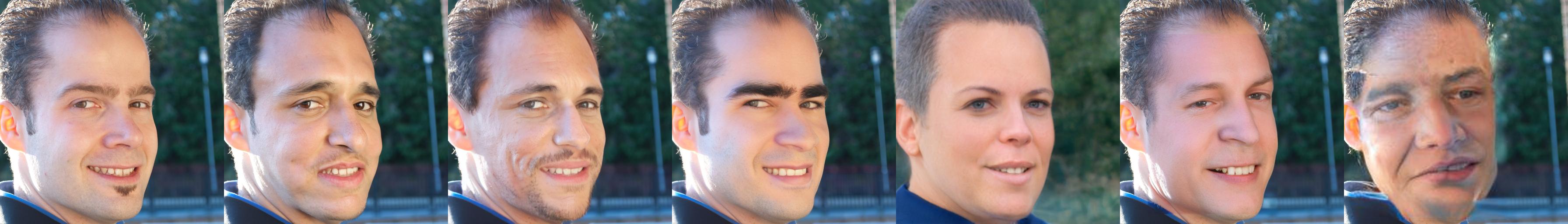}} \\
    \multicolumn{7}{@{}c@{}}{\includegraphics[width=\linewidth]{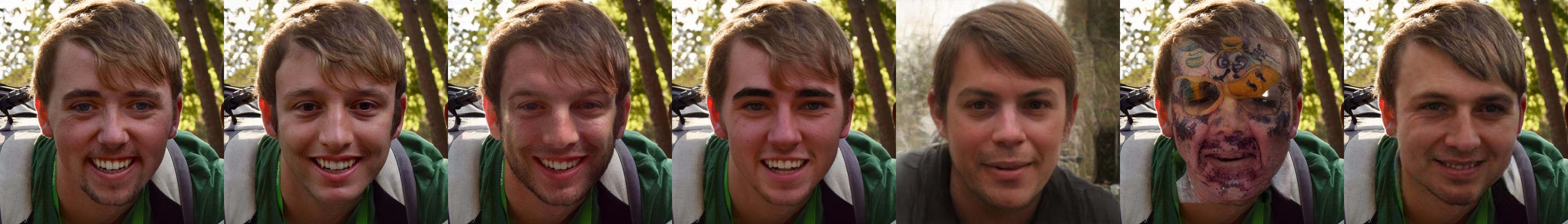}} \\
    \multicolumn{7}{@{}c@{}}{\includegraphics[width=\linewidth]{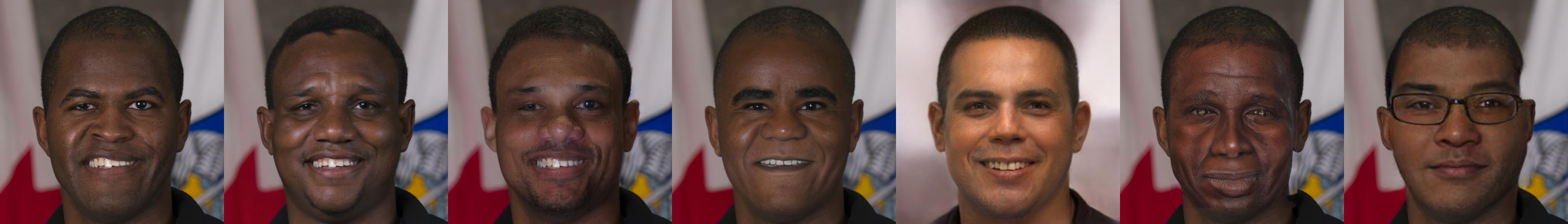}} \\
    \multicolumn{7}{@{}c@{}}{\includegraphics[width=\linewidth]{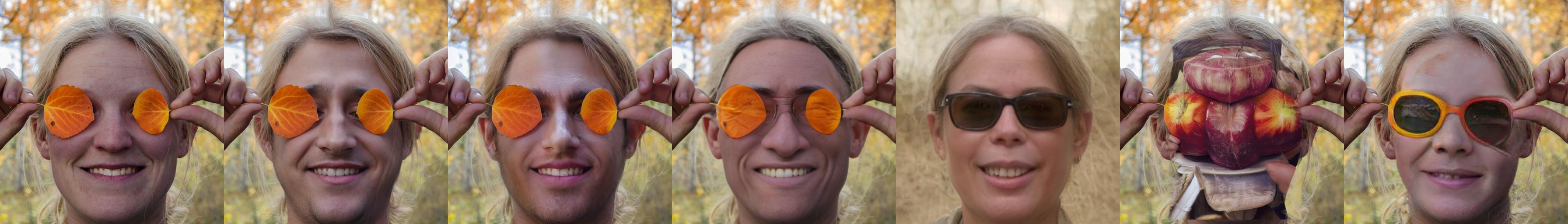}} \\
    \multicolumn{7}{@{}c@{}}{\includegraphics[width=\linewidth]{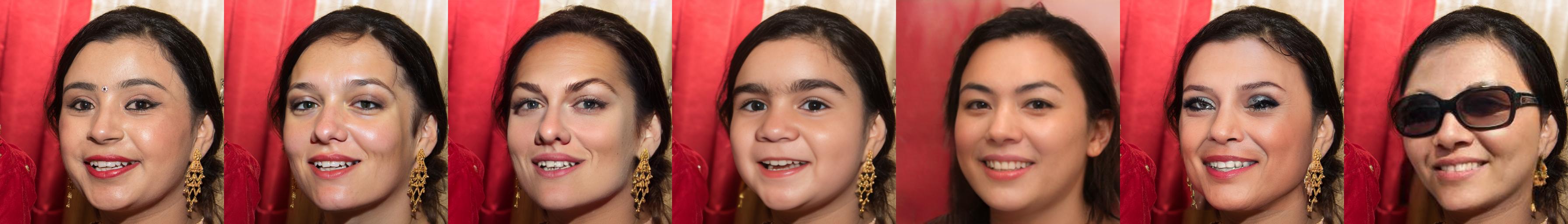}} \\
    \centering Input & \centering Ours & \centering NullFace~\cite{kung2025nullface} & \centering FAMS~\cite{Kung_2025_WACV} & \centering RiDDLE~\cite{li2023riddle} & \centering LDFA~\cite{klemp2023ldfa} & \centering DP2~\cite{hukkelaas2023deepprivacy2} \\
  \end{tabularx}
  \caption{Qualitative comparison of anonymization results on FFHQ~\cite{karras2019style}.}
  \label{fig:comp_ffhq_plus_1}
\end{figure*}

\begin{figure*}
  \centering
  \begin{tabularx}{\linewidth}{@{}X@{}X@{}X@{}X@{}X@{}X@{}X@{}}
    \multicolumn{7}{@{}c@{}}{\includegraphics[width=\linewidth]{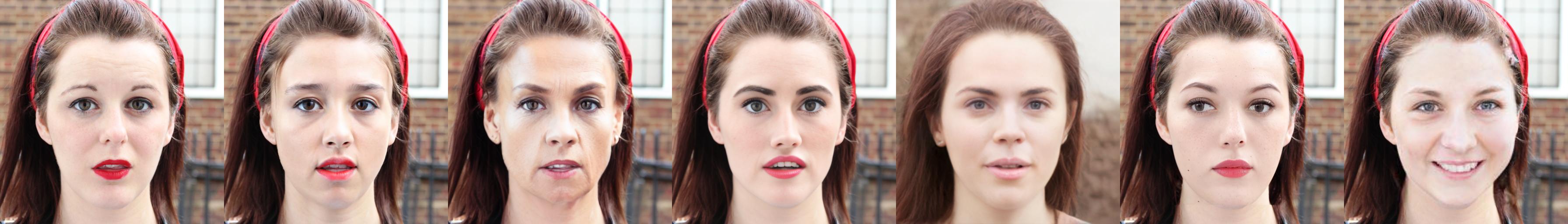}} \\
    \multicolumn{7}{@{}c@{}}{\includegraphics[width=\linewidth]{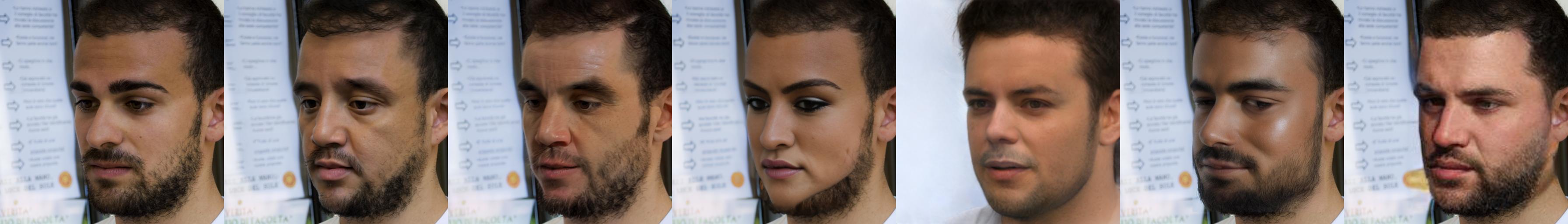}} \\
    \multicolumn{7}{@{}c@{}}{\includegraphics[width=\linewidth]{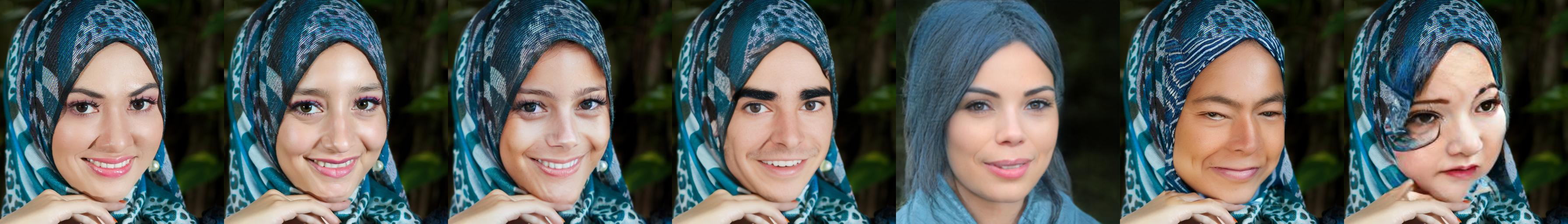}} \\
    \multicolumn{7}{@{}c@{}}{\includegraphics[width=\linewidth]{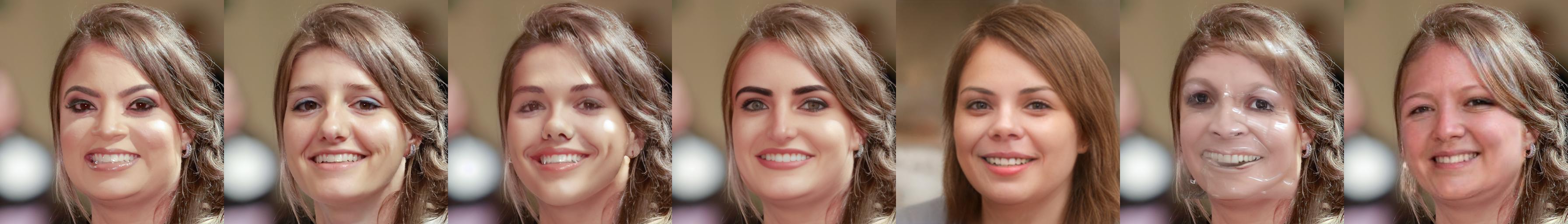}} \\
    \multicolumn{7}{@{}c@{}}{\includegraphics[width=\linewidth]{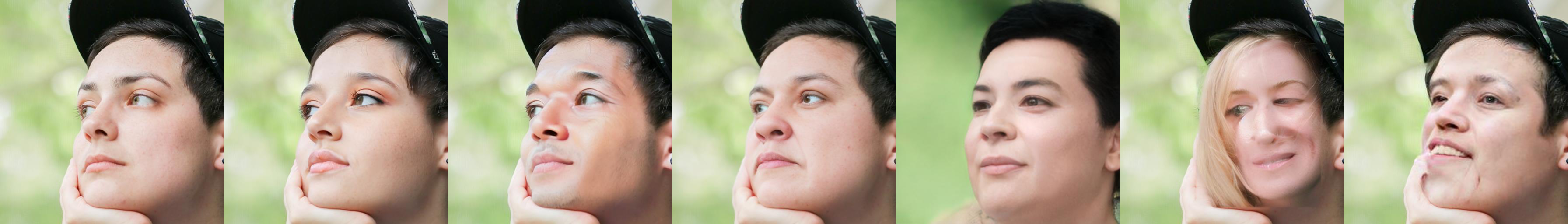}} \\
    \multicolumn{7}{@{}c@{}}{\includegraphics[width=\linewidth]{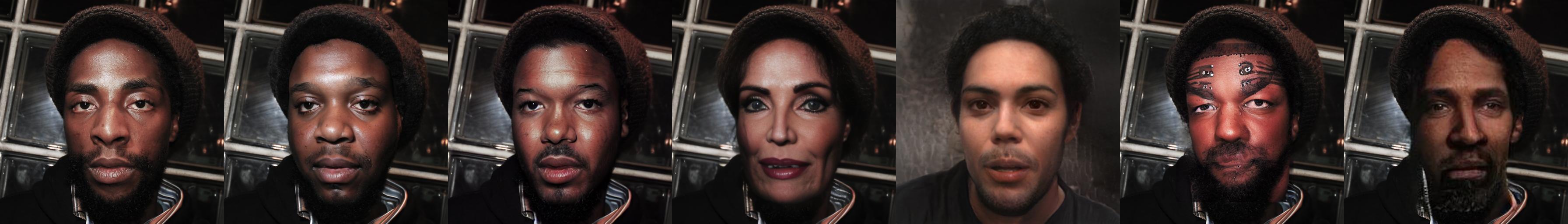}} \\
    \multicolumn{7}{@{}c@{}}{\includegraphics[width=\linewidth]{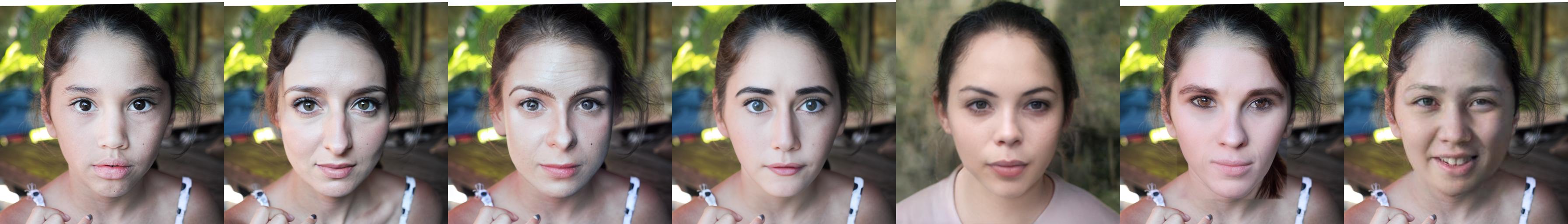}} \\
    \multicolumn{7}{@{}c@{}}{\includegraphics[width=\linewidth]{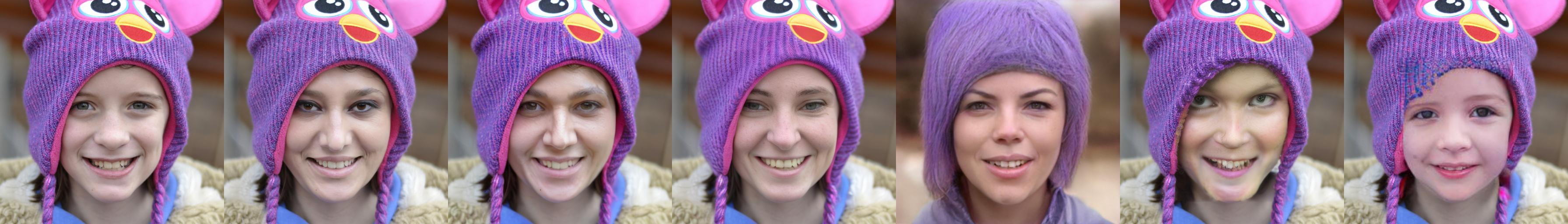}} \\
    \centering Input & \centering Ours & \centering NullFace~\cite{kung2025nullface} & \centering FAMS~\cite{Kung_2025_WACV} & \centering RiDDLE~\cite{li2023riddle} & \centering LDFA~\cite{klemp2023ldfa} & \centering DP2~\cite{hukkelaas2023deepprivacy2} \\
  \end{tabularx}
  \caption{Qualitative comparison of anonymization results on FFHQ~\cite{karras2019style}.}
  \label{fig:comp_ffhq_plus_2}
\end{figure*}

Qualitative comparisons between our method and six state-of-the-art approaches---NullFace~\cite{kung2025nullface}, FAMS~\cite{Kung_2025_WACV}, FALCO~\cite{barattin2023attribute}, RiDDLE~\cite{li2023riddle}, LDFA~\cite{klemp2023ldfa}, and DP2~\cite{hukkelaas2023deepprivacy2}---are shown for CelebA-HQ~\cite{karras2017progressive} and FFHQ~\cite{karras2019style}. CelebA-HQ~\cite{karras2017progressive} results are in \cref{fig:comp_cele_plus_0,fig:comp_cele_plus_1,fig:comp_cele_plus_2}, and FFHQ~\cite{karras2019style} results are in \cref{fig:comp_ffhq_plus_0,fig:comp_ffhq_plus_1,fig:comp_ffhq_plus_2}.

\end{document}